# COLIN: Planning with Continuous Linear Numeric Change

**Amanda Coles**                                       AMANDA.COLES@KCL.AC.UK
**Andrew Coles**                                       ANDREW.COLES@KCL.AC.UK
**Maria Fox**                                            MARIA.FOX@KCL.AC.UK
**Derek Long**                                         DEREK.LONG@KCL.AC.UK
*Department of Informatics, King's College London,*
*Strand, London WC2R 2LS, UK*

## Abstract

In this paper we describe COLIN, a forward-chaining heuristic search planner, capable of reasoning with COntinuous LINear numeric change, in addition to the full temporal semantics of PDDL2.1. Through this work we make two advances to the state-of-the-art in terms of expressive reasoning capabilities of planners: the handling of continuous linear change, and the handling of duration-dependent effects in combination with duration inequalities, both of which require tightly coupled temporal and numeric reasoning during planning. COLIN combines FF-style forward chaining search, with the use of a Linear Program (LP) to check the consistency of the interacting temporal and numeric constraints at each state. The LP is used to compute bounds on the values of variables in each state, reducing the range of actions that need to be considered for application. In addition, we develop an extension of the Temporal Relaxed Planning Graph heuristic of CRIKEY3, to support reasoning directly with continuous change. We extend the range of task variables considered to be suitable candidates for specifying the gradient of the continuous numeric change effected by an action. Finally, we explore the potential for employing mixed integer programming as a tool for optimising the timestamps of the actions in the plan, once a solution has been found. To support this, we further contribute a selection of extended benchmark domains that include continuous numeric effects. We present results for COLIN that demonstrate its scalability on a range of benchmarks, and compare to existing state-of-the-art planners.

## 1. Introduction

There has been considerable progress in the development of automated planning techniques for domains involving independent temporal and metric conditions and effects (Eyerich, Mattmüller, & Röger, 2009; Coles, Fox, Long, & Smith, 2008a; Gerevini, Saetti, & Serina, 2006; Edelkamp, 2003; Coles, Fox, Long, & Smith, 2008b). The development of powerful heuristics for propositional planning has been shown to offer benefits in the solution of extended planning problems, including planning under uncertainty (Palacios & Geffner, 2009), planning with numbers and planning with time. However, the combination and integration of metric and temporal features, in which metric quantities change in time-dependent ways, remains a challenge that has received relatively little attention.

Interaction between time and numbers in planning problems can occur in many ways. In the simplest case, using PDDL2.1 (Fox & Long, 2003), the numeric effects of actions are only updated instantaneously, and only at the start or end points of actions which are known (and fixed) at the point of action execution. The corpus of domains from past International Planning Competitions adhere to these restrictions. Time and numbers can interact in at least two more complex ways. First, actions can have variable, possibly constrained, durations and the (instantaneous) effects of these actions can depend on the values of the durations. This allows domain models to capture the effects of processes as discretised step effects, but adjusted according to the demands of specific problem instances. Second, the effects of actions can be considered to be continuous across their execution, so that the values of metric variables at any time point depend on how long the continuous effects have been acting on them.

For example, a problem in which sand is loaded into a lorry can be modelled so that the amount of sand loaded depends on the time spent loading. The first approach is to capture the increase in the quantity of loaded sand as a step function applied at the end of the loading action. In the second approach, the process of loading sand is modelled as a continuous and linear function of the time spent loading, so that the amount of sand in the lorry can be observed at any point throughout the loading process. If a safety device must be engaged before the lorry is more than three-quarters full, then only the second of these models will allow a planner to have the necessary access to the underlying process behaviour to make good planning choices about how to integrate this action into solutions. There are alternative models exploiting duration-dependent effects to split the loading action into two parts around the time point at which the safety device must be engaged, but these alternatives become very complicated with relatively modest changes to the domain.

Continuous change in both of these forms is common in many important problems. These include: energy management, the consumption and replenishment of restricted continuous resources such as fuel, tracking the progress of chemicals through storage tanks in chemical plants, choreographing robot motion with the execution of tasks, and managing the efficient use of time. In some cases, a model using discrete time-independent change is adequate for planning. However, discretisation is not always practical: to find a reasonable solution (or, indeed, to find one at all) identifying the appropriate granularity for discretisation is non-trivial, perhaps requiring a range of choices that are so fine-grained as to make the discrete model infeasibly large. In other cases, the numeric change cannot be appropriately discretised, where it is unavoidably necessary to have access to the values of numeric variables during the execution of actions, in order to manage interactions between numeric values.

In this paper we present a planner, COLIN, capable of reasoning with both variable, duration-dependent, linear change and linear continuous numeric effects. The key advance that COLIN makes is to be able to reason about time-dependent change through the use of linear programs that combine metric and temporal conditions and effects into the same representation. COLIN is a satisficing planner that attempts to build good quality solutions to this complex class of problems. Since COLIN is a forward-searching planner it requires a representation of states, a means to compute the progression of states and a heuristic function to guide the search for a path from the initial to the goal state. COLIN is built on the planner CRIKEY3 (Coles, Fox, Long et al., 2008a). However, CRIKEY3 requires numeric change to be discrete and cannot reason with continuous numeric change, or duration dependent change (where the duration of actions is not fixed in the state in which the action begins). Being able to reason successfully with problems characterised by continuous change, coping efficiently with a wide range of practical problems that are inspired by real applications, is the major contribution made by COLIN.

The organisation of the paper is as follows. In Section 2 we explain the features of PDDL2.1 that COLIN can handle, and contrast its repertoire with that of CRIKEY3. In Section 4 we define the problem that is addressed by COLIN. In Section 5 we outline the background in temporal and metric planning that supports COLIN, before, in Section 6, describing the details of the foundations of COLIN that lie in CRIKEY3. COLIN inherits its representation of states from CRIKEY3, as well as the machinery for confirming the temporal consistency of plans and the basis for the heuristic function. In Section 7 we describe systems in the literature that have addressed similar hybrid discrete-continuous planning problems to those that COLIN is designed to handle. Section 8 explains how state progression is extended in COLIN to handle linear continuous change, and Section 9 describes the heuristic that guides the search for solutions. In Section 10 we consider several elements of COLIN that improve both efficiency and plan quality, without affecting the fundamental behaviour of the planner. Since time-dependent numeric change has been so little explored, there are few benchmarks in existence that allow a full quantitative evaluation. We therefore present a collection of continuous domains that can be used for such analysis, and we show how COLIN fares on these. An appendix containing some explanations of technical detail and some detailed summaries of background work on which COLIN depends, ensures that the paper is complete and self-contained.

## 2. Language Features in CRIKEY3 and COLIN

COLIN builds on CRIKEY3 by handling the continuous features of PDDL2.1. CRIKEY3 was restricted to management of discrete change, while COLIN can handle the full range of linear continuous numeric effects. The only metric functions of PDDL2.1 that are not in the repertoire of COLIN are `scale-up` and `scale-down`, which are non-linear updates, and the general form of plan metrics. Managing plan metrics defined in terms of domain variables remains a challenge for planning that has not yet been fully confronted by any contemporary planner. COLIN does handle a restricted form of quality metric, which exploits an instrumented variable called `total-cost`. This allows COLIN to minimise the overall cost of the shortest plan it can find using `total-time` (the default metric used by most temporal planners).

In common with CRIKEY3, COLIN can cope with Timed Initial Literals, an important feature that was introduced in PDDL2.2 (Hoffmann & Edelkamp, 2005). PDDL2.1 is backward compatible with McDermott's PDDL (McDermott, 2000) and therefore supports ADL (Pednault, 1989). COLIN does not handle full ADL, but it can deal with a restricted form of conditional effect as seen in the airplane-landing problem described in section 11. This restricted form allows the cost of an action to be dependent on the state in which it is applied. More general forms of conditional effect cannot be handled.

With this collection of features, COLIN is able to fully manage both the discrete and continuous numeric change that occur directly as a result of its actions. PDDL+ (Fox & Long, 2006) further supports the modelling of continuous change brought about by exogenous processes and events. These are triggered by actions, but they model the independent continuous behaviour brought about by the world rather than by the planner's direct action. The key additional features of PDDL+ that support this are processes and events. COLIN does not handle these features but is restricted to the management of continuous change as expressed through the durative action device.

For detailed explanations of the syntaxes and semantics of PDDL2.1 and PDDL+, including the semantics on which implementations of state representation and state progression must be constructed, readers should refer to the work of Fox and Long (2003, 2006).

| Language | Language Feature | Crikey3 | Colin | Comment | Section |
|---|---|---|---|---|---|
| PDDL2.1 | Numeric conditions and effects | yes | yes | Basic treatment follows Metric-FF | Appendix B |
| PDDL2.1 | Continuous numeric effects | no | yes | Modification to state representation | Section 8 |
| | | | | Modification to heuristic | Section 9 |
| PDDL2.1 | General plan metrics | no | no | | |
| PDDL2.1 | Use of `total-cost` | no | yes | Limited form | Section 10 |
| PDDL2.1 | Assign (to discrete variables) | yes | yes | Treatment follows Metric-FF | |
| PDDL2.1 | Scale-up/down | no | no | | |
| PDDL2.1 | #t | no | yes | As continuous effects | |
| PDDL2.1 | Durative actions | yes | yes | Includes required concurrency | Section 6 and Appendix C |
| PDDL2.1 | Duration inequalities | limited | yes | Colin handles duration-dependent effects | Sections 8 and 9 |
| PDDL2.2 | TILs | yes | yes | | Section 6 |
| PDDL | Conditional Effects | no | partial | Only for limited effects | Section 10 |
| PDDL | Other ADL | no | no | | |

Table 1: Language features handled by Crikey3 and Colin.

## 3. Motivation

There are a number of accounts of planning having been successfully applied to real problems, and the frequency with which applications are reported is increasing. The following examples involve domains with hybrid discrete-continuous dynamics. These dynamics are typically being dealt with by discretising time, packaging continuous numeric effects into step functions, or integrating propositional planning techniques with specialised solvers. They are all examples in which hybrid discrete-continuous reasoning could be exploited to improve plan quality or solution time.

- Operations of refineries (Boddy & Johnson, 2002; Lamba, Dietz, Johnson, & Boddy, 2003) or chemical plants (Penna, Intrigila, Magazzeni, & Mercorio, 2010), where the continuous processes reflect flows of materials, mixing and chemical reactions, heating and cooling.

- Management of power and thermal energy in aerospace applications in which power management is critical, such as management of the solar panel arrays on the International Space Station (Knight, Schaffer, & B.Clement, 2009; Reddy, Frank, Iatauro, Boyce, Kürklü, Ai-Chang, & Jónsson, 2011). For example, Knight et al. (2009) rely on a high-fidelity power model (TurboSpeed) to provide support for reasoning about the continuous power supply in different configurations of the solar panels. Power management is a critical problem for most space applications (including planetary rovers and landers, inspiring the temporal-metric-continuous Rovers domain used as one of our benchmark evaluation domains in Section 11). Chien et al. (2010) describe the planner used to support operations on Earth Observing 1 (EO-1), where the management of thermal energy generated by instruments is sufficiently important that the on-board planner uses some of its (highly constrained) CPU cycles to model and track its value. EO-1 inspires the temporal-metric-continuous Satellite benchmark described in Section 11.

- Management of non-renewable power in other contexts, such as for battery powered devices. The battery management problem described by Fox et al. (2011) relies on a non-linear model, which COLIN must currently reduce to a discrete or linear approximation, coupled with iterated validation and solution refinement, in order to optimise power use. Battery management is an example of a continuous problem that cannot be solved if the continuous dynamics are removed.

- Assignment of time-dependent costs as in the Aircraft Landing domain (Dierks, 2005), in which continuous processes govern the changing costs of the use of the runway as the landing time deviates from the optimal landing time for each aircraft. This problem inspires the Aircraft-Landing benchmark domain described in Section 11.

- Choreography of mobile robotic systems: in many cases, operations of robotic platforms involve careful management of motion alongside other tasks, where the continuous motion of the robot constrains the accessibility of specific tasks, such as inspection or observation. Existing examples of hybrid discrete-continuous planning models and reasoning for problems of this kind include work using flow tubes to capture the constraints on continuous processes (Léauté & Williams, 2005; Li & Williams, 2008). Problems involving autonomous underwater vehicles (AUVs) inspired the temporal-metric-continuous AUV benchmark presented in Section 11.

## 4. Problem Definition

COLIN is designed to solve a class of problems that are temporal and metric, and that feature linear continuous metric change. We refer to this as the class of temporal-metric-continuous problems, and it contains a substantial subset of the problems that can be expressed in PDDL2.1.

As a step towards the class of temporal-metric-continuous problems, we recall the definition of a simple temporal-metric planning problem — one in which there is no time-dependent metric change. Simple temporal-metric problems can be represented as a tuple $\langle I, A, G, M \rangle$, where:

- $I$ is the initial state: a set of propositions and an assignment of values to a set of numeric variables. Either of these sets may be empty. For notational convenience, we refer to the vector of numeric values in a given state as $\mathbf{v}$.

- $A$, a set of actions, each $\langle dur, pre_\vdash, eff_\vdash, pre_\leftrightarrow, pre_\dashv, eff_\dashv \rangle$, where:

    - $pre_\vdash$ ($pre_\dashv$) are the *start* (*end*) conditions of $a$: at the state in which $a$ starts (ends), these conditions must hold (for a detailed account of some of the subtleties in the semantics of action application, see Fox & Long, 2003).

    - $eff_\vdash$ ($eff_\dashv$) are the *start* (*end*) effects of $a$: starting (ending) $a$ updates the world state according to these effects. A given collection of effects $eff_x$, $x \in \{\vdash, \dashv\}$, consists of:
        * $eff_x^-$, propositions to be deleted from the world state;
        * $eff_x^+$, propositions to be added to the world state;
        * $eff_x^n$, effects acting upon numeric variables.

    - $pre_\leftrightarrow$ are the invariant conditions of $a$: these must hold at every point in the open interval between the start and end of $a$.

    - $dur$ are the *duration constraints* of $a$, calculated on the basis of the world state in which $a$ is started, and constraining the length of time that can pass between the start and end of $a$. They each refer to the special parameter `?duration`, denoting the duration of $a$.

- $G$, a goal: a set of propositions and conditions over numeric variables.

- optionally $M$, a metric optimisation function, defined as a function of the values of numeric variables at the end of the plan, and the special variable `total-time`, denoting the makespan of the plan.

A solution to such a problem is a time-stamped sequence of actions, with associated durations, that transforms the initial state into a state satisfying the goal, respecting all the conditions imposed. The durations of the actions must be specified explicitly, since it is possible that the action specifications can be satisfied by different duration values.

PDDL2.1 numeric conditions used in $pre_\vdash$, $pre_\dashv$, $pre_{\leftrightarrow}$, $dur$ and $G$ can be expressed in the form:

$$\langle f(\mathbf{v}), op, c \rangle, \quad \text{such that} \quad op \in \{\leq, <, =, >, \geq\}, c \in \Re$$

where $\mathbf{v}$ is the vector of metric fluents in the planning problem, $f(\mathbf{v})$ is a function applied to the vector of numeric fluents and $c$ is an arbitrary constant. Numeric effects used in $eff_\vdash$ and $eff_\dashv$ are expressed as:

$$\langle v, op, f(\mathbf{v}) \rangle, \quad \text{such that} \quad op \in \{\times=, +=, =, -=, \div=\}$$

A restricted form of numeric expressions is the set of expressions in Linear Normal Form (LNF). These are expressions in which $f(\mathbf{v})$ is a weighted sum of variables plus a constant, expressible in the form $\mathbf{w} \cdot \mathbf{v} + c$, for a vector of constants, $\mathbf{w}$. A notable consequence of permitting $dur$ to take the form of a set of LNF constraints over `?duration` is that `?duration` need not evaluate to a single fixed value. For instance, it may constrain the value of `?duration` to lie within a range of values, e.g. (`?duration` $\geq v_1$) $\wedge$ (`?duration` $\leq v_2$), for some numeric variables $v_1$ and $v_2$. Restricting conditions and effects to use only LNFs allows the metric expressions to be captured in a linear program model, a fact that we exploit in COLIN.

The class of temporal-metric problems is extended to temporal-metric-continuous problems by two additions:

1. Each action $a \in A$ is described with an additional component: a set of linear continuous numeric effects, $cont$, of the form $\langle v, k \rangle, k \in \Re$, denoting that $a$ increases $v$ at the rate of $k$ per unit of time. This corresponds to the PDDL2.1 effect (`increase (v) (* #t k)`).

2. The start or end effects of actions ($eff_\vdash^n$ and $eff_\dashv^n$ may, additionally, include the parameter `?duration`, denoting the duration of the action, and hence are written:

$$\langle v, op, \mathbf{w} \cdot \mathbf{v} + k.(\texttt{?duration}) + c \rangle \text{ s.t. } op \in \{+=, =, -=\}, c, k \in \Re$$

In temporal-metric-continuous problems the relationship between time and numbers is more complex than in temporal-metric problems. The first extension allows the value of a variable $v$ to depend on the length of time elapsed since the continuous effect acting upon it began. The second extension implies that, if `?duration` is not fixed, then the value of variables can depend on the duration assigned to the action. In fact , very few planners allow the literal `?duration` to appear in effects, even in actions where the value of the parameter is constrained to take a single fixed value by the duration constraint (e.g. (= ?duration 10)). A typical idiom is to name the intended value of the duration with a metric fluent in the initial state (e.g. (= (durationOfAction) 10)) and then use this fluent in the effects.

```
(:durative-action saveHard              (:durative-action lifeAudit
 :parameters ()                          :parameters ()
 :duration (= ?duration 10)              :duration (= ?duration (patience))
 :condition                              :condition
    (and (at start (canSave))               (and (at start (saving))
         (over all (>= (money) 0)))               (at end (boughtHouse))
 :effect                                          (at end (>= (money) 0)))
    (and (at start (not (canSave)))      :effect (and (at end (happy)))))
         (at end (canSave))
         (at start (saving))
         (at end (not (saving)))
         (increase (money) (* #t 1))))

(:durative-action takeMortgage
 :parameters (?m - mortgage)
 :duration (= ?duration (durationFor ?m))
 :condition
    (and (at start (saving))
         (at start (>= (money) (depositFor ?m)))
         (over all (<= (money) (maxSavings ?m))))
 :effect
    (and (at start (decrease (money) (depositFor ?m)))
         (decrease (money) (* #t (interestRateFor ?m)))
         (at end (boughtHouse))))
```

Figure 1: Actions for the Borrower Domain.

Temporal-metric-continuous problems form a significant subset of problems expressible in the PDDL+ language (Fox & Long, 2006), including those with linear continuous change within durative actions. The problems do not include non-linear continuous change, nor do they explicitly represent events or processes, although the use of certain modelling tricks can capture similar behaviours.

## 4.1 An Example Problem

As a running example of a temporal-metric-continuous domain we use the problem shown in Figure 1. In this, the Borrower Domain, a borrower can use a mortgage to buy a house. The domain is simplified in order to focus attention on some key aspects of continuous reasoning and is not proposed as a realistic application. Furthermore, the domain does not exploit variable duration actions, even though the ability to handle these is a key feature of COLIN. The example illustrates required concurrency, by means of interesting interactions between multiple actions affecting a single continuous variable, and allows us to demonstrate the differences between alternative heuristics described in Section 9. Management of required concurrency is also a key feature of COLIN, and domains with variable durations are discussed later in the paper.

In this domain, to obtain a mortgage it is necessary to have an appropriate active savings plan and to be able to lay down a deposit. These conditions are both achieved by saving hard, an action that cannot be applied in parallel with itself, preventing the borrower from building up capital at an arbitrarily high rate by multiple parallel applications of `saveHard`. For the sake of the example we restrict the saving periods to durations of 10 years to produce interesting interactions with the

```
(:objects shortMortgage longMortgage - mortgage)
(:init (= (money) 0)
        (canSave)
        (= (patience) 4)
        (= (depositFor shortMortgage) 5)
        (= (durationFor shortMortgage) 10)
        (= (interestRateFor shortMortgage) 0.5)
        (= (maxSavings shortMortgage) 6)
        (= (depositFor longMortgage) 1)
        (= (durationFor longMortgage) 12)
        (= (interestRateFor longMortgage) 0.75)
        (= (maxSavings longMortgage) 6))
(:goal (and (happy)))
(:metric minimize (total-time))
```

Figure 2: An example problem for the Borrower Domain.

durations of the mortgages in the sample problem. Once a person starts saving he or she is tied into a 10-year savings plan.

The constraint on being able to start a mortgage leads to required concurrency between saving and taking a mortgage. The effects of saving and repaying interest therefore combine to yield different linear effects on the value of the `money` variable, while the saving action requires this variable to remain non-negative throughout the duration of the `saveHard` action. Furthermore, in order to qualify for tax relief, each mortgage carries a maximum allowed level of savings throughout the mortgage (which prevents the mortgage being taken too late in the savings plan). Finally, the `lifeAudit` action places a constraint on the gap between the end of the saving action and the point at which the mortgage is completed (and also ensures that the borrower does not end up in debt). This action acknowledges that borrowers will only be happy if they manage to complete their mortgages within short periods (limited by their patience) of having to save hard.

The simple problem instance we will consider is shown in Figure 2. Two possible solutions to this are shown in Figure 3. In the first solution the borrower takes the longer mortgage, which has the advantage that it can start earlier because it requires a lower deposit. Money rises at rate 1 over the first part of the saving action, then decreases by 1 when the mortgage starts. It then rises at rate 0.25 (the difference between the saving and mortgage rates) until the saving action concludes, when it continues to decrease at rate 0.75 until the mortgage ends. The life audit action must start during a saving action and cannot end until after the end of a mortgage action. In the second solution the borrower takes the shorter mortgage, but that cannot start as early because it requires a much larger deposit. As a consequence, the life audit cannot start during the first saving action: the mortgage finishes too late to be included inside a life audit beginning within the first saving action. To meet the initial condition of the life audit, the borrower must therefore perform a second saving action to follow the first. Clearly the first solution is preferable since we are interested in minimising the makespan.

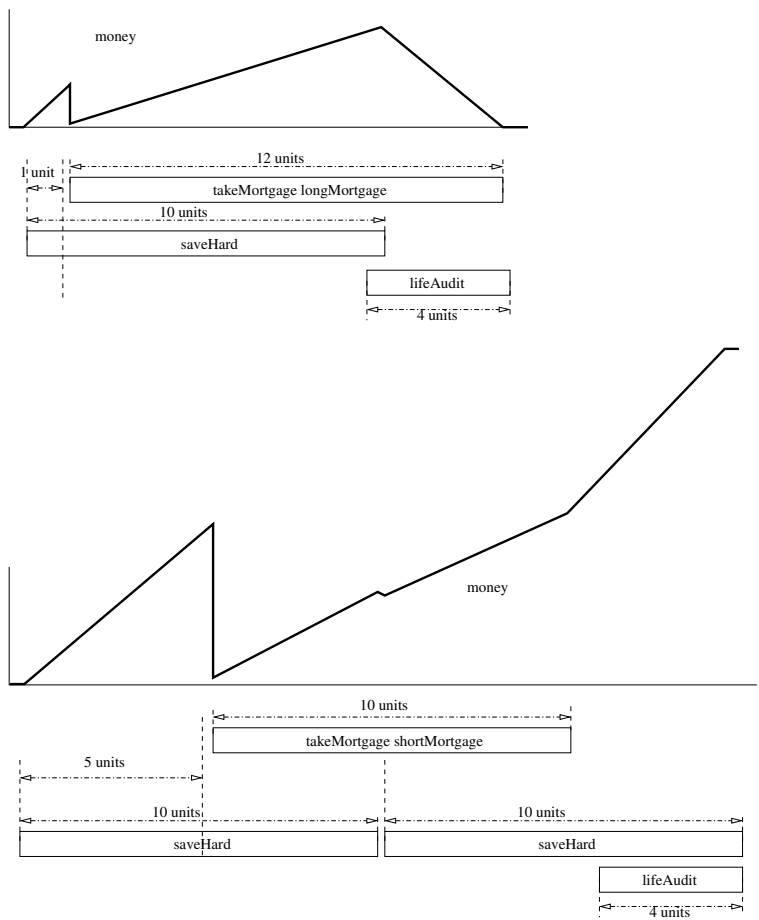

Figure 3: Possible solutions to the Borrower problem.

## 5. Background in Metric and Temporal Planning

Most recent work on discrete numeric planning is built on the ideas introduced in the planner Metric-FF (Hoffmann, 2003). A discrete numeric planning problem introduces numeric variables into the planning domain that can hold any real numeric value (or be undefined, if they have not yet been given a value). Actions can have conditions expressed in terms of these variables, and have effects that act upon them. To provide heuristic guidance, Metric-FF introduced an extension of the relaxed planning graph (RPG) heuristic (Hoffmann & Nebel, 2001), the Metric RPG heuristic, supporting the computation of a relaxed plan for a problems involving discrete numeric change. As with the propositional RPG heuristic, it performs a forwards-reachability analysis in which the delete effects of actions are relaxed (ignored). For numeric effects, ignoring decrease effects does not always relax the problem, as conditions can require that a variable hold a value *less* than a given constant. Thus, as the reachability analysis extends forwards, upper- and lower- bounds on the values of numeric variables are computed: decrease effects have no effect upon the upper bound and increase effects have no effect upon the lower bound, while assignment effects replace the value of the upper (lower) bound if the incumbent has a lower (greater) value (respectively) than that which would be assigned. Deciding whether a precondition is satisfied in a given layer is performed (optimistically) on the basis of these: for a condition $\mathbf{w} \cdot \mathbf{v} \geq c$[1], then an optimistically high value for $\mathbf{w} \cdot \mathbf{v}$ can be computed by using the upper bound on each fluent $v$ assigned a value in $\mathbf{v}$ if its corresponding weight in $\mathbf{w}$ is positive, or, otherwise, using its lower bound.

An alternative to the use of a Metric RPG is proposed in LPRPG (Coles, Fox, Long et al., 2008b), where a linear program is constructed incrementally to capture the interactions between actions. This approach is restricted to actions with linear effects, so is not as general as Metric-FF, but it provides a more accurate heuristic guidance in handling metric problems and can perform significantly better in problems where metric resources must be exchanged for one another in order to complete a solution.

Numeric planning also gives the opportunity to define metric optimisation functions in terms of metric variables within the problem description. For example, an objective to minimise fuel consumption can be defined for domains where the quantity of fuel available is a metric variable. This optimisation function can also include the special variable `total-time`, representing the makespan (execution duration) of the plan. Most planners are restricted to a weighted sum across variables (although PDDL2.1 syntax allows it to be an unrestricted expression across variables). In general, planners are not yet capable of optimising metric functions effectively: the task of finding any plan remains difficult. However, there are some planners that attempt to optimise these functions, the most notable being LPG (Gerevini & Serina, 2000) (and, in domains where the only numeric effects are to count action cost, LAMA, due to Richter & Westphal, 2010).

Although the introduction of PDDL2.1 led to an increased interest in temporal planning, earlier work on planning with time has been influential. IxTeT (Ghallab & Laruelle, 1994) introduced *chronicles*, consisting of temporal assertions and constraints over a set of state variables, and *time-lines* which are chronicles for single state variables. Timelines have since been widely used by planners that have followed a different trajectory of development than that led by the PDDL family of languages (Pell, Gat, Keesing, Muscettola, & Smith, 1997; Frank & Jónsson, 2003; Cesta, Cortellessa, Fratini, & Oddi, 2009). IxTeT also pioneered the use of many important techniques, including simple temporal networks and linear constraints.

The language introduced for the planner 'Temporal Graph Plan' (TGP) (Smith & Weld, 1999) allowed (constant) durations to be attached to actions. The semantics of these actions required their preconditions, *pre*, to be true for the entire duration of the action, and the effects of the actions, *eff*, to become available instantaneously at their ends. The values of affected variables are treated as undefined and inaccessible during execution, although the intended semantics (at least in TGP) is that the values should be considered *unobservable* during these intervals and, therefore, plans should be conformant with respect to all possible values of these variables over these intervals. TGP solves these problems using a temporally extended version of the Graphplan planning graph (Blum & Furst, 1995) to reason with temporal constraints. A temporal heuristic effective for this form of temporal planning was developed by Haslum and Geffner (2001) and Vidal and Geffner (2006) have explored a constraint propagation approach to handling these problems.

Even when using the more expressive temporal model defined in PDDL2.1, many temporal planners make use of the restricted TGP semantics, exploiting a simplification of the PDDL2.1 encoding known as 'action compression'. The compression is performed by setting *pre* to be the weakest preconditions of the actions, and *eff*$^+$ (*eff*$^-$) to be their strongest add (delete) effects. In the propo-

---

1. Conditions $\mathbf{w} \cdot \mathbf{v} \leq c$ can be rewritten in this form by negating both sides. Further, those stating $\mathbf{w} \cdot \mathbf{v} = c$ can be rewritten as a pair of conditions, $\mathbf{w} \cdot \mathbf{v} \geq c$ and $-(\mathbf{w} \cdot \mathbf{v}) \geq -c$

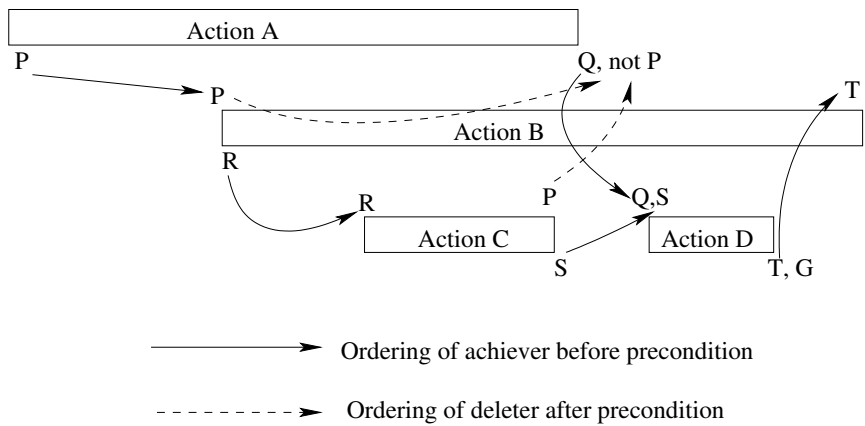

Figure 4: A problem for SAPA.

sitional case, in terms of the action representation introduced earlier, these are:

$$pre = pre_{\vdash} \cup ((pre_{\leftrightarrow} \cup pre_{\dashv}) \setminus eff^+_{\vdash})$$

$$eff^+ = (eff^+_{\vdash} \setminus eff^-_{\dashv}) \cup eff^+_{\dashv}$$

$$eff^- = ((eff^-_{\vdash} \setminus eff^+_{\vdash}) \cup eff^-_{\dashv}) \setminus eff^+_{\dashv}$$

Many modern temporal planners, such as MIPS-XXL (Edelkamp & Jabbar, 2006) and earlier versions of LPG (Gerevini & Serina, 2000), make use of this action compression technique. However, applying the compression can lead to incompleteness (Coles, Fox, Halsey, Long, & Smith, 2008) (in particular, a failure to solve certain temporal problems). The issues surrounding incompleteness were first discussed with reference to the planner CRIKEY (Fox, Long, & Halsey, 2004) and, later, the problem structures causing this were said to introduce *required concurrency* (Cushing, Kambhampati, Mausam, & Weld, 2007). The Borrower domain is one example of a problem in which the compression prevents solution. Both the `lifeAudit` and `takeMortgage` actions have initial preconditions that can only be satisfied inside the interval of the `saveHard` action, since this action adds `saving` at its start, but deletes it at its end.

Required concurrency is a critical ingredient in planning with continuous effects, as both *when* change occurs and *what* change occurs are important throughout the execution of actions. In order to avoid producing poor quality plans or, indeed, excluding possible solutions, we must allow concurrency between actions wherever the problem description permits it. A naïve extension of the compression approach would discretise continuous numeric change into step function effects occurring at the ends of the relevant actions, precluding any possibility of managing the interaction between numeric variables during execution of actions with continuous effects. We therefore build our approach on a planner capable of reasoning with required concurrency. In the Borrower domain, the mortgage action must overlap with the saving action, but it cannot be too early (to meet the deposit requirement) or too late (to meet the maximum savings constraint and to ensure that the life audit can be performed as early as possible). As this example illustrates, problems that include reasoning with continuous linear change typically also require concurrency.

Several planners are, currently, capable of reasoning with the PDDL2.1 start–end semantics, as opposed to relying on a compression approach. The earliest PDDL2.1 planner that reasons successfully with the semantics is VHPOP (Younes & Simmons, 2003), which is a partial-order planner.

This planner depends on heuristic guidance based on the same relaxed planning graph that is used in FF, so the guidance can fail in problems with required concurrency. Nevertheless, the search space explored by VHPOP includes the interleavings of action start and end points that allow solution of problems with required concurrency. VHPOP suffers from some of the problems encountered in earlier partial-order planners and its performance scales poorly in many domains. TPSYS (Garrido, Fox, & Long, 2002; Garrido, Onainda, & Barber, 2001) is a Graphplan-inspired planner that can produce plans in domains with required concurrency. Time is represented by successive layers of the graph, using a uniform time increment for successive layers. This approach is similar to the way that TGP uses a plan graph to represent temporal structure, but TPSYS supports a model of actions that separates the start and end effects of actions as dictated by PDDL2.1 semantics.

Another planner that adopts a Graphplan-based approach to temporal planning is LPGP (Long & Fox, 2003a), but in its case the time between successive layers is variable. Instead of using layers of the graph to represent the passage of fixed-duration increments of time, they are used to represent successive *happenings* — time points at which state changes occur. The time between successive state changes is allowed to vary within constraints imposed by the action durations whose end points are fixed at particular happenings. A linear program is constructed, incrementally, to model the constraints and the solution of the program is interleaved with the selection of action choices. This approach suffers from most of the weaknesses of a Graphplan planner: the exhaustive iterative deepening search is impractical for large problems, while computation and storage of mutex relations becomes very expensive in larger problems. Nevertheless, LPGP provides a useful approach to the treatment of PDDL2.1 durative actions, by splitting them into their end points which are treated as instantaneous 'snap' actions. A solution to the (original) planning problem can be expressed in terms of these, subject to four conditions:

1. Each start snap-action is paired with an end snap-action (and no end can be applied without its corresponding start having been applied earlier);

2. Between the start and end of an action, the invariants of the action $pre_{\leftrightarrow}$ are respected;

3. No actions must be currently executing for a state to be considered to be a goal state;

4. Each step in the plan occurs after the preceding step, and the time between the start and end of an action respect its duration constraints.

SAPA (Do & Kambhampati, 2003) is one of the earliest forward-search planners to solve temporal PDDL2.1 problems. It works with a priority queue of events. When a durative action is started its end point is queued at the time in the future at which it will be executed. The choice points of the planner include starting any new action, but also a special *wait* action, which advances time to the next entry in the queue, and the corresponding action end point is executed. This allows SAPA to reason with concurrency and to solve some problems with required concurrency. Unfortunately, its search space does not include all necessary interleavings to achieve a complete search. For example, consider the problem illustrated in Figure 4. To solve this problem, action $A$ must start, then action $B$ must start early enough to allow $C$ to complete before $A$ ends (and deletes $P$) and late enough that action $D$ can start before $B$ ends but end after $A$ ends. All of the actions are required in order to allow $D$ to be applied, achieving the goal $G$. After SAPA starts action $A$, the queue will contain the end of $A$. The choices now open are to start $B$ immediately, but this will then end too early to allow $D$ to execute successfully, or else to complete $A$, which advances time too far to allow $B$ to exploit effect $P$ of $A$, preventing $C$ from being executed. In fact, a simpler problem defeats SAPA: if $B$ were to have end condition $Q$ instead of $T$ and end effect $G$ then $C$ and $D$ can be dispensed with. However, the additional complexity of the existing example is that it is impossible to infer when to start $B$ by examination of $A$ and $B$ alone, because the timing constraints on the start of $B$ depends on both actions $C$ and $D$ and it is not immediately obvious how their temporal constraints will affect the placement of $B$. The difficulty in adopting the waiting approach is that it is hard to anticipate how long to wait if the next interesting time point depends on the interaction of actions that have not yet even been selected.

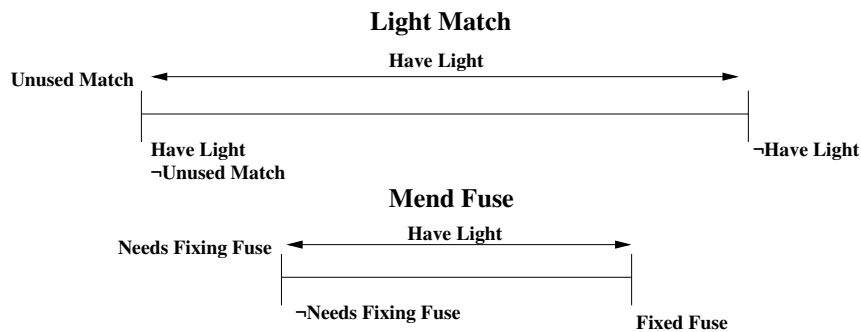

Figure 5: Required Concurrency

A different approach to forward-search temporal planning is explored in the CRIKEY family of planners (Coles, Fox, Halsey et al., 2008; Coles, Fox, Long et al., 2008a). These planners use the same action splitting approach used in LPGP, but work with a heuristically guided forward search. The heuristics in these planners use a relaxed planning graph as a starting point (Hoffmann & Nebel, 2001), but extend it by adding some guidance about the temporal structure of the plan, pruning choices that can be easily demonstrated to violate temporal constraints and inferring choices where temporal constraints imply them. The planners use a Simple Temporal Network to model and solve the temporal constraints between the action end points as they are accumulated during successive action choices. Split actions have also been used to extend LPG into a temporal version that respects the semantics of PDDL2.1 (Gerevini, Saetti, & Serina, 2010) (earlier versions of LPG use the compressed action models described above). Recent work by Haslum (2009) has explored other ways in which heuristics for temporal planning can be constructed, while remaining admissible.

Temporal Fast Downward (Eyerich et al., 2009), based on Helmert's Fast Downward planner (Helmert, 2006), uses an approach that is a slight refinement of the compressed action model, allowing some required concurrency to be managed. The authors demonstrate that this planner can solve the Match problem shown in Figure 5. They mistakenly claim that SAPA cannot solve this problem because it cannot consider applying an action between starting and ending lighting the match: in fact, SAPA can apply the `mend_fuse` action after the match is lit, in much the same way as is done in Temporal Fast Downward. The problem that both planners face is in situations in which an action must be started some time after the last happening, but before the next queued event: neither planner includes this choice in its search space.

Huang et al. (2009) developed a temporal planner exploiting the planning-as-SATisfiability paradigm. This uses a Graphplan-to-SAT encoding, starting with an LPGP action-splitting compilation, and using a fixed time increment between successive layers of the graph. This approach is adequate for problems where an appropriate time increment can be identified, but this is not possible, in general, when there are time-dependent effects in a domain. Furthermore, the approach is ineffective when there is significant difference between the durations of actions, so that the time increment becomes very short relative to some actions. The planner can produce optimal (makespan) plans using iterative deepening search. The planner combines existing ideas to achieve its objectives and it is mainly of interest because of its relationship to other SAT-based approaches to temporal planning, such as TM-LPSAT discussed below.

CRIKEY3, and the other planners mentioned, are only capable of solving the simple temporal planning problems described above. They are restricted to the management of discrete change. Duration-dependent change cannot be handled by these planners. In fact, not all of these planners can manage any kind of reasoning with numbers outside the durations of actions. COLIN therefore significantly extends the competence of other PDDL-compliant temporal planners.

## 6. CRIKEY3: A Forward-Chaining Temporal Planner

Temporal forward-chaining planners have two kinds of choices to make during the construction of plans. Firstly, as in the non-temporal case, a choice must be made of which actions to apply (these choices can be considered to be the 'planning' element of the problem). Secondly, choices must be made of *when* to apply the actions (these can be seen as the 'scheduling' choices in construction of solutions). CRIKEY3 (Coles, Fox, Long et al., 2008a), a temporal forward-chaining planner, exploits the distinction between these choices, using separate procedures to make the planning decisions (which actions to start or end) and the scheduling decisions (when to place actions on the timeline). Both of these decisions must be checked for consistency with respect to the existing temporal constraints to confirm that all the actions can be completely scheduled. In this section, we briefly describe how CRIKEY3 performs planning and scheduling, since its architecture forms the basis for COLIN and the work subsequently described in this paper. Full details of temporal management in CRIKEY3 are provided by Coles et al.

CRIKEY3 uses a forward-chaining heuristic state-space search to drive its planning decisions. It makes use of the Enforced Hill-Climbing (EHC) algorithm introduced in FF (Hoffmann & Nebel, 2001) and repeated, for convenience, as Algorithm 1. EHC is incomplete, so if a solution cannot be found CRIKEY3 plans again, using a weighted A* search. We now discuss how the search described within the basic enforced hill-climbing algorithm of FF can be extended to perform temporal planning. In order to do this, a number of modifications are required. In particular:

1. $get\_applicable\_actions(S)$: the planner must reason with two actions per durative action, a start action and an end action, rather than applying an action and immediately considering it to have finished (as in the non-temporal case).

2. $get\_applicable\_actions(S)$, $apply(a, S)$: invariant conditions of durative actions must be maintained throughout their execution, which requires active invariants to be recorded in the state in order to prevent the application of actions that conflict with them.

3. $is\_goal\_state(S)$: for a state to be a goal state (i.e. for the path to it to be a solution plan) all actions must have completed.

---

**Algorithm 1**: Enforced Hill-Climbing Algorithm

---

   **Data**: $P = \langle A, I, G \rangle$ - a planning problem

   **Result**: $P$, a solution plan

**1**   $best\_heuristic \leftarrow evaluate\_heuristic(I)$;

**2**   **if** $best\_heuristic = 0$ **then**

**3**      **return** [];

**4**   $closed \leftarrow \{I\}$;

**5**   $open\_list \leftarrow [\langle I, [] \rangle]$;

**6**   **while** $open\_list \neq []$ **do**

**7**      $\langle S, P \rangle \leftarrow$ element removed from the front of $open\_list$;

**8**      $applic(S) \leftarrow get\_applicable\_actions(S)$;

**9**      $apply\_helpful\_filter(applic(S))$;

**10**     **foreach** $a \in applic(S)$ **do**

**11**        $S' \leftarrow apply(a, S)$;

**12**        **if** $S' \notin closed$ **then**

**13**          add $S'$ to $closed$;

**14**          $P' \leftarrow P$ followed by $a$;

**15**          **if** $is\_valid\_plan(P')$ **then**

**16**            **if** $is\_goal\_state(S')$ **then**

**17**              **return** $P'$;

**18**            $h \leftarrow evaluate\_heuristic(S')$;

**19**            **if** $h < best\_heuristic$ **then**

**20**              $open\_list \leftarrow [\langle S', P' \rangle]$;

**21**              $best\_heuristic \leftarrow h$;

**22**              **break**;

**23**            **else**

**24**              **if** $h < \infty$ **then**

**25**                append $\langle S', P' \rangle$ onto $open\_list$;

**26**   $return\_with\_failure$;

---

4. $is\_valid\_plan(P)$: the temporal (scheduling) constraints of candidate plans must be respected. In particular, the duration constraints of durative actions must be satisfied. This is discussed in Section 6.1.

We consider each of these modifications in turn. First, durative actions are compiled into two non-temporal actions. A modified version of the LPGP action compilation (Long & Fox, 2003a) is used for this, as described by Coles et al. (2008). Each durative action $a$, of the form $\langle dur, pre_\vdash, eff_\vdash, pre_\leftrightarrow, pre_\dashv, eff_\dashv \rangle$, is split into two non-temporal (in fact, instantaneous) 'snap actions' of the form $\langle pre, eff \rangle$:

- $a_\vdash = \langle pre_\vdash, eff_\vdash \rangle$

- $a_\dashv = \langle pre_\dashv, eff_\dashv \rangle$

By performing search with these snap actions, and taking appropriate care to ensure that the other constraints are satisfied, the restrictions on expressivity imposed by the use of action compression are avoided. It becomes possible to search for a plan in which the start and end points of different actions are coordinated, solving problems with required concurrency. The price for this is that the search space is much larger: each original action is replaced by two snap-actions, so the length of solution plans is doubled. In some circumstances this blow-up can be avoided by identifying actions that are *compression safe* (Coles, Coles, Fox, & Long, 2009a), i.e. those for which the use of action compression does not compromise soundness or completeness. In the approach described by Coles et al., these actions are still split into start and end snap-actions, but the end points of compression-safe actions are inserted when either their effects are needed or their invariants would otherwise be violated by another action chosen for application. As a consequence, only one search decision point is needed per compression-safe action (choose to apply its start), rather than two. Recent versions of both CRIKEY3 and COLIN make use of this restricted action compression technique in search.

Having split actions into start and end points, modifications to the basic search algorithm are needed to handle the constraints that arise as a consequence. CRIKEY3 makes use of an extended state representation, adding two further elements to the state tuple. The resulting state is defined as $S = \langle F, P, E, T \rangle$, where:

- $F$ represents the facts that hold in the current world state: a set of propositions that are currently true, $W$, and a vector, $\mathbf{v}$, recording the values of numeric variables.

- $P$ is an ordered list of snap actions, representing the plan to reach $S$ from the initial state.

- $E$ is an ordered list of start events, recording actions that have started but not yet finished;

- $T$ is a collection of temporal constraints over the actions in the plan to reach $F$.

The purpose of the start event list $E$ is to record information about the currently executing actions, to assist in the formation of sound plans. Each entry $e \in E$ is a tuple $\langle op, i, dmin, dmax \rangle$ where:

- $op$ is the identifier of an action, for which the start snap-action $op_{\vdash}$ has been added to the plan;

- $i$ is the index at which this snap-action was added in the plan to reach $S$;

- $dmin, dmax$ are the minimum and maximum duration of $op$, determined in the state in which $op$ is started.

The minimum and maximum duration of an action can depend on the state in which it is applied (e.g. the duration of a recharge action may depend on the level of charge at the time of execution), so durations must be computed based on the state preceding step $i$. However, once a given action has started, the bounds on the duration remain fixed. PDDL2.1 also allows actions to have durations constrained by conditions that hold at the end of the action, but such actions are not supported by our planners.

This extended state definition leads to corresponding extensions to $get\_applicable\_actions(S)$. As before, a snap-action is deemed to be *logically applicable* in a state $S$ if its preconditions *pre* are satisfied in $S$. However, an additional condition must be satisfied: its effects must not violate any active invariants. The invariants active in a given state are determined from $E$ — we denote the invariants in a state $S$ with event list $E$ as:

$$inv(S) = \underset{e \in E}{\cup} e.op.pre_{\leftrightarrow}$$

To apply the end snap-action, $a_{\dashv}$, there is required to be an entry $e \in E$ whose operator entry $op$ is equal to $a$. This prevents the planner from attempting to apply the ends of actions that have not yet been started.

Assuming an action, $a$, is found to be applicable and chosen as step $i$ of a plan, the function $apply(a, S)$, applied to a temporally-extended state, $S$, yields a successor $S' = \langle F', P', E', T' \rangle$. The first two elements are updated as in the non-temporal case: $F' = apply(a, F)$, and $P' = P + [a]$. To obtain $T'$, we begin by setting $T' = T$. Furthermore, if $i > 0$:

$$T' = T' \cup \{\epsilon \leq t(i) - t(i-1)\}$$

where $t(i)$ is the variable representing the time at which step $i$ is scheduled to be executed. That is, the new step must come at least $\epsilon$ (a small unit of time) after the preceding step. This separation respects the requirement that interfering actions must be separated by at least $\epsilon$ (Fox & Long, 2003), but it is strictly stronger than required where actions are not actually mutually exclusive. A more accurate realisation of the PDDL2.1 semantics could be implemented, but it would incur a cost while offering very little apparent benefit. Finally, the resulting value of $E'$ (and whether $T'$ is changed further) depends on whether $a$ is a start or end snap-action:

- if a start action $a_{\vdash}$ is applied, $E' = E + [\langle a, i, dmin, dmax \rangle]$, where $dmin$ and $dmax$ correspond to the lower- and upper-bounds of the duration of $a$, as evaluated in the context of valuation $F$.

- if an end action $a_{\dashv}$ is applied, a start entry $\{e \in E \mid e.op = a\}$ is chosen, and then $E'$ is assigned a value $E' = E \setminus e$. It will often be the case that there is only one instance of an action open, so there is only one choice of pairing, but in the case where multiple instances of the same action are executing concurrently, search branches over the choice of each such $e$. For the $e$ chosen, a final modification is then made to $T'$ to encode the duration constraints of the action that has just finished:

$$T' = T' \cup \{e.dmin \leq t(i) - t(e.i) \leq e.dmax\}$$

With this information encoded in each state about currently executing actions, the extension needed to $is\_goal\_state(S)$ is minor: a state $S$ is a goal state if it satisfies the non-temporal version of $is\_goal\_state(S)$, and if the event list of the state, $E$, is empty.

This search strategy leads to a natural way to handle PDDL2.2 Timed Initial Literals (TILs) directly. Dummy 'TIL actions' are introduced, comprising the effects of the TILs at each time point, and these can be added to the plan if all earlier TIL actions have already been added, and if they do not delete the invariants of any open action. As a special case, TIL actions do not create an entry in $E$: only the facts in $F$ are amended by their execution. They do, however, produce an updated set of temporal constraints. As with snap actions, if a TIL is added as step $i$ to a plan, the TIL must fall no earlier than $\epsilon$ after the preceding step. Then, $T' = T' \cup \{ts \leq t(i) - t(\alpha) \leq ts\}$, where $ts$ is the time-stamp at which the TIL is prescribed to happen, $\alpha$ is the name denoting the start of the plan and $t(\alpha) = 0$. As can be seen, these constraints ensure that the TIL can only occur at an appropriate time, that any step prior to the TIL must occur before it, and that any step after the TIL must occur after it.

The changes described in this subsection ensure that the plans produced by CRIKEY3 are logically sound: the check for logical applicability, coupled with the maintenance of $E$ throughout search, ensures that no preconditions, either propositional or numeric, can be broken. Use of $get\_applicable\_actions(S)$ only guarantees that actions are logically applicable: there is no guarantee that adding a snap-action to the plan, judged applicable in this way, will not violate the temporal constraints. For example, it is possible that all preconditions are satisfied in the plan $P = [a_\vdash, b_\vdash, b_\dashv, a_\dashv]$, so that $P$ is *logically* sound. However, if the duration of $b$ is greater than the duration of $a$ then $P$ is not *temporally* sound. In the next section we discuss how the function $is\_valid\_plan(P)$ is modified to identify and reject temporally inconsistent plans.

## 6.1 Temporal Plan Consistency

A state $S$ is only temporally consistent if the steps $[0...n-1]$ in the plan, $P$, that reaches it can be assigned values $[t(0)...t(n-1)]$, representing the times of execution of each of the corresponding steps, respecting the temporal constraints, $T$. This is checked through the use of $is\_valid\_plan(P')$, called at line 15 of Algorithm 1 — this function call is trivial in the non-temporal case, but in the temporal case serves to check the temporal consistency of the plan. Any state for which the temporal constraints cannot be satisfied is immediately pruned from search, since no extension of the action sequence can lead to a solution plan that is valid.

The temporal constraints $T$ built by CRIKEY3 in a state $S$ are each expressed in the form:

$$lb \leq t(b) - t(a) \leq ub \qquad \text{where } lb, ub \in \Re \text{ and } 0 \leq lb \leq ub$$

These constraints are conveniently expressible as a Simple Temporal Problem (STP) (Dechter, Meiri, & Pearl, 1989). The variables within the STP consist of the timestamps of actions, and between them inequality constraints can be specified in the above form. Crucially, for our purposes, the validity of an STP (and the assignment of timestamps to the events therein) can be determined in polynomial time by solving a shortest-path problem within a Simple Temporal Network (STN), a directed-graph representation of an STP. Each event in the STP is represented by a vertex in the STN. There is an additional node $t(\alpha)$ to represent time 0 and the time of the first action in the plan, $t(0)$, is constrained to fall within $\epsilon$ of $t(\alpha)$. Each constraint in the above form adds two edges to the graph: one from $a$ to $b$ with weight $ub$, and one from $b$ to $a$ with weight $-lb$. Attempting to solve the shortest-path problem from $t(\alpha)$ to each event yields one of two outcomes: either it terminates successfully, providing a time-stamp for each step, or it terminates unsuccessfully due to the presence of a negative-cost cycle within the STN indicating a temporal inconsistency (any schedule would require at least one step to be scheduled before itself).

In CRIKEY3, an STP is used to check the temporal consistency of the choices made to reach each step $S$, based on the temporal constraints $T$ that must hold over the plan $P$ to reach $S$, and additional constraints that can be determined from $E$: the list of actions that have started, but not yet finished. The variables $vars$ in the STP can be partitioned into two sets: the '$t$' variables, $t(i)$ for step $i \in P$ and the '$f$' variables, one $f(i)$ for each entry $\langle op, i, dmin, dmax \rangle \in E$. The $t$ variables correspond to the times of steps that have already been added to the plan, which might be the times of start or end points of actions. Some of these time points might correspond to the starts of actions that have not yet finished and it is this subset of actions (only) that will have associated $f$ variables associated with the pending end times of those actions. For consistency with the terminology we introduced in CRIKEY3 (Coles, Fox, Long et al., 2008a), we use *now* to refer to the time at which the next event in the plan will occur (which could be during the execution of the last actions applied). It is the time point at which the next choice is to be made, either the start of a new action or the completion of an existing one, and can therefore be seen as the time associated with the final state, $S$, generated by the current plan head. There is only ever one timepoint called *now* and its value moves forward as the plan head extends. The constraints are then as follows:

- $T$, constraining the $t$ variables — these ensure the temporal consistency of the steps in the plan to reach $S$ (and include any constraints introduced for timed initial literals);

- $\{dmin \leq f(i) - t(i) \leq dmax \mid \langle op, i, dmin, dmax \rangle \in E\}$ — that is, for each future action end point that has been committed to (but has yet to be applied), the recorded duration constraint must be respected;

- $\{\epsilon \leq f(i) - t(n-1) \mid \langle op, i, dmin, dmax \rangle \in E\}$ — that is, each future action end point must come after the last step in the current plan, to ensure it is in the future.

- $t(now) - t(n-1) \geq \epsilon$ — that is, the current time (the time at which the next event in the plan can occur) is at least $\epsilon$ after the last event in the plan.

Solving this STP confirms the temporal consistency of the decisions made so far. If the STP cannot be solved, the state $S$ can be pruned: the plan induced from the start–end action representation is temporally invalid. The last two of these categories of constraints are particularly important: without them, pruning could only be undertaken on the basis of the plan $P$ to reach $S$. Including them, however, allows the STP to identify cases where the end point of an action can never be added to the plan, as doing so would lead to temporal inconsistency. As goal states cannot contain any executing actions (i.e. $E$ must be empty), this allows CRIKEY3 to prune states earlier from which there can definitely be no path to a state in which all end points have been added to the plan.

Timed initial literals are easily managed in the STP using the dummy TIL actions described earlier. The constraints for each dummy TIL action that has already been applied are included in $T$. Each dummy TIL action yet to occur is automatically treated as the end of an action that has yet to be applied. Thus, an $f$ variable is added for each, and in doing so, the last step in the plan so far is constrained to come before each TIL event that has yet to happen.

## 7. Planning with Continuous Numeric Change

The most challenging variants of temporal and numeric problems combine the two to arrive at problems with time-dependent metric fluents. Although problems exhibiting hybrid discrete-continuous dynamics have been studied in other research communities for some time, for example, in verification (Yi, Larsen, & Pettersson, 1997; Henzinger, Ho, & Wong-Toi, 1995; Henzinger, 1996), where timed automata capture exactly this kind of behaviour, there has been relatively little work on continuous dynamics in the planning community.

In PDDL2.1 the model of mixed discrete-continuous change extends the propositional state transition model to include continuous change on the state variables. There is a state transition system in which discrete changes transition instantaneously between states. While the system is in a particular state, continuous change can occur on the state variables and time passes. As soon as a discrete change occurs the system changes state. In PDDL+ (Fox & Long, 2006) this is extended to allow exogenous events and processes (controlled by nature) as well as durative actions. This leads to a formal semantics that is based in the theory of Hybrid Automata (Henzinger, 1996). An action causes a discrete state change which might trigger a continuous process. This continues over time until an event is triggered leading into a new state. Some time later another action might be taken.

Early work exploring planning with continuous processes includes the Zeno system of Penberthy and Weld (1994), in which processes are described using differential equations. Zeno suffers from the same limitations as other partial order planners of its time, being unable to solve large planning problems without significant aid from a carefully crafted heuristic function. More importantly, a fundamental constraint on its behaviour is that it does not allow concurrent actions to apply continuous effects to the same variable. This imposes a very significant restriction on the kinds of problems that can be solved, making Zeno much less expressive than COLIN. This constraint follows, in part, from the way that the model requires effects to be specified as differential equations, rather than as continuous update effects, so that simultaneous equations must be consistent with one another rather than accumulating additive effects. As the authors say "We must specify the entire continuous behaviour over the interval [of the durative action] as our semantics insist that all continuous behaviours are the result of direct, explicit action".

Another early planner to handle continuous processes is McDermott's OPTOP system (McDermott, 2003), which is a heuristic search planner, using a regression-based heuristic. The 'plausible progression' technique used within OPTOP to guide search is not sufficiently powerful to recognise interactions that could prevent future application of actions, thereby restricting its scalability on problems of the form we consider here. OPTOP competed in the International Planning Competition in 2004, where it solved only a small subset of the problems (although, interestingly, those it solved involved an expressive combination of ADL and temporal windows that no other planner could manage). OPTOP is an interesting variant on the heuristic forward search approach, since it avoids grounding the representation, using an approach that is similar to a means-ends linear planning approach to generate relaxed plan estimates of the number of actions required to achieve the goal from a given state.

## 7.1 TM-LPSAT

More recently, Shin and Davis developed TM-LPSAT (Shin & Davis, 2005), based on the earlier LPSAT system (Wolfman & Weld, 1999). TM-LPSAT was the first planner to implement the PDDL+ semantics. It is implemented as a compilation scheme by which a horizon-bounded continuous planning problem is compiled into a collection of SAT formulas that enforce the PDDL+ semantics, together with an associated set of linear metric constraints over numeric variables. This compiled formulation is then passed to a SAT-based arithmetic constraint solver, LPSAT. LPSAT consists of a DPLL solver and an LP solver. The SAT-solver passes triggered constraints to the LP-solver, which hands back conflict sets in the form of nogoods if the constraints cannot be resolved. If there is no solution the horizon is increased and the process repeats, otherwise the solution is decoded into a plan. In order to support concurrency the compilation exploits the LPGP separation of action start and end points. There are different versions of TM-LPSAT exploiting different solvers: LPSAT and MathSAT-04 (Audemard, Bertoli, Cimatti, Kornilowicz, & Sebastiani, 2002) have both been exploited. The novelty of TM-LPSAT lies in the compilation and decoding phases, since both solvers are well-established systems.

The compilation scheme of TM-LPSAT implements the full PDDL+ semantics. Although this includes events and processes, which are specific to PDDL+, TM-LPSAT can also handle variable duration durative actions, durative actions with continuous effects and duration-dependent end-effects. The continuous effects of concurrent actions on a quantity between two time-points are summed over all actions active on the quantity over the period. Therefore, TM-LPSAT supports concurrent updates to continuous variables.

TM-LPSAT is an interesting approach, in theory capable of solving a large class of problems with varied continuous dynamics. However, reported empirical data suggests that the planner is very slow and unable to solve problems requiring plans of more than a few steps. It is not possible to experiment further because there is no publicly available implementation of the system.

## 7.2 Kongming

Hui Li and Brian Williams have explored planning for hybrid systems (Li & Williams, 2008, 2011). This work has focussed on model-based control, using techniques based on constraint reasoning. The continuous dynamics of a system are modelled as *flow tubes* that capture the envelopes of the continuous behaviours (Léauté & Williams, 2005). The dimensions of these tubes are a function of time (typically expanding as they are allowed to extend), with the requirement being made that successive continuous behaviours must be connected by connecting the start of one tube (the precondition surface) to the cross-section of the preceding tube; i.e. the intersection of the two spaces must be non-empty. The most relevant work in this area is in the development of the planner Kongming, described by Li and Williams.

Kongming solves a class of control planning problems with continuous dynamics. It is based on the construction of fact and action layers and flow tubes, within the iterative plan graph structure introduced in Graphplan (Blum & Furst, 1995). As the graph is developed, every action produces a flow tube which contains the valid trajectories as they develop over time. Starting in a feasible region, actions whose preconditions intersect with the feasible region can be applied and the reachable states at any time point can be computed using the state equations of the system. In the initial state of the system all the variables have single known values. A valid trajectory must pass through a sequence of flow tubes, but must also meet the constraints specified in the dynamics of the actions selected. The mutex relation used in Graphplan is extended to the continuous dynamics as well as the propositional fragment of the language. The graph is iteratively extended as in Graphplan, with a search for a plan conducted after each successive extension.

The plan-graph encoding of a problem with continuous dynamics is translated into a Mixed Logical-Quadratic Program (MLQP). The metric objective functions used by the planner to optimise its behaviour can be defined in terms of quadratic functions of state variables. An example problem considered by Li and Williams (2008) is a 2-d representation of a simple autonomous underwater vehicle (AUV) problem where the AUV can glide, ascend and descend while avoiding obstacles. The language used is a version of PDDL2.1 extended to enable dynamics to be encoded. The continuous nature of the problem lies in the fact that, after a continuous action, the AUV will be in one of a continuous range of positions determined by the control system. Because Kongming depends on translation of the planning problems into MLQPs the constraints describing the dynamics of the problem must be linear. Since the effects of continuous actions involve the product of rate of change with time, only one of these values can be treated as a variable. In Kongming it is the rate of change that is variable, but time is discretised, which contrasts with COLIN in which rates of change remain constant over continuously variable length intervals. The discretisation of time in Kongming is exploited to support state updates within the plan graph: successive layers of the graph are separated by a constant and uniform time increment. This approach suffers from a disadvantage that the duration of a plan is limited by the number of happenings in the plan, since the solver cannot realistically solve problems with more than a few tens of layers in the plan graph.

Kongming does not support concurrent continuous updates to the same state variable, so, in this respect, PDDL2.1 is more expressive than the extended language used in Kongming. In part this is due to a difficulty in resolving precisely what is the semantics of the dynamics described in the actions used by Kongming. Each dynamic constraint specifies limits on the rate of change of a specific variable: it is unclear whether concurrent actions should be combined by taking the union or the intersection of the bounds each constraint specifies on the rate of change of a given fluent.

### 7.3 UPMurphi

One other recently developed planner that uses PDDL2.1 and reasons with continuous processes is UPMurphi (Penna, Intrigila, Magazzeni, & Mercorio, 2009). UPMurphi takes a completely different approach to those considered so far. Instead of reasoning about continuous change directly, UPMurphi works by guessing a discretisation and iteratively refining it if the solution to the discretised problem does not validate against the original problem specification. The iterative driver is the coarseness of the discretisation, as well as the planning horizon, making it an interestingly different basic architecture from TM-LPSAT.

UPMurphi begins with the continuous representation of the problem and starts by discretising it. First the actions are discretised by taking specific values from their feasible ranges. This results in several versions of each action. Then UPMurphi explores the state space, by explicitly constructing it under the current discretisation. Plans are constructed using the planning-as-model-checking paradigm (Cimatti, Giunchiglia, Giunchiglia, & Traverso, 1997): there is no heuristic to guide search. Once a plan has been found it is then validated against the original continuous model, using the plan validator (Fox, Howey, & Long, 2005). If it is invalid, the discretisation is refined and the search resumes. If UPMurphi fails to find a plan at one discretisation it starts again at a finer grained discretisation. Subsequent refinements lead to ever denser feasible regions, but they are increasingly complex to construct.

UPMurphi can be used to build partial policies to handle the uncertainty that is likely to arise in practice during the execution of hybrid control plans. A controller table is initially synthesised, consisting of the (state,action) pairs of the plan it first constructs. However, this table might lack some of the states that could be visited by the controller, so it is not robust. The subsequent step is to "robustify" the controller by randomly perturbing some of the states and finding new paths from these new states. Because some of the perturbed states are not reachable, a probability distribution is used to identify the most likely ones. These are called the safe states. The controller table is then extended with the safe (state, action) pairs. The controller table, or policy, is referred to as a Universal Plan.

### 7.4 Other Approaches to Continuous Reasoning

A completely different way to manage continuous quantities is to model continuous resource consumption and production in terms of uncertainty about the amount consumed or produced. This is the approach taken in the HAO* algorithm (Meuleau, Benazera, Brafman, Hansen, & Mausam, 2009) where a Markov Decision Process (MDP) is constructed consisting of hybrid states. Each state contains a set of propositional variables and also a collection of distributions over resource consumption and production values. Because the states are hybrid, standard value iteration approaches cannot be used to find policies. A hybrid AO* approach is described which can be used to find the best feasible policy. The feasible region constructed by HAO* is a continuous distribution of resource values and the resource is considered to be uncontrollable (unlike in Kongming, where it is assumed that the executive maintains control over which values in the region are eventually chosen).

Planning with continuous processes has important applications and, as with many other application areas of planning, this has led to the development of systems that combine generic planning technology with more carefully tuned domain-specific performance to achieve the necessary combination of problem coverage and performance. A good example of this is the work by Boddy and Johnson (2002) and colleagues (Lamba et al., 2003) on planning oil refinery operations. This work uses a quadratic program solver, coupled with heuristically guided assignment to discrete decision variables (corresponding to actions), to solve real problems.

## 8. COLIN: Forward Chaining Planning With Continuous Linear Change

In this section we will describe how CRIKEY3 is extended to reason with duration-dependent and continuous numeric change, building the planner COLIN ( for COntinuous LINear dynamics). We decided to give the planner a specific name to highlight its capabilities. As demonstrated in Section 4.1, the key difference introduced with continuous numeric change is that logical and numeric constraints can no longer be neatly separated from temporal constraints: the values of the numeric variables in a state depend on the timestamps and durations of actions, and vice versa. The relative benefits of handling temporal and numeric constraints together, rather than separating them out, are apparent in the motivating domains outlined in Section 3 and have been amply rehearsed in the paper describing PDDL+ (Fox & Long, 2006).

The need to cope with integrated numeric and temporal constraints raises a number of important issues for planning with these domains. First, checking whether an action choice is consistent can no longer be achieved using an STP, as the numeric constraints now interact with the temporal constraints, and an STP is not sufficiently expressive to capture this. Second, the changing values of numeric variables over time brings new challenges for determining action applicability: if a precondition is not satisfied immediately following the application of an action, it might become satisfied after allowing a certain amount of time to elapse. Finally, there is the need to provide heuristic guidance. We will cover the first two of these issues in this section, and defer discussion of the heuristic guidance to the next.

### 8.1 Temporal-Numeric Plan Consistency Through Linear Programming

We begin with the problem of temporal-numeric plan consistency, as the techniques used in dealing with this issue can also be amended for use in solving the issues encountered when determining action applicability. Considering the definition of the STP given in Section 6.1, we make the observation that the STP could equally well be written as a linear program (LP). In CRIKEY3, the STP is more efficiently solved using a shortest-path algorithm. However, this observation becomes important when we wish to reason with continuous change in numeric resources alongside the temporal constraints. In this case, we can use an LP to capture both temporal constraints and numeric constraints, including the interaction between the two. We will now describe how the LP is built, serving as a replacement for the $is\_valid\_plan(S)$ function called during search, which invokes the STP solver in CRIKEY3. A diagram of the structure of the LP we create is shown in Figure 6, for a plan $P = [a_0, ..., a_{n-2}, a_{n-1}]$ to reach a state $S$, where $a_{n-1}$ is the action most recently added to the plan. (For simplicity, it shows a case where the event queue $E$ is empty.)

The construction of the LP begins with the variables and (a subset of) the constraints of the STP. Each STP variable $t_i$ (the time-stamp of the (snap) action $a_i$) has a corresponding LP variable $step_i$ (shown across the top of Figure 6), and each STP variable $e_i$ (for the future end of the action at step $i$) has a corresponding LP variable $estep_i$. We also construct the constraints corresponding to the total-ordering of action steps, just as in the STP: each step in $P$ is still sequenced (i.e. $\epsilon \leq step_i - step_{i-1}$ for all $n > i > 0$), and each future end snap-action has to be later than $step_{n-1}$ (i.e. $\epsilon \leq estep_i - step_{n-1}$ for all $estep$ variables).

We then extend the LP with the numeric constraints of the problem, beginning with the effects of actions. Since numeric effects can be both discrete and continuous, we create two additional vectors of variables per step in the plan. The first of these, $\mathbf{v}_i$, represents the values of the state variables $\mathbf{v}$ immediately *prior* to $a_i$ being executed (in the case of step 0, $\mathbf{v}_i$ is equal to the values of $\mathbf{v}$ in the initial state, $I$). The second, $\mathbf{v}'_i$, contains the values of $\mathbf{v}$ immediately *after* $a_i$ is executed. In Figure 6, the variables in $\mathbf{v}_0$ are enumerated as $v_0...v_{m-1}$ and, similarly, those in $\mathbf{v}'_0$ are shown as $v'_0...v'_{m-1}$. To avoid proliferation of indices we do not further index these values with their time stamp in Figure 6, so $v_i$ is the $i$th value in $\mathbf{v}$ at the time step corresponding to the layer in which the variable appears. The use of two vectors at each layer is required in order to represent discrete changes caused by actions: a snap-action can cause the value of a variable to be different immediately after its execution. To represent this within the LP, if an action at step $i$ has no effect on a variable $v$ then $v'_i = v_i{}^2$. Otherwise, for a discrete effect $\langle v' \mathrel{+}= \mathbf{w} \cdot \mathbf{v} + k.(\texttt{?duration}) + c \rangle$, a constraint is introduced to define the value of $v'_i$ :[3]

$$v'_i = v_i + \mathbf{w} \cdot \mathbf{v} + k.(ce(i) - cs(i)) + c$$

where the functions $cs(i)$ and $ce(i)$ denote the time-stamp variables for the corresponding start and end of the action at step $i$. If step $i$ is the end of an action, then $ce(i) = step_i$, and $cs(i)$ is the $step$ variable for the start of the action that finished at step $i$. Similarly, if step $i$ initiates an action, then $cs(i) = step_i$, and $ce(i)$ is either $estep_i$ if the action has not yet finished or, otherwise, the $step$ variable for the end of the action started at step $i$. Therefore, substituting $ce(i) - cs(i)$ for `?duration` captures the relationship between the effect of the action and its duration.

---

2. Note that identities such as this are implemented efficiently by simply not introducing the unnecessary additional variable. Similarly, while a variable is subject to no effects or conditions it is not added to the LP, but it is only introduced once it becomes relevant.

3. For effects using the operator -=, i.e. decrease effects, all but the first term on the right-hand side are negated. For assignment effects, where the operator is =, the first term on the right-hand side (i.e. $v_i$) is omitted entirely (the value of $v$ after such an assignment does not depend on the value of $v$ beforehand).

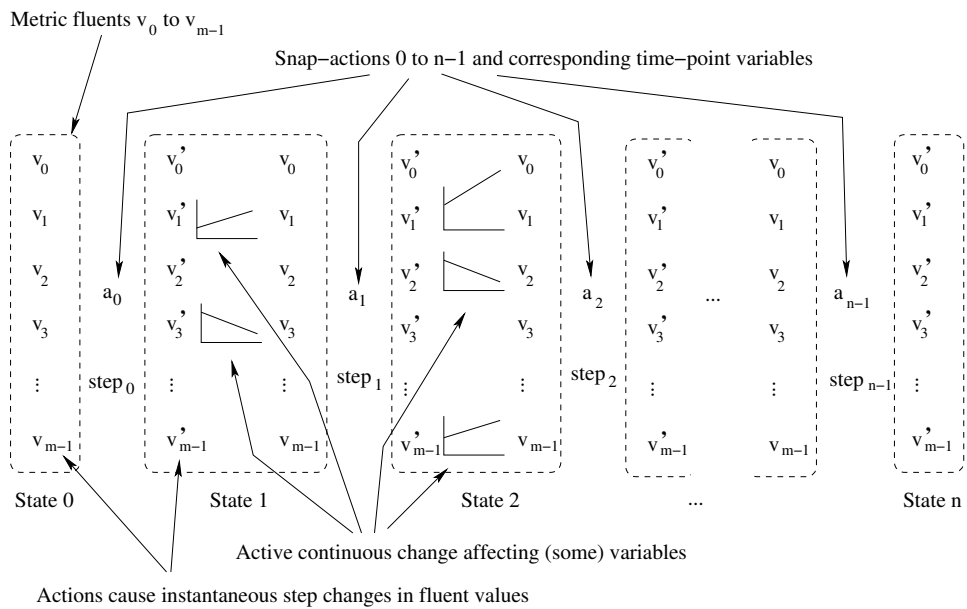

Figure 6: Diagrammatic Representation of the LP Used in COLIN. Note that the subscripts attached to the $v$ and $v'$ fluents in this diagram are indices into the vector of fluents in the state, while indices on *step* and $a$ represent different time steps in the plan. The metric fluents are also notionally indexed by the time step, but this is not shown in the diagram in order to avoid clutter.

Continuous numeric change occurs *between* the steps in the plan, rather than at the instant of execution of the step itself. To capture continuous effects, when building the LP we consider each step in turn, from the start of the plan, recording the gradient of the total (linear) continuous change acting upon each variable $v \in \mathbf{v}$, where $\delta v$ denotes the gradient active after $a_{i-1}$ and before the execution of action $a_i$. Under the restrictions on the language handled by COLIN, described in Section 4, and the total-order constraints between snap-actions, the value of each variable $\delta v_i$ is known and constant within each interval between successive actions: all continuous change is linear. The gradient on a variable $v$ can only be changed by either starting an action (initiating an adjustment to the prevailing continuous effect on $v$ given by $\frac{dv}{dt} += k$, for some $k \in \Re$) or ending an action (terminating the effect initiated by its start). The values of the $\delta$ constants can be computed as follows[4]:

- For all variables, $\delta v_0 = 0$; that is, there is no continuous numeric change active on any variable before the start of the plan.

- If $a_i$ has no continuous numeric effect on $v$ then $\delta v_{i+1} = \delta v_i$;

- If $a_i$ initiates a continuous numeric effect, $\frac{dv}{dt} += k$, then $\delta v_{i+1} = \delta v_i + k$;

- If $a_i$ terminates a continuous numeric effect, $\frac{dv}{dt} += k$, then $\delta v_{i+1} = \delta v_i - k$;

On the basis of these values, we now add constraints to the LP:

$$v_{i+1} = v'_i + \delta v_{i+1}(step_{i+1} - step_i)$$

Again, the distinction between $v_i$ and $v'_i$ is important: $v_i$ is determined on the basis of any continuous change in the interval between steps $i$ and $i-1$, but immediately prior to any discrete effect that may occur at that step.

Having created variables to represent the values of fluents at each step and having introduced constraints to capture the effects of actions on them, we now consider the constraints that arise from the preconditions of each snap-action, the invariants that must be respected between the starts and ends of actions, and any constraints on the durations of each of the actions in the plan. For each numeric precondition of the form $\langle v, \{\geq, =, \leq\}, \mathbf{w} \cdot \mathbf{v} + c \rangle$, that must hold in order to apply step $i$, we add a constraint to the LP:

$$v_i\{\geq, =, \leq\}\mathbf{w} \cdot \mathbf{v}_i + c$$

For an action $a$ starting at $step_i$ and ending at $step_j$, the invariants of $a$ are added to the LP in this form, once for each of the vectors of variables $[\mathbf{v}'_i, \mathbf{v}'_{j-1}]$ and $[\mathbf{v}_{i+1}, \mathbf{v}_j]$ ($\mathbf{v}_i$ and $\mathbf{v}'_j$ are excluded because the PDDL2.1 semantics does not require invariants of an action to hold at its end points). In the case where the end of the action $a$ (starting at $i$) has not yet appeared in the plan, the invariants of $a$ are imposed on all vectors of variables from $\mathbf{v}'_i$ onwards: as $a$ must end in the future, its invariants must not be violated at any step in the current plan after the point where it started.

Finally, we add the duration constraints. For an action $a$ starting at $step_i$, we denote the variable corresponding to the time at which $a$ finishes as $ce(i)$, where $ce(i) = step_j$ if the end of the action has been inserted into the plan at step $j$, or $ce(i) = estep_i$ otherwise (as defined above). Then, for each duration constraint of $a$, of the form $\langle \text{?duration}, \{\geq, =, \leq\}, \mathbf{w} \cdot \mathbf{v} + c \rangle$, we add a constraint:

$$ce(i) - step_i\{\geq, =, \leq\}\mathbf{w} \cdot \mathbf{v}_i + c$$

This process constructs a LP that captures all the numeric and temporal constraints that govern a plan, and the interactions between them. As with the STP in CRIKEY3, a solution to the LP contains values for the variables $[step_0...step_n]$, i.e. an assignment of time-stamps to the actions in the plan. To prevent the LP assigning these variables arbitrarily large (but valid) values, we set the LP objective function to be to minimise $step_n$, where $a_n$ is the last step in the plan so far. For the purposes of the $is\_valid\_plan(S)$ function, if the LP built for a plan $P$ to reach a state $S$ cannot be solved, we can prune the state $S$ from the search space and need not consider it any further: there is no path from $S$ to a legal goal state. In this way, the LP scheduler can be used as a replacement for the STP in order to determine plan validity.

---

4. Variables that can be trivially shown to be constant (i.e. where no action has an effect referring to that variable) can be removed from the LP and replaced throughout by their values in the initial state.

| Plan Action | Delta value | LP Variable | LP Constraints | | |
|---|---|---|---|---|---|
| `saveHard_start` | $\delta m_0 = 0$ | $step_0$ | | $= 0$ | |
| | | $m_0$ | | $= 0$ | |
| | | $m_0'$ | $= m_0$ | $\geq 0$ | |
| `takeMortgage_start` | $\delta m_1 = 1$ | $step_1$ | $\geq step_0 + \epsilon$ | $\leq step_0 + 10$ | |
| | | $m_1$ | $= m_0' + 1.(step_1 - step_0)$ | $\geq 0$ | |
| | | $m_1'$ | $= m_1 - 1$ | $\geq 0$ | $\leq 6$ |
| `lifeAudit_start` | $\delta m_2 = \frac{1}{4}$ | $step_2$ | $= step_1 + \epsilon$ | $\leq step_0 + 10$ | $\leq step_1 + 12$ |
| | | $m_2$ | $= m_1' + \frac{1}{4}.(step_2 - step_1)$ | $\geq 0$ | $\leq 6$ |
| | | $m_2'$ | $= m_2$ | $\geq 0$ | $\leq 6$ |
| `saveHard_end` | $\delta m_3 = \frac{1}{4}$ | $step_3$ | $\geq step_2 + \epsilon$ | $= step_0 + 10$ $\leq step_2 + 4$ | $\leq step_1 + 12$ |
| | | $m_3$ | $= m_2' + \frac{1}{4}.(step_3 - step_2)$ | $\leq 6$ | |
| | | $m_3'$ | $= m_3$ | | |
| `takeMortgage_end` | $\delta m_4 = -\frac{3}{4}$ | $step_4$ | $\geq step_3 + \epsilon$ | $= step_1 + 12$ | $\leq step_2 + 4$ |
| | | $m_4$ | $= m_3' - \frac{3}{4}.(step_3 - step_2)$ | | |
| | | $m_4'$ | $= m_4$ | | |
| `lifeAudit_end` | $\delta m_5 = 0$ | $step_5$ | $\geq step_4 + \epsilon$ | $= step_2 + 4$ | |
| | | $m_5$ | $= m_4' - 0.(step_3 - step_2)$ | $\geq 0$ | |
| | | $m_5'$ | $= m_5$ | | |

Table 2: Variables and constraints for the Borrower problem

## 8.2 Example: LP for the Borrower Problem

In order to illustrate LP construction for a plan we consider the example Borrower problem introduced in Section 4.1. Recall that one solution plan for this problem has the following structure:

```
0: saveHard_start
1: takeMortgage_start longMortgage
2: lifeAudit_start
3: saveHard_end
4: takeMortgage_end longMortgage
5: lifeAudit_end.
```

The LP for this six-step Borrower solution plan contains the variables and constraints shown in Table 2. The six $step$ variables represent the time-stamps of the six snap-actions in the plan, and the variable $m$ represents the money that has been saved by the Borrower. In the initial state, $m = 0$, and hence $m_0 = 0$. Starting the `saveHard` action has no instantaneous numeric effects, introducing the constraint $m'_0 = m_0$ (if it did have an effect on $m$, for instance an instantaneous increase in the savings by $k$, then the constraint would be $m'_0 = m_0 + k$). Due to the invariant condition of the `saveHard` action, that the savings remain above zero, the constraint $m'_0 \geq 0$ is added: it can be seen this constraint is duplicated for each $m_i$ and $m'_i$ during the execution of the `saveHard` action, to ensure that the invariant continues to hold. Notice, also, when the action `takeMortgage` is started, the invariant for that action (the savings level remains less than or equal to the `maxSavings` cap) also appears, and applies to all values of $m$ during its execution. Additional constraints capture discrete change by connecting the value of $m'_i$ to $m_i$. In most cases in this example these values are equal, but one constraint shows a discrete effect: $m'_1 = m_1 - 1$ captures the deduction of the deposit caused by initiating the `takeMortgage` action.

As previously described, the temporal constraints in the LP take two forms. First, there are constraints of the form $step_{i+1} \geq step_i + \epsilon$, forcing $step_{i+1}$ to follow $step_i$, enforcing the sequencing of the snap-actions. Second, duration constraints restrict the duration of actions, e.g. $step_3 = step_0 + 10$ forces that $step_3$ (the end point of `saveHard`) occurs precisely 10 units (the duration of `saveHard`) after $step_0$, its start snap-action.

The final constraints to consider are those modelling the continuous numeric change. The first constraint of this type gives the value of $m_1$ after the execution of `saveHard_start` and before the execution of `takeMortgage_start`. This constraint, $m_1 = m'_0 + 1.(step_1 - step_0)$, is based on the value of $\delta m_1$, which is 1: the only action currently executing with continuous change on $m$ is `saveHard`, which increases it by 1 per unit of time. The second such constraint, $m_2 = m'_1 + \frac{1}{4}.(step_2 - step_1)$, is based on the value of $\delta m_2$ which is now $(1 - \frac{3}{4}) = \frac{1}{4}$, found by adding the active gradients from both of the actions that have started but not yet finished. This illustrates how two actions can have active linear continuous effects on the same variable simultaneously. Note that when `saveHard_end` is applied (at $step_3$) the gradient of continuous change ($\delta m_4$) becomes $-\frac{3}{4}$ as the only active continuous effect is now that of the `takeMortgage` action.

Solving the temporal constraints in this problem without considering the metric fluents yields a solution in which $step_0 = 0$, $step_1 = \epsilon$, $step_2 = 8 + 2\epsilon$, $step_3 = 10$, $step_4 = 12 + \epsilon$ and $step_5 = 12 + 2\epsilon$. Unfortunately, this proposal violates the constraint $m'_1 \geq 0$, since:

$$m'_1 = m_1 - 1 = m'_0 + 1.(step_1 - step_0) - 1 = m_0 + \epsilon - 1 = 0 + \epsilon - 1 = \epsilon - 1$$

and $\epsilon \ll 1$. The constraint on the start time of the `takeMortgage` action cannot be identified because it is dependent on the discrete initial effect of that action, the active continuous effect of the `saveHard` action and the invariant of `saveHard`. This simple example illustrates the strength of using the LP to perform the scheduling alongside the resolution of numeric constraints: the timestamps then satisfy both temporal and numeric constraints.

## 8.3 Temporal–Numeric Search

When performing state-space search, a state, $S$, is a snapshot of the world along some plan trajectory, coming after one action step and before another. In the absence of continuous numeric change, the valuations that define $S$ are known precisely: both which propositions hold, and the values of the numeric variables $\mathbf{v}$. In the presence of continuous numeric change, however, the same does not hold: if a variable $v$ is undergoing continuous numeric change (or is subject to active duration-dependent change) the valuations in a state depend on which snap-actions have been applied so far, on the times at which those snap-actions were applied and on how much time has passed since the last action was applied. Within our representation of the state the time-stamps of the snap-actions in the plan are not fixed (during plan-construction, the LP is used only to confirm that the plan can be scheduled subject to the current constraints), so the valuation of numeric fluents in $S$ is constrained only within ranges determined by the constraints on the temporal variables and the interactions between them.

As a consequence of the flexibility in the commitment to values for temporal and continuously changing variables, COLIN requires a different state representation to the one used in CRIKEY3. Rather than representing the values of the numeric variables by a single vector $\mathbf{v}$, we use two vectors: $\mathbf{v}_{max}$ and $\mathbf{v}_{min}$. These hold the maximum and minimum values, respectively, for each numeric variable in $S$. The computation of these bounds on variables can be achieved using a small extension of the LP described in Section 8.1. For a state $S$, reached by plan $P$ (where $a_n$ is the last step in $P$), we add another vector of variables to the LP, denoted $\mathbf{v}_{now}$, and another time-stamp variable, $step_{now}$. The variables in $\mathbf{v}_{now}$ represent the values of each state variable at some point (at time $step_{now}$) along the state trajectory following $a_n$. The numeric variables and time-stamp for $now$ are constrained as if it were an additional action appended to the plan:

- $now$ must follow the previous step, i.e. $step_{now} - step_n \geq \epsilon$

- $now$ must precede or coincide with the ends of any actions that have started but not yet finished, i.e. for each $estep(i)$, $estep(i) \geq step_{now}$

- For each variable $v_{now} \in \mathbf{v}_{now}$, we compute its value based on any continuous numeric change:

$$v_{now} = v'_n + \delta v_{now}(step_{now} - step_n)$$

- Finally, for every invariant condition $\langle v, \{\geq, =, \leq\}, \mathbf{w} \cdot \mathbf{v} + c\rangle$ of each action that has started but not yet finished:

$$v_{now}\{\geq, =, \leq\}\mathbf{w} \cdot \mathbf{v}_{now} + c$$

The LP can then be used to find the upper and lower bounds on variables. For each of the variables $v_{now} \in \mathbf{v}_{now}$, two calls are made to the LP solver: one with objective set to to maximise $v_{now}$, and one to minimise $v_{now}$. These are then taken as the values of $v_{max}$ and $v_{min}$ in $S$. In the simplest case, where a variable $v$ is not subject to (direct or indirect) continuous or duration-dependent change, the value of $v$ is time-independent, so $v_{max} = v_{min}$, and its value can be determined through the successive application of the effects of the actions in $P$, i.e. the mechanism used in CRIKEY3, or indeed classical (non-temporal) planning.

Since we have upper and lower bounds on the value of each variable, rather than a fixed assignment, the action applicability function, $get\_applicable\_actions(S)$, must be modified. In CRIKEY3, an action is said to be applicable in a state $S$ if its preconditions are satisfied. In COLIN, the definition of what it means for a numeric precondition to be satisfied is different. To preserve completeness, we employ the mechanism used in metric relaxed planning graphs, as discussed in more detail in Section B. Specifically, for a numeric precondition $\mathbf{w} \cdot \mathbf{x} \geq c$, we calculate an optimistic value for $\mathbf{w} \cdot \mathbf{x}$ by using the upper bound on a $v \in \mathbf{x}$ if its corresponding weight in $\mathbf{w}$ is positive, or, otherwise, using its lower bound. Then, if this resulting value is greater than or equal to $c$, the precondition is considered to be satisfied. (As before, for numeric conditions $\mathbf{w} \cdot \mathbf{x} \leq c$, an equivalent precondition in the appropriate form can be obtained by multiplying both sides of the inequality by $-1$ and constraints of the form $\mathbf{w} \cdot \mathbf{x} = c$ are replaced with the equivalent pair of conditions $\mathbf{w} \cdot \mathbf{x} \geq c$, $-\mathbf{w} \cdot \mathbf{x} \geq -c$.)

| Plan Action | Delta value | LP Variable | LP Constraints | | | |
|---|---|---|---|---|---|---|
| saveHard_start | $\delta m_0 = 0$ | $step_0$ | | $= 0$ | | |
| | | $m_0$ | | $= 0$ | | |
| | | $m_0'$ | $= m_0$ | $\geq 0$ | | |
| takeMortgage_start | $\delta m_1 = 1$ | $step_1$ | $\geq step_0 + \epsilon$ | $\leq step_0 + 10$ | | |
| | | $m_1$ | $= m_0' + 1.(step_1 - step_0)$ | $\geq 0$ | | |
| | | $m_1'$ | $= m_1 - 1$ | $\geq 0$ | $\leq 6$ | |
| Now | $\delta m_{now} = \frac{1}{4}$ | $step_{now}$ | $\geq step_1 + \epsilon$ | $\leq step_0 + 10$ | $\leq step_1 + 12$ | |
| | | $m_{now}$ | $= m_1' + \frac{1}{4}.(step_{now} - step_1)$ | $\geq 0$ | $\leq 6$ | |
| | | $m_{now}'$ | $= m_{now}$ | $\geq 0$ | $\leq 6$ | |

Table 3: Variables and constraints for the first stages of the Borrower Problem

This test for applicability of an action is relaxed, so it serves only as a filter, eliminating actions that are certainly inapplicable. For instance, a precondition $a + b \geq 3$ could be satisfied if the upper bounds on $a$ and $b$ are both 2, even if the assignment of timestamps to actions within the LP to attain $a = 2$ conflicts with that needed to attain $b \geq 1$. We rely on the subsequent LP consistency check to determine whether actions are truly applicable. Nonetheless, filtering applicable actions on the basis of the variable bounds in a state is a useful tool for reducing the number of candidates that must be individually verified by the LP.

### 8.3.1 EXAMPLE OF USE OF *now* IN THE BORROWER PROBLEM

We briefly illustrate the way in which the *now* variable is constructed and used in the context of the Borrower problem. Consider the situation after the selection of the first two actions (saveHard_start and takeMortgage_start). The LP construction yields the constraints shown in Table 3. Solving this LP for minimum and maximum values of $step_{now}$ gives values of $1 + \epsilon$ and 10 respectively, meaning that the earliest time at which the third action can be applied will be $1 + \epsilon$ and the latest will be 10.[5] Similarly, solving the LP for minimum and maximum values of $m_{now}$ gives bounds of $\frac{\epsilon}{4}$ and 6. This information could, in principle, constrain what actions can be applied in the current state.

## 8.4 Some Comments on LP Efficiency

An LP is solved at every node in the search space, so it is important that this process is made as efficient as possible. When adding the variable vectors to the LP for each step $i$, it is only necessary to consider a state variable, $v$, if it has become *unstable* prior to step $i$, because of one of the following effects acting on it:

1. direct continuous numeric change, i.e. changing $v$ according to some gradient;

---

5. In practice, for efficiency, COLIN does not actually solve the LP for minimum and maximum values of $step_{now}$, but uses the variable only to communicate constraints to the metric variables in this state.

2. direct duration-dependent change, i.e. a change on $v$ dependent on the duration of an action (whose duration is non-fixed);

3. discrete change, where the magnitude of the change was based on one or more variables falling into either of the previous two categories.

All variables that do not meet one of these conditions can be omitted from the LP, as their values can be calculated based on the successive effects of the actions applied up to step $i$, and substituted as a constant within any LP constraints referring to them. This reduces the number of state variables and constraints that must be added to the LP and also reduces the number of times the LP must be solved at each state to find variable bounds: irrelevant variables can be eliminated from the vector $\mathbf{v}_{now}$. A similar simplification is that, if applying a plan $a_0...a_{n-1}$ reaches a state $S$ where $v_{min} = v_{max}$, then if there is no continuous numeric change acting on $v$, $v$ has become *stable*, i.e. its value is independent of the times assigned to the preceding plan steps. In this case, until the first step $k$ at which $v$ becomes unstable, the value of $v$ can be determined through simple application of discrete effects, and hence $v$ can be omitted from all $\mathbf{v}_j, \mathbf{v}'_j, n-1 < j$.

A further opportunity we exploit is that the LP solved in each state is similar to that being solved in its parent state: it represents the same plan, but with an extra snap-action appended to the end. The lower bounds of the time-stamp variables in the LP can therefore be based on the values computed in the parent states. Suppose a state $S$ is expanded to reach a state $S'$ by applying a snap action, $a$, as step $i$ of the plan. At this point, the LP corresponding to the plan will be built and solved with the objective being to minimise $step_i$. Assuming the plan can indeed be scheduled (if it cannot, then $S'$ is pruned and no successors will be generated from it), the value of the objective function is stored in $S'$ as a lower bound on the time-stamp of $a$. In all states subsequently reached from $S'$, this stored value can be used in the LP as a lower bound on $step_i$ — appending actions to the plan can further constrain and hence increase the value of $step_i$, but it can never remove constraints in order to allow it to decrease.

As well as storing lower bounds for time-stamp variables, we can make use of the bounds $\mathbf{v}_{min}, \mathbf{v}_{max}$ in the state $S'$ when generating successors from it. In a state $S$ reached via plan of length $i$, applying an action $a$ leads to a state $S'$ in which the new action at $step_{i+1}$ inherits the constraints imposed previously on $step_{now}$ when calculating the variable bounds in $S'$. Therefore, the values of $\mathbf{v}_{max}$ and $\mathbf{v}_{min}$ in $S$ serve as upper and lower bounds (respectively) for $\mathbf{v}_{i+1}$ in the LP built to determine the feasibility of $S'$. Similarly, we can combine any discrete numeric effects of $a$ with the values of $\mathbf{v}_{max}$ and $\mathbf{v}_{min}$ in $S$ to give bounds on $\mathbf{v}'_{i+1}$. For each variable $v$ subject to an effect, an optimistically large (small) outcome for that effect can be computed on the basis of $\mathbf{v}_{max}$ and $\mathbf{v}_{min}$, and taken as the upper (lower) bound of $v'_{i+1}$. Otherwise, for variables upon which $a$ has no discrete effect, $v'_{i+1} = v_i$.

Finally, the presence of timed initial literals (TILs) allows us to impose stricter bounds on the time-stamp variables. If step $j$ of a plan is the dummy action corresponding to a TIL at time $t$, the upper bound on $step_i$, $i < j$, is $t - \epsilon$ and the lower bound on each $step_k$, $j < k$ (or any $estep$ variable) is $t + \epsilon$. Similarly, if the plan does not yet contain a step corresponding to a TIL at time $t$, the upper bound on all $step$ variables is $t - \epsilon$. Furthermore, a TIL at time $t$ corresponds to a deadline if it deletes some fact $p$ that is present in the initial state, never added by any action, and never reinstated by any other TIL. In this case:

- if a plan step $i$ requires $p$ as a precondition, then $step_i \leq t - \epsilon$;

- if $estep_i$ is the end of an action with an end condition $p$, then $estep_i \leq t - \epsilon$;

- if $estep_i$ is the end of an action with an invariant condition $p$, then $estep_i \leq t$.

## 9. Heuristic Computation

The search algorithms described so far in this paper all make use of a heuristic to guide the planner efficiently through the search space towards the goal. Having introduced the necessary machinery to support linear continuous numeric and duration-dependent effects we now turn our attention to the construction of an informed heuristic in the face of time-dependent change.

In Appendices B and C we revisit the standard Metric-FF Relaxed-Planning Graph (RPG) heuristic and the Temporal RPG (TRPG) used in CRIKEY3, and provide the details of these approaches for reference. Both of these depend on the initial construction of a reachability graph, based on the plan graph introduced in Graphplan (Blum & Furst, 1995). The graph consists of alternating layers of facts ($fl$) and actions ($al$). In the TRPG, for convenience, we index these layers by the earliest time they could represent, although they can still be enumerated by consecutive integers because only finitely many times can be relevant in the process of construction. In this section we explain how the heuristic computation techniques introduced by these planners can then be modified to reason with interacting temporal–numeric behaviour. We describe two variants of the heuristic: a basic version, in which active continuous change is relaxed to discrete step changes, and a refined variant in which this relaxation is replaced with a more careful approximation of the continuous values. We show, using the Borrower example, the benefits of the refined approach.

The heuristics are based on the underlying use of a relaxed plan step-count. We use the relaxed plan makespan as a tie-breaker in ordering plans with the same step-count. Step-count dominates our heuristic because our first priority is to find a feasible solution to a planning problem and this means attempting to minimise the number of choices that must be made and resolved during the search. Of course, the emphasis on rapidly finding a feasible plan can compromise the quality of the plan, particularly in problems where the step-count is poorly correlated with the makespan. Subsequent attempts to improve the quality of an initial feasible solution, either by iteratively improving the solution itself or by further search using the bound derived from the feasible solution to prune the search space, are possible, but we do not consider them in this work.

### 9.1 The Basic 'Integrated' Heuristic Computation with Continuous Numeric Effects

The first version of COLIN (Coles, Coles, Fox, & Long, 2009b) introduced three significant modifications to the TRPG used in CRIKEY3, in order to generate heuristic values in the presence of continuous and duration-dependent effects. The first modification simply equips the heuristic with the means to approximate the effects of continuous change.

- If an action $a$ has a continuous effect equivalent to $\frac{dv}{dt} += k$ it is relaxed to an instantaneous *start* effect $\langle v, +=, k * dmax(a) \rangle$. That is, the effect on the changing variable is treated as the integral of the effect up to an upper bound on the duration of the action and is applied at the start of the action. Doing this ensures that the behaviour is *relaxed*, in contrast to, say, applying the effect at the end of the action. $dmax(a)$ is calculated at the point where the action is added to the TRPG, based on the maximum duration constraints of $a$ that refer only to variables that cannot change after that time (that is, they are state-independent). If no such constraints exist, the duration is allowed to be infinite (and variables affected by continuous effects of the action will then have similarly uninformed bounds).

- If an action $a$ has a discrete duration-dependent effect on a variable $v$ then, when calculating the maximum (minimum) effect of $a$ upon $v$ (as discussed, in the non-temporal case, in Appendix B), the `?duration` variable is relaxed to whichever of $dmin(a)$ or $dmax(a)$ gives the largest (smallest) effect. Relaxation of this effect is achieved without changing its timing, so it is associated with the start or end of the action as indicated in the action specification.

The second modification affects any action that has a continuous numeric effect on some variable and either an end precondition or invariant that refers to the same numeric variable. If the invariant or end precondition places a constraint on the way in which the process governed by the action can affect the value of a variable, then this constraint is reflected in the corresponding upper or lower bounds of the value of the variable. Specifically, if an action $a$ decreases $v$ at rate $k$ and has an invariant or end precondition $v \geq c$, then the upper bound on $v$ by the end of the action must be at least $k.(dmin(a) - elapsed(a)) + c$, where $elapsed(a)$ is the maximum amount of time for which $a$ could have been executing in the state being evaluated (0 if $a$ is not currently executing, otherwise, the maximum from all such entries in $E$). This condition ensures that the variable could achieve the necessary value to support the application of the action. It might appear strange that the bound is set to be higher than $c$, but the reason is that the relaxation *accumulates* increase effects and ignores decrease effects in assessing the upper bound, so it will be necessary, by the end of the action, to have accumulated increases in the value of the variable that allow for the outstanding consumption from $a$ in order to still meet the $c$ bound at the end of the action. A corresponding condition is required for an action that $a$ increases $v$ at rate $k$, and has an invariant or end precondition $v \leq c$, where the lower bound on $v$ cannot be more than $k.(dmin(a) - elapsed(a)) + c$. These conditions are added as explicit additional preconditions to $a_\dashv$ for the purposes of constructing the TRPG.

The third modification deals with the problem of constructing an appropriate initialisation of the bounds for the numeric variables in the first layer of the TRPG. In CRIKEY3 these values are initialised to the actual values of the metric variables, since their values in the current state do not change if time passes without further actions being applied. The same is not true in COLIN, since any actions that have started, but not yet finished, and which govern a process, will cause variables to change simply as a consequence of time passing. As the basic heuristic proposed here relies on being able to integrate continuous numeric change, we determine the variable bounds in $fl(0.0)$ in two stages. First, the bounds on a variable $v$ are set according to those obtained from the LP in Section 8.3. Then, for each entry $e \in E$, corresponding to the start of an action, $a$, with a continuous effect on $v$ having positive gradient $k$, the upper bound on $v$ in $fl(0.0)$ is increased by $k.remaining(e)$. Here, $remaining(e)$ is the maximum amount of time that could elapse between the state being evaluated and the future end snap-action paired with start event $e$. The maximum remaining execution time is calculated by subtracting the lower bound for the amount of time that *has* to have elapsed since the start of action $a$ from its maximum duration. In the case where the gradient is negative, the lower bound is decreased.

## 9.2 The Refined Integrated Heuristic

Time-dependent change arises from two sources: continuous numeric effects, initiated by start snap-actions, and discrete duration-dependent effects which can apply at either end of durative actions.

For the purposes of the refined heuristic described in this section, we treat continuous effects and discrete duration-dependent effects at the ends of actions of these in the same way, attaching a continuous linear effect acting on each relevant variable to the effects of the appropriate snap-action, $a$, denoting the set of all such continuous effects by $g(a)$. For continuous effects, $cont(a)$, initiated by $a_{\vdash}$, $cont(a) \subseteq g(a_{\vdash})$. That is, the gradient effects of the start of $a$ include all of the continuous effects of $a$. For duration-dependent effects of an end snap-action $a_{\dashv}$ we split the effect into two parts:

- a discrete effect of $a_{\dashv}$, $\langle v, \{+=, -=, =\}, \mathbf{w} \cdot \mathbf{v} + k.dmin(a) + c \rangle$ and

- a gradient effect on $v$, added to $g(a_{\dashv})$. The effect is defined as $\langle v, k \rangle$ if the original effect used the operator += or = otherwise, it is $\langle v, -k \rangle$.

Thus, instantaneously, at the end of $a_{\dashv}$, the effect of $a$ is available assuming the smallest possible duration for $a$ is used. As $a$ executes with a greater duration, a continuous effect is applied with the gradient of the change being taken from the coefficient $k$ of the `?duration` variable in the corresponding effect in $a$.

Unfortunately, the treatment proposed above cannot be applied to duration-dependent start effects, since the effects are always available at the start of the action, regardless of the duration. Thus, we employ the approach taken with the basic heuristic used in COLIN: when calculating the maximum (minimum) effect of $a_{\vdash}$ on the affected variable, $v$, the `?duration` variable is substituted with whichever of $dmin(a)$ or $dmax(a)$ gives the largest (smallest) effect.

Once we have a collection of linear continuous effects, $g(a)$, associated with each snap-action, $a$, we can adjust the construction of the TRPG. First, we identify, for each variable, $v$, an associated maximum rate of change, $\delta v_{max}(t)$, following the layer $al(t)$. We set this to be the sum of all the positive rates of change, affecting $v$, of any snap-actions in $al(t)$:

$$\delta v_{max}(t) = \sum_{a \in al(t)} \sum_{\langle v,k \rangle \in g(a)} k$$

This definition relies on the restriction that only one instance of any action can execute at any time. If this restriction does not hold, but there is a clear finite bound $p(a)$ on the number of instances of an action that can execute concurrently, then we incorporate this into the calculation of $\delta v_{max}(t)$ as follows:

$$\delta v_{max}(t) = \sum_{a \in al(t)} p(a) \times \sum_{\langle v,k \rangle \in g(a)} k$$

Where no such finite bound exists, an action could, in principle, be applied arbitrarily many times in parallel and hence we set $\delta v_{max}(t) = \infty$.[6] Following any layer $al(t)$ at which $\delta v_{max}(t) = \infty$ we no longer need to reason about the upper bound of the continuous change on $v$ since the upper bound on $v$ itself will become $\infty$ immediately after this layer. It should be noted that this degradation of behaviour will, in the worst case, lead to the same heuristic behaviour as the basic heuristic where, again, if arbitrarily many copies of the same action can execute concurrently, the magnitude of its increase or decrease effects becomes unbounded. The extension of the heuristic to consider continuous effects in a more refined way does not worsen its guidance in this situation. For the remainder of this section, we consider only variables whose values are modified by actions for which there are finite bounds on the number of concurrently executing copies allowed.

---

6. We note that, in our experience, the presence of infinitely self-overlapping actions with continuous numeric change is often a bug in the domain encoding: it is difficult to envisage a real situation in which parallel production is unbounded.

Armed with an upper bound value for the rate of change of each variable following layer $al(t)$, we can deduce the maximum value of each variable at any time $t' > t$, by simply applying the appropriate change to the maximum value of the variable at time $t$. The remaining challenge is to decide how far to advance $t'$ in the construction of the TRPG. During construction of the TRPG in CRIKEY3 time is constrained to advance by $\epsilon$ or until the next action end point, depending on whether any new facts are available following the most recent action layer (lines 29–34 of Algorithm 2). In order to manage the effects of the active continuous processes, we add a third possibility: time can advance to the earliest value at which the accumulated effect of active continuous change on a variable can satisfy a previously unsatisfied precondition. The set of preconditions of interest will always be finite, so, assuming that the variable is subject to a non-zero effect, the bound on the relevant advance is always defined (or, if the set of preconditions is empty, no advance is required). We can compute the value of this time as follows. Each numeric precondition may be written as a constraint on the vector of numeric variables, $\mathbf{v}$, in the form $\mathbf{w} \cdot \mathbf{v} \geq c$, for vectors of constants $\mathbf{w}$ and $\mathbf{c}$. We define the function $ub$ as follows:

$$ub(\mathbf{w}, \mathbf{x}, \mathbf{y}) = \sum_{\mathbf{w}[i] \in \mathbf{w}} \left\{ \begin{array}{ll} \mathbf{w}[i] \times \mathbf{y}[i] & \text{if } \mathbf{w}[i] \geq 0 \\ \mathbf{w}[i] \times \mathbf{x}[i] & \text{otherwise} \end{array} \right.$$

The upper bound on $\mathbf{w} \cdot \mathbf{v}$ at $t'$ is then: $ub(\mathbf{w}, \mathbf{v}_{min}(t'), \mathbf{v}_{max}(t'))$.

The earliest point at which the numeric precondition $\mathbf{w} \cdot \mathbf{v} \geq c$ will become satisfied is then the smallest value of $t'$ for which $ub(\mathbf{w}, \mathbf{v}_{min}(t'), \mathbf{v}_{max}(t')) \geq c$.

As an example, suppose there is an action with a precondition $x + 2y - z \geq c$, so that $\mathbf{w} = \langle 1, 2, -1 \rangle$ (assuming $x$, $y$ and $z$ are the only numeric fluents in this case). Substituting this into the previous equation yields:

$$
\begin{aligned}
ub(\langle 1, 2, -1 \rangle, \langle x, y, z \rangle_{min}(t'), \langle x, y, z \rangle_{max}(t')) &= 1.x_{max}(t') + 2.y_{max}(t') - 1.z_{min}(t') \\
&= 1.(\delta x_{max}(t) \times (t' - t - \epsilon) + x_{max}(t + \epsilon)) \\
&\quad + 2.(\delta y_{max}(t) \times (t' - t - \epsilon) + y_{max}(t + \epsilon)) \\
&\quad - 1.(\delta z_{min}(t) \times (t' - t - \epsilon) + z_{min}(t + \epsilon))
\end{aligned}
$$

(The values of $x$, $y$ and $z$ are based on their starting points at $t + \epsilon$ because this accounts for any instantaneous changes triggered by actions in $al(t)$.) If the value of $t'$ produced by this computation is infinite, then the maximum possible rate of increase of the expression $x + 2y - z$ must be zero.[7] Otherwise, $t'$ is the time at which a new numeric precondition will first become satisfied due to active continuous effects and, if this is earlier than the earliest point at which an action end point can be applied, then the next fact layer in the TRGP will be $fl(t')$.

### 9.2.1 IMPROVING THE BOUNDS ON VARIABLES IN FACT-LAYER ZERO

Previously, setting the bounds in fact-layer zero could be thought of as consisting of two stages: finding initial bounds using the LP and then, because the passage of time could cause these bounds to further diverge due to active continuous numeric change, integrating this change prior to setting bounds for layer zero of the TRPG. With an explicit model of numeric gradients in the planning graph, we can now reconsider this approach. The intuition behind our new approach here is as follows:

1. For each variable $v$, create an associated variable $t_{now}(v)$ in the LP, and solve the LP to minimise the value of this variable.

2. Fixing the value of $t_{now}(v)$ to this lower-bound, maximise and minimise the value of $v$ to find the bounds on it *at this point* — these are then used as the bounds on $v$ in $fl(0.0)$.

3. If $\delta v > 0$ in the current state, then all $\delta v_{max}(t)$ values in the TRPG are offset by $\delta v$ or, similarly, if $\delta v < 0$, all $\delta v_{min}(t)$ values are offset.

---

7. To find $t'$ requires only a simple rearrangement of the formula to extract $t'$ directly.

The first of these steps is based on the ideas described in Section 8.3, but the process is subtly different because we are trying to determine the bounds on $v$ at a given point in time, rather than those that appear to be reachable. As before, $t_{now}(v)$ must still come after the most recent plan step and is used to determine the value of $v$. This is reflected by the pair of constraints:

$$t_{now}(v) - step_i \geq \epsilon$$

$$v_{now} = v'_i + \delta v_{now}(t_{now}(v) - step_i)$$

Additionally, since the 'now' variable is associated with only a single $v$, rather than having to be appropriate for *all* $v$, we can further constrain it if, necessarily, $v$ cannot be referred to (either in a precondition, duration or within an effect) until at least after certain steps in the plan, rather than the weaker requirement of just after the most recent step. For our purposes, we observe that if all actions referring to $v$ require, delete and then add a fact $p$, and all possible interaction with $p$ is of this require-delete-add form, then $t_{now}(v)$ must come after any plan step that adds $p$. More formally, the require-delete-add idiom holds for $p$ if $p$ is true in the initial state, and for each action $a$ with preconditions/effects on $p$, the interaction between the action and $p$ can be characterised as one of the following patterns:

1. $p \in pre_{\vdash}(a), p \in \mathit{eff}_{\vdash}^-(a), p \in \mathit{eff}_{\vdash}^+(a)$

2. $p \in pre_{\dashv}(a), p \in \mathit{eff}_{\dashv}^-(a), p \in \mathit{eff}_{\dashv}^+(a)$

3. $p \in pre_{\vdash}(a), p \in \mathit{eff}_{\vdash}^-(a), p \in \mathit{eff}_{\dashv}^+(a)$

(An action may exhibit either or both of the first two interactions, or just the third.)

The LP variable corresponding to the point at which $p$ is added, which we denote $step_p$, is determined in one of two ways. First, if $p$ is present in the state being evaluated, $step_p$ is the LP variable corresponding to the plan step that most recently added $p$. Otherwise, from case 4 above, we know that $p \in \mathit{eff}_{\dashv}^+(a)$ for some action $a$ that is currently executing. In this case, $step_p$ is the LP variable $estep_i$ corresponding to the end of $a$. With this defined variable, we can add the constraint to the LP:

$$t_{now}(v) \geq step_p + \epsilon$$

Solving the LP with the objective being to minimise $t_{now}(v)$ finds the earliest possible time at which $v$ can be referred to. Then, fixing $t_{now}(v)$ to this minimised value, we minimise and maximise the bounds on $v_{now}$. This gives us bounds on $v$ that are appropriate as early as possible after the actions in the plan so far.

Having obtained variable bounds from the LP we must, as before, account for the fact that the passage of time causes the bounds to change if there is active continuous numeric change. Whereas before we integrated this change prior to the TRPG, we now have a mechanism for handling gradients directly during TRPG expansion. Thus, for each start-event-queue entry $e \in E$ corresponding to the start of an action, $A$, with a continuous effect on $v$ with a positive (negative) gradient $k$, we add a gradient effect on the upper (lower) bound on $v$ to the TRPG. Just as we previously restricted the integrated effect of $e$ by $remaining(e)$, the maximum remaining time until the action must end, so here we limit how long the gradient effect is active: it starts at $al(0.0)$ and finishes at $al(remaining(e))$. Then, for a given fact layer $t$ the value of $\delta v_{max(t)}$ is updated accordingly:

$$\delta v_{max}(t) += \sum_{e \in E} \sum \{k \mid \langle v, k \rangle \in g(op(e)) \wedge k > 0 \wedge t \leq remaining(e)\}$$

Similarly, $\delta v_{min(t)}$ is amended to account for effects $\langle v, k \rangle, k < 0$.

### 9.3 Using the Two Variants of the Integrated Heuristic in the Borrower Problem

We now illustrate the computation of the two heuristic functions for a choice point in the Borrower problem. This example shows that the refined heuristic guides the planner to a shorter makespan plan than the basic heuristic, because the improved heuristic information leads to the selection of better choices of helpful actions. Consider the situation following execution of the first action, `saveHard_start`. Figure 7 (top) shows the TRPG and relaxed plan constructed using the basic heuristic.

The heuristic generates a cost for this state of 5: the four actions shown in the relaxed plan, together with an extra one to end the `saveHard` action that has already started. This relaxed plan generates two helpful actions, to start the `lifeAudit` and to start `takeMortgage short`. An attempt to start the `lifeAudit` action can quickly be dismissed as temporally inconsistent, depending as it does on `boughtHouse` becoming true before it ends, so the other helpful action is chosen. Unfortunately, once this action is selected the interaction between the saving process and the deposit requirement (at least five savings must have been acquired) forces the action to start no earlier than time 5. This constraint is invisible in the TRPG, because the continuous effect of `saveHard` has been abstracted to a start effect, and a full ten savings therefore appear to be available immediately. A plan can be constructed using the short mortgage, but only by introducing a second saving action as shown in the lower plan in Figure 3. This is because the start of the short mortgage is pushed so late that the life audit cannot both overlap the end of the first `saveHard` action and finish after the mortgage action.

The lower part of Figure 7 shows what happens when the refined heuristic is used to solve this problem. The `saveHard` action starts as before, but this time the heuristic does not relax the behaviour of the continuous savings process so the long mortgage, which requires a smaller deposit to initiate it, becomes available before the short mortgage. As a consequence of this, the relaxed plan selects the long mortgage, and this action starts early enough that the life audit can overlap both its end and the end of the `saveHard` action. The planner is correctly guided to the optimal plan, as shown at the top of Figure 3. The crucial difference between the two heuristics, is that the refined heuristic is able to access more accurate information about the value of the savings at timepoints after the start of the savehard action. This leads to a finer-grained structure of the TRPG, which can be seen in the fact that there are six action layers before arrival at the goal, rather than four as in the case when the basic heuristic is used. The estimated makespan of the final plan is $12 + \epsilon$, while the makespan according to the basic heuristic is $10 + 2\epsilon$. The basic heuristic leads to a non-optimal solution because it requires the extra saveHard action, giving a solution makespan of $20 + 2\epsilon$, in contrast to the makespan of $12 + \epsilon$ of the optimal plan.

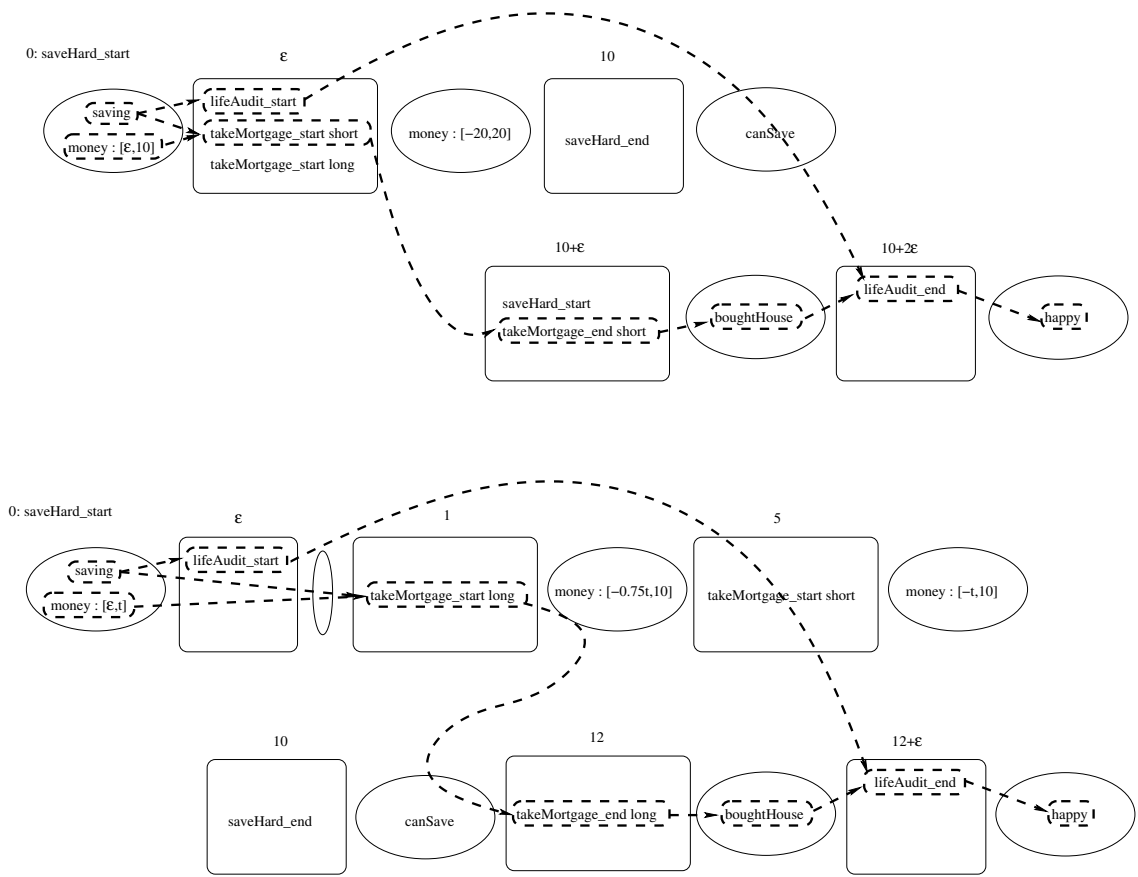

Figure 7: The TRPG and relaxed plan for the Borrower problem, following initial execution of saveHard_start at time 0, as constructed using the original version of COLIN (top — described in Section 9.1) and the revised version (bottom — described in Section 9.2). Action layers are depicted in rounded rectangles and fact layers in ovals. The action layers are labelled with their times constructed during the reachability analysis.

The benefit of the refined heuristic, and the extra work involved in constructing the modified TRPG, is that better helpful actions are chosen and the makespan estimate is therefore more accurate. The choice between similar length plans is made based on makespan. The TRPG, constructed by the refined heuristic in the Borrower problem, does not even contain the short mortgage action at an early enough layer for it to be considered by the relaxed plan.

## 10. Improving Performance

In this section we present two techniques we use to improve the performance of COLIN. The first technique, described in Section 10.1, is a generalisation of our earlier exploitation of *one-shot actions* (Coles et al., 2009a) to the situation in which they encapsulate continuous processes, leading to faster plan construction in problems with these action types. The second technique, described in Section 10.2, exploits the LP that defines the constraints within the final plan to optimise the plan metric. This leads to better quality plans in many cases.

### 10.1 Reasoning with One-Shot Actions

In earlier work (Coles et al., 2009a) we have observed that there is a common modelling device in planning domains that leads to use of actions that can only be applied once. We call these actions *one-shot actions*. They arise, in particular, when there is a collection of resources that can each be used only once. The key difference that one-shot actions imply for the TRPG is that continuous effects generated by one-shot actions lapse once a certain point has been reached:

- If a one-shot action $a$ has a continuous numeric effect on $v$, and $a_{\vdash}$ first appears in action layer $al(t)$, then the gradient on $v$ due to this effect of $a$ finishes, at the latest, at $al(t + dmax(a))$.

- If the end $a_{\dashv}$ of a one-shot action has a duration-dependent effect on $v$, then the (implicit) continuous effect acting on $v$ finishes, at the latest, at layer $al(t + dmax(a))$

The termination point is implied, in both cases, by the fact that the action is one-shot.

We modify the TRPG construction to reflect these restrictions by extending the data recorded in each action layer to include, for each snap-action action $a$, the maximum remaining execution time of $a$, denoted $rem(t, a)$. For one-shot actions, in the layer $al(t)$ in which $a_{\vdash}$ first appears, $rem(t, a_{\vdash}) = dmax(a)$, and when $a_{\dashv}$ first appears, $rem(t, a_{\dashv}) = dmax(a) - dmin(a)$. For actions that are not one-shot $rem(t, a_{\vdash})$ and $rem(t, a_{\dashv})$ are both initialised to $\infty$. We make three minor changes to the layer update rules to accommodate the $rem$ values. First, when calculating the active gradient on a variable $v$ following action layer $al(t)$:

$$\delta v_{max}(t) = \sum_{a \in al(t) | rem(a,t) > 0} p(a) \times \sum_{\langle v,k \rangle \in g(a)} k$$

As can be seen, only the subset of actions with execution time remaining is considered. Second, at the next action layer $al(t + \Delta t)$ following $al(t)$, the value of each positive $rem$ is decremented by $\Delta t$, the amount of time elapsed since the previous layer. Third, as a consequence of this, an additional criterion must be considered when calculating the time-stamp of the next fact-layer, $t'$, described in Section 9.2. Since the time remaining to complete an action may expire, we may need to insert an additional fact layer to denote the point at which a $rem$ value reaches 0 and the continuous effects acting on one or more variables need to be recalculated. The time-stamp of the earliest such layer is:

$$t' = t + \min\{rem(t, a) > 0 \mid a \in al(t)\}$$

One-shot actions can be exploited still further by improving the upper bound on the duration of the action $a$. In the case of actions with state-dependent duration constraints (i.e. where the upper-bound is calculated based on variables that can be subjected to the effects of actions), $dmax(a)$ may be a gross over-estimate of the duration of $a$. Suppose the maximum duration of $a$ is bounded by a formula $w \cdot v + c$. In the layer $al(t)$ in which $a_\vdash$ appears, we can compute the maximum duration of $a$, were it to be started in that layer, based on the variable bounds recorded in $fl(t)$. We could use this value to determine a bound on the remaining execution time for $a$. However, at some future layer $fl(t')$, the variable bounds might have changed, so that beginning $a$ in $al(t')$, and calculating its maximum duration based on $fl(t')$, would have allowed $a$ to execute for a possibly longer period of time, allowing its continuous effects to persist for longer.

To remain faithful to the relaxation, the possibility of exploiting this increased duration of $a$ (by starting $a$ at $t'$) must be included in the TRPG, as well as allowing the possibility of $a$ to start at $t$, thereby obtaining its effects sooner. Therefore, each one-shot action is allowed to start in the earliest layer $al(t)$ in which its preconditions are satisfied, giving it an initial maximum duration of $dmax(a, t)$ based on the fact later $fl(t)$. But, if a later fact layer $fl(t')$ admits a greater duration ($dmax(a, t')$, the value of $dmax$ for action $a$ at layer $t'$), the remaining execution time for $a$ is reconsidered. First, in the simple case, the variables in the duration constraint are changed in $fl(t')$, but not subject to any active continuous effects. In this case, we apply a pair of dummy effects to fact layer $t'' = t' + dmax(a, t)$:

$$\langle rem(a_\vdash, t'') \mathrel{+}= (dmax(a, t') - dmax(a, t)) \rangle$$

and

$$\langle rem(a_\dashv, t'') \mathrel{+}= (dmax(a, t') - dmax(a, t)) \rangle.$$

Note that the increase of the $rem$ values is delayed until layer $t''$ because, in order to benefit from the longer duration of $a$, $a$ must have started in layer $t'$.

In the more complex case, the variables in the duration constraint are changed in $fl(t')$ but the duration is also affected by continuous effects on some of the variables it depends on. In this situation, each subsequent fact layer might admit a marginally bigger duration for $a$ than the last. To avoid having to recalculate the new duration for $a$ repeatedly, we schedule a pair of dummy effects based on the global, layer-independent, maximum value for the duration of $a$:

$$\langle rem(a_\vdash, t'') \mathrel{+}= (dmax(a) - dmax(a, t)) \rangle$$

and

$$\langle rem(a_\dashv, t'') \mathrel{+}= (dmax(a) - dmax(a, t)) \rangle.$$

This relaxation is weaker than it might be, but is efficient to compute.

## 10.2 Plan Optimisation

A plan metric can be specified in PDDL2.1 problem files to indicate the measure of quality to use in evaluating plans. The metric is expressed in terms of the task numeric variables and the total execution time of the plan (by referring to the variable `total-time`). The use of an LP in COLIN offers an opportunity for the optimisation of a plan with respect to such a metric: for a plan consisting of $n$ steps, the numeric variables $\mathbf{v}'_{n-1}$ are those at the end of the plan, the $step_{n-1}$ is the time-stamp of the final step (i.e. the action dictating the makespan of the plan) and the LP objective can be set to minimise a function over these. The LP must be solved to minimise the time-stamp of the last action (the makespan of the plan) in order to arrive at a lower bound on the time for the next action. However, it can also be solved to optimise the plan metric.

Although it is possible to consider ways to use the metric-optimising LP value during plan construction, to guide the search, we have focussed on a much more limited, but less costly, use: we only attempt *post hoc* optimisation, attempting to exploit any flexibility in the temporal structure of the final plan to optimise the plan quality at the last stage of plan construction.

In order for such *post hoc* optimisation to be useful, planning problems must have the property that it is possible to vary the quality metric of a plan by scheduling the same actions to occur at different times. This is possible in a wide range of interesting situations, such as scheduling aircraft to land as close to a given target time as possible, taking images from satellites at certain times of day when the view is clearer, or minimising wasted fuel by penalising the time elapsing between starting the engine of a plane and its take off. The last of these represents a general class of problems in which it may be desirable to minimise the amount of time between two activities: a different metric to the total time taken for plan execution. To capture these interesting cases, we first extend the language supported by COLIN to allow a limited subset of ADL conditional effects, to allow the conditions under which an action is executed to vary the effects the action has on the metric value of the plan. Second, we discuss how a MILP can be built, based on the LP described in Section 8.1, to support *post hoc* plan optimisation.

Our planner handles most conditional effects by a standard compilation. However, conditional effects on metric variables that appear in the plan quality metric and not in the preconditions of any actions are dealt with differently. We call such variables *metric tracking variables* and we exploit the fact that rescheduling a plan can affect the values of these variables without changing the validity of the plan. An example is shown in Figure 10.2 of an action to land an airplane, with conditional effects on the metric tracking variable `total-cost`. The domain is structured so that all land actions must start at the beginning of the plan, and their end points represent the actual landing times of the aircraft in the problem. The duration of the actions are set to correspond to the earliest and latest possible points at which each plane could land. As can be seen, the propositional effects of the action are the same whether the plane lands early or late: the plane has landed, and is no longer flying. However, the numeric effects, which are all effects on the metric tracking variable `total-cost`, depend on the duration of the action and, in particular, whether the plane has landed early or late. If the plane lands early, a penalty is paid at a certain rate per unit time that the plane lands before the desired target value. If the plane lands late, then a fixed cost is paid, in addition to a penalty (at a different rate) per unit of time the plane lands after the desired target value. By considering this action as a single action, with a pair of conditional effects, the planner can decide upon the actions needed to construct a sound plan (in which all the planes have landed) whilst leaving to the subsequent optimisation phase the decision about whether a plane should be landed early, late, or on time.

In general, it is straightforward to exploit the LP described in Section 8.1 to attempt to reschedule the actions in a plan to optimise the value of the plan metric (provided that the metric function is linear). However, if the plan contains actions with conditional effects on metric tracking variables, it becomes possible to exploit the representation of these effects in an extended LP, using integer variables, in order to offer a more powerful optimisation step. Each conditional effect will either be activated or not: we introduce a 0-1 variable to represent which of these is the case for each effect. The variable is connected to corresponding constraints that determine whether or not the condition associated with the effect is true or not.

We deal with two kinds of constraints on the 0-1 variables in our MILP encoding of the plan optimisation problem: one is the special case where actions are scheduled against fixed time-windows

```
(:durative-action land
:parameters (?p - plane ?r - runway)
:duration (and (>= ?duration (earliest ?p)) (<= ?duration (latest ?p)))
:condition
    (and
       (at start (takeOff))
       (over all (flying ?p))
       (at end (scheduled ?p ?r)))
:effect
    (and
       (at start (flying ?p))
       (at end (landed ?p))
       (at end (not (flying ?p)))
       (when (at end (< (?duration) (target ?p)))
         (at end
            (increase (total-cost)
       (* (earlyPenaltyRate ?p) (- (target ?p) ?duration)))))

       (when (at end (> ?duration (target ?p)))
         (at end
            (increase (total-cost)
              (+ (latePenalty ?p)
                  (* (latePenaltyRate ?p) (- ?duration (target ?p)))))))))
  )
)
```

Figure 8: PDDL Domain with Conditional Effects in the Airplane Landing Problem. The literal `takeOff` is a special proposition manipulated by a dummy action to force all the landing actions to be anchored to the same point in time.

governed by timed initial literals that affect whether the conditions are satisfied or not and the other is the case where the satisfaction of the conditions is determined by the status of continuous effects controlled by the actions in the plan (so, for example, the cost of an action might depend on whether a continuously changing value has passed some threshold or not at the time the action is executed). Both of these cases can be handled by a straightforward encoding of the linkage between the value of the 0-1 condition variable and the corresponding conditions (the details are given in Appendix D). We have also extended the conditional effects to allow them to affect the `?duration` variable in an action, with similar devices for encoding this in the MILP.

The MILP can then be solved as a single and final step in the construction of the plan, optimising the plan metric quality by rescheduling the actions to best exploit the precise timing of the actions and their interaction, through these limited conditional effects, on the plan quality.

## 11. Continuous Linear Benchmark Domains

As COLIN is one of the first planners to support PDDL2.1 models featuring continuous linear change and duration-dependent effects[8] there are currently no benchmarks available that exploit these fea-tures. To support our evaluation and to foster future comparisons between planners designed to solve these problems, we have produced a number of domains with these features[9].

---

8. Specifically, duration-dependent effects that depend on non-fixed durations.

The first of our domains is an extension of the 'Metric Time' variant of the Rovers domain, from the 2002 International Planning Competition (IPC 2002) (Long & Fox, 2003b). Our focus here is on the action `navigate`, responsible for moving a rover from one location to another. In the original model, it has a discrete effect, at the start of the action, to decrease the energy level of the rover by 8 units, coupled with a precondition that there must be at least 8 units of energy available. We replace this with a continuous numeric effect on energy, and an `over all` condition that energy must be at least zero during the action. As the duration of the original action was specified as 5, we use an effect with gradient $-8/5$. Written thus, the action has the same net effect and conditions: energy is decreased by 8 units, and must not become negative. This continuous change models more accurately the use of power during the navigate action: whilst power use may not actually be linear, it is closer to linear than it is to being instantaneous. To make the model still more realistic, we introduce a new action to the domain: `journey-recharge`, shown in Figure 9. By exploiting interaction between continuous numeric effects on the same variable, we use this action to capture the option of the rover tilting its solar panels to face the sun whilst navigating between two points. To account for the power use in reorienting the solar panels, at the start and end of the action, $0.2$ units of energy are used. The benefit for this consumption is that, whilst the action is executing, the energy of the rover is increased according to a constant positive gradient. For our final modification to the domain, we alter the duration constraint on the existing `recharge` action. In the original encoding, the constraint is:

```
(= ?duration (/ (- 80 (energy ?x)) (recharge-rate ?x))).
```

This forces the duration of the action to be sufficient to restore the level of charge to 80 (full capacity). In our new formulation, we replace the = with <= so that the duration constraint specifies the *maximum* duration for which the battery can be charged: it need not be restored to full capacity every time the action is applied. Following all three of these modifications, the domain can be used with the standard IPC 2002 benchmark problems. In addition to this, we have also created some problems considering just a single rover, where the issue of battery power management is of much greater importance.

The next of our domains is an extension of the 'Time' variant of the Satellite domain, again taken from IPC 2002. Here, in our continuous variant of the domain, we make three key changes to the domain model. First, in the original formulation, a proposition was used to indicate whether power was available to operate the instrumentation on a given satellite. Switching an instrument on required and then deleted this fact, and switching it off added it again. Thus, there was no scope for parallel power usage, and all instrumentation effectively used unit power. Now, we use a numeric variable to represent power, with preconditions and effects on this variable replacing the preconditions and effects on the proposition previously used. Second, exploiting the potential we now have for differing power requirements, instruments can be operated in one of two modes: cooled, or uncooled. In cooled mode, active sensor cooling is used to reduce sensor noise, enabling images to be taken in less time. This cooling, however, requires additional energy. Third, and finally, there is a compulsory 'sunrise' phase at the start of the plan, during which the satellites

---

9. PDDL domain and problem descriptions for all evaluation tasks are available in the online appendix maintained by JAIR for this paper.

```
(:durative-action journey-recharge
:parameters (?x - rover ?y - waypoint ?z - waypoint)
:duration (>= ?duration 0.2)
:condition (and (over all (moving ?x ?y ?z))
                (over all (<= (energy ?x) 80))
                (at start (>= (energy ?x) 0.2))
                (at end   (>= (energy ?x) 0.2))
           )
:effect (and (at start (decrease (energy ?x) 0.2))
             (increase (energy ?x) (* #t (recharge-rate ?x)))
             (at end   (decrease (energy ?x) 0.2))
        )
)
```

Figure 9: The `journey-recharge` action in the continuous-numeric Rovers domain

move from being shaded by the planet, to being in direct sunlight. This leads to an increase in power availability, modelled as a linear continuous numeric effect attached to an action, `sunrise`, that must be applied. Interaction between this effect and the preconditions on powering instruments ensures they can be operated no sooner than power is available. The problem files we use for this domain are slightly modified versions of the IPC competition problems, updated to define power availability as a numeric variable and to encode the power requirements of cooled and uncooled sensor operation. The problems in this domain have characteristics that are very similar to the Borrower problem we have used as a running example.

Further exploring the use of continuous numeric effects, our next domain models the operations of cooperating Autonomous Underwater Vehicles (AUVs). The AUVs move between waypoints underwater and can perform two sorts of science gathering operations. The first is taking a water sample from a given waypoint, which can be performed by any AUV in the appropriate location, and whose water sample chamber is empty. The second is taking an image of a target of interest. This requires two AUVs to cooperate: one to illuminate the target with a torch, and one to take an image of it. The AUV domain was inspired by the problem described by Maria Fox in her invited lecture at the 2009 International Conference on Automated Planning and Scheduling. Once data has been acquired, it must be communicated to a ship on the surface. As in the Satellite and Rovers domains, the AUVs are energy-constrained — they have finite battery power — and the power usage by actions is continuous throughout their execution. The more interesting continuous numeric aspects of the domain arise from the use of a model of drift. We introduce a variable to record how far each AUV has drifted from its nominal position, and update this in two ways. First, all activity in the plan is contained within an action `drift` with small, positive continuous numeric effect on the drifted distance. Second, we add a `localise` action that sets this drifted distance to zero, with its duration (and hence energy requirements) depending on the drifted distance prior to its application. This drifting then affects the other domain actions. In the simplest case, to sample water or take an image at a given location, an AUV cannot have drifted more than two metres, hence introducing the need to first localise if this is the case. More interestingly, for an AUV shining a torch, drifting affects how much light is falling on the target. Thus, the `shine-torch` action for an AUV `?v` has three effects on the amount of light falling on a given target `?t`:

- start: `(increase (light-level ?t) (- 1000 (distance-from-waypoint ?v)))`

- throughout: `(decrease (light-level ?t) (* #t (fall-off)))`

- end: decrease `(light-level ?t)` by any remaining contribution `?v` was making to its illumination.

The constant (`fall-off`) is pessimistically derived from formulæ involving the inverse-square law, giving a linear approximation of the decay in illumination levels due to drift. Then, for the `take-image` action itself, its duration is a function of (`light-level ?t`): the less light available, the longer it requires to take the image.

The final domain we use is the Airplane Landing domain (Dierks, 2005), first posed as a challenge by Kim Larsen in his invited lecture at the 2009 International Conference on Automated Planning and Scheduling. This problem models the scheduling of landing aircraft on an airport runway. For each plane, three landing times are specified: the earliest possible landing time, the latest possible landing time, and the target (desired) landing time. Since time must be allowed for airplanes to clear the runway once they have landed, and the use of the runway is a heavily subscribed resource, it is not possible for all planes to land at their ideal time. Planes can, therefore, land early or late, but doing so incurs a penalty. This penalty is modelled by a duration-dependent effect, as shown earlier in the paper (Figure 10.2 in Section 10). We have been able to construct a set of airplane landing problems using real data from the Edinburgh Airport arrivals board. Results from running Colin on these problems are reported in Section 12.

## 12. Evaluation

Colin is a temporal planner, able to solve problems with required concurrency, that can handle both discrete and continuous metric variables. The first question we address is how costly is the extension of the underlying Crikey3 system to allow Colin to manage continuous effects? Colin is a particularly powerful planner and there are no other general PDDL2.1 planners with similar expressive power available for comparison on the continuous problems. However, the extensions necessary to support continuous reasoning will add an overhead to the cost of solving problems where there are no continuous effects. We compare the performance of Colin with other temporal planners on a selection of temporal problems without continuous effects (Section 12.1) in order to evaluate how much overhead is paid by Colin in setting up and managing (redundant) structures, in comparison with state-of-the-art planners that do not pay this price.

We then move on to considering the performance of Colin on problems with continuous dynamics. Our second question is: how much improvement do we obtain from using the refined heuristic instead of the basic heuristic, when dealing with problems with continuous change? The planners discussed in Section 7 are not able to scale to large and complex problems, so we compare the two versions of Colin. We present their performances on new benchmark problems with continuous processes, setting the foundation for future comparative evaluation of alternative approaches to these problems.

The third question considered concerns the quality of the solutions produced by Colin, in comparison with optimal solutions where these can be found. Colin is a satisficing planner that can perform efficiently on a wide range of continuous planning problems, and we are interested in understanding how much solution quality must be sacrificed in order to obtain the efficiency achieved by Colin.

Finally, we consider the question: just how expensive is the move from solving an STP (sufficient for purely discrete temporal planning) to solving an LP (necessary for handling continuous effects)? In particular, is it practical to solve multiple LPs in performing heuristic state evaluations? Since LP construction and solution is central to the architecture of COLIN it is important that this can be relied upon to scale appropriately with the range and complexity of problems that COLIN is expected to solve.

The following experiments consider a large number of domains and domain variants. For the temporal comparisons we use the Simple Time and Time variants of Depots, Driverlog, Rovers, Satellite and Zeno, all from IPC 2002, and Airport and Pipes-No-Tankage from IPC 2004. The Airport variant used here is the Strips Temporal variant.

For the comparisons between the basic and refined heuristics on continuous domains, we use the new continuous benchmark domains introduced in Section 11: Airplane Landing, Rovers, Satellite Cooled (the Satellite variant with sensor cooling) and the AUV domain.

For the post-hoc optimisation experiments we use the Airplane Landing problem, the Cafe domain introduced in the empirical analysis of CRIKEY2 (Coles, Fox, Halsey et al., 2008), a variant of Airport in which the amount of fuel burned is to be minimised, and a version of Satellite with time windows, where rewards are obtained by scheduling observations into the tighter windows.

In all cases we use the competition benchmark sets of instances where available. For the continuous Rovers and Satellite domains we used the IPC 2002 Complex Time problem sets. These instances work with the continuous domain variants and it is possible to get better makespan plans for them, by respecting the continuous dynamics, than is possible when the same instances are solved using the discrete domain variants. We generated increasing sized instances for the Airplane Landing domain in which the number of planes to be landed increased (in the $n^{th}$ instance of the problem, $n$ planes must be landed). We wrote a problem generator for the AUV domain that increases the number of AUVs, waypoints and goals in the instances (they range from 2 AUVs, 4 waypoints and 1 goal, to 6 AUVs, 16 waypoints and 6 goals). All experiments were run on a 3.4GHz Pentium D machine, limited to 30 minutes and 1GB of memory.

## 12.1 Comparison with Existing Temporal Planners

Few temporal planners can actually solve a full range of temporal problems. As we have already observed, many temporal planners cannot solve problems with required concurrency. Even within the class of problems that have required concurrency, there are easier problems, which can be solved by a *left packing* of actions within the plan and harder ones for which this is not possible. By *left packing* we mean that actions that must be executed concurrently with other actions in the plan can be started at the same time as each other. This property means that the approach adopted in Sapa, of extending forward search to include a choice to either start a new action or else to advance time to the earliest point at which a currently executing action terminates, is sufficient to solve the problem. In contrast, a problem that cannot be left packed will require the possibility of advancing time to some intermediate point during execution of an action in order to coordinate the correct interleaving of other actions with it. We describe such problems as requiring *temporal coordination*. One of the few planners that can also handle problems requiring temporal coordination is LPG-s (Gerevini et al., 2010).

We therefore compare COLIN with LPG-td, LPG-s, Sapa and the temporal baseline planner developed for the temporal satisficing track at the 2008 International Planning Competition. Neither the temporal baseline planner, Sapa nor LPG-td can solve problems requiring any kind of temporal coordination. The temporal baseline planner compiles away temporal information, by using action compression, and solves problems as if they were non-temporal metric or propositional problems. When solutions are found, using Metric-FF as the core planning system, the temporal information is reintroduced by annotating the plan with suitable timestamps based on a critical path analysis. No details are published about this planner, but the source code and brief information are available from the IPC 2008 web site. This approach cannot therefore solve problems with required concurrency, but is fast and effective on simpler problems where the temporal actions can be sequenced. It is straightforward to identify many cases when action compression can be applied safely and this analysis is implemented in COLIN to reduce the overhead of reasoning with action end points where it is unnecessary. Therefore, the behaviour of the temporal baseline planner is similar to that of COLIN when all actions can be safely compressed. In Figures 10, 11 and 12 we show CPU time comparisons between COLIN and the best performances of Sapa, LPG-td, LPG-s and the temporal baseline planner, across a wide and representative collection of temporal benchmark domains. Figure 10 shows performance on simple temporal problems, where action durations are all fixed, while Figures 11 and 12 show results for more complex temporal problems, including those where the duration of actions is determined by the context in which they are executed (although none in which action effects depend on this), and problems with metric variables. None of these problems feature required concurrency or other forms of temporal coordination. In these figures, planners not appearing in a dataset were not the best on any problems in that domain.

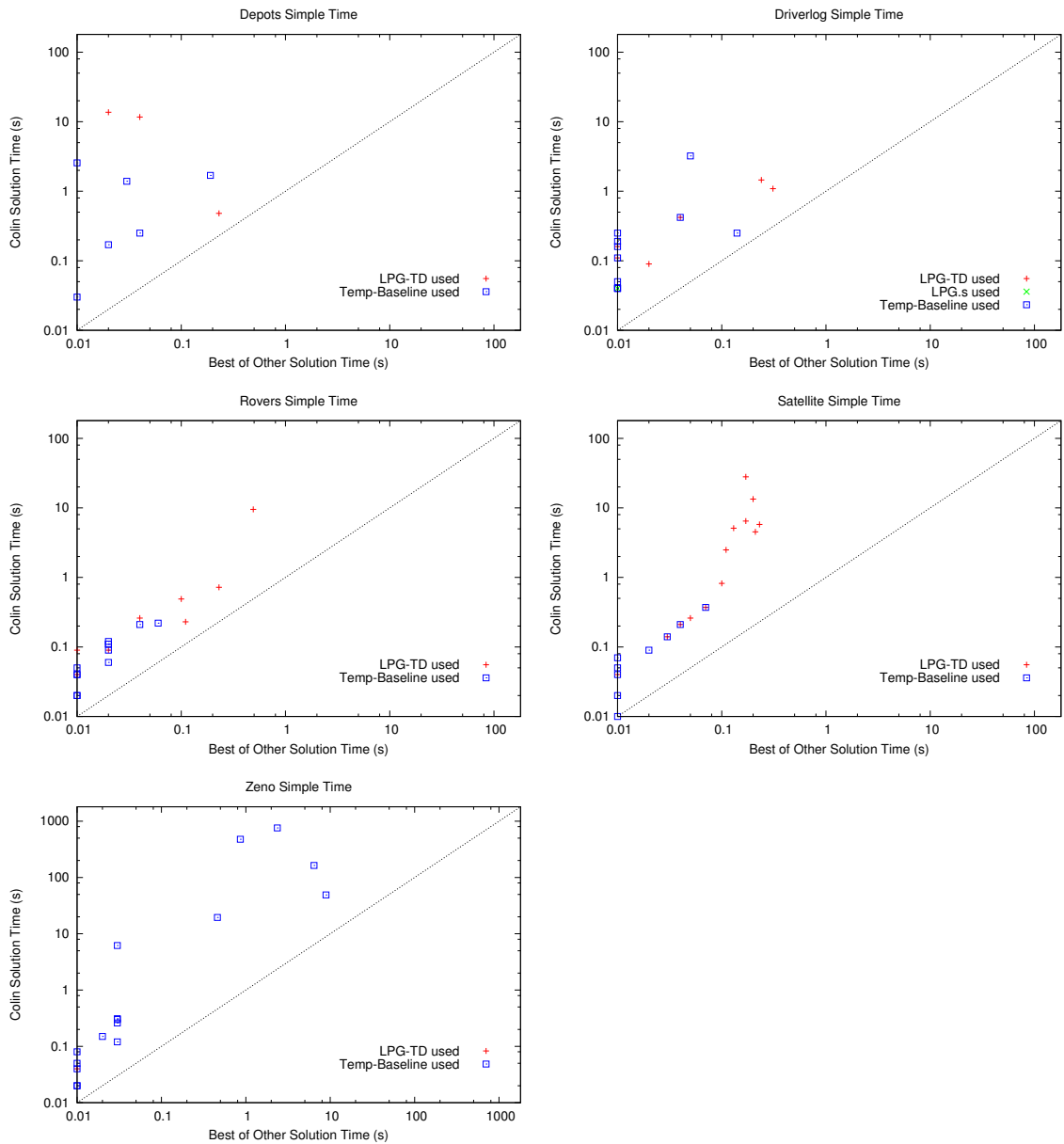

Figure 10: Comparison of time taken to solve problems in simple temporal planning benchmarks. COLIN is compared to the best of LPG-td, LPG.s, Sapa and a temporal baseline planner, on each problem file — the shape and colour of the points indicate which planner was the best and was therefore used in the plot. Planners not appearing in a particular dataset were not the best on any of the problems in that collection.

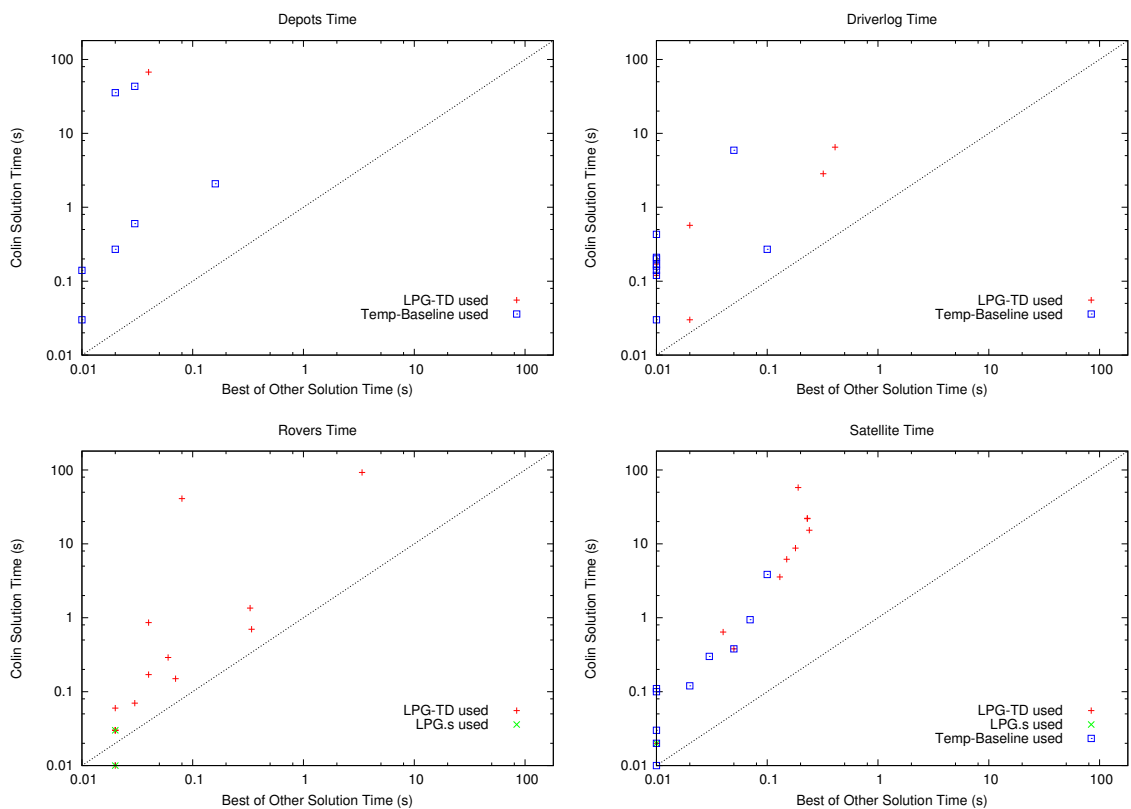

Figure 11: Comparison of time taken to solve problems in more complex temporal planning benchmarks (first set). COLIN is compared to the best of LPG-td, LPG.s, Sapa and a temporal baseline planner, on each problem file — the shape and colour of the points indicate which was the best. Planners not appearing in a particular dataset were not the best on any of the problems in that collection.

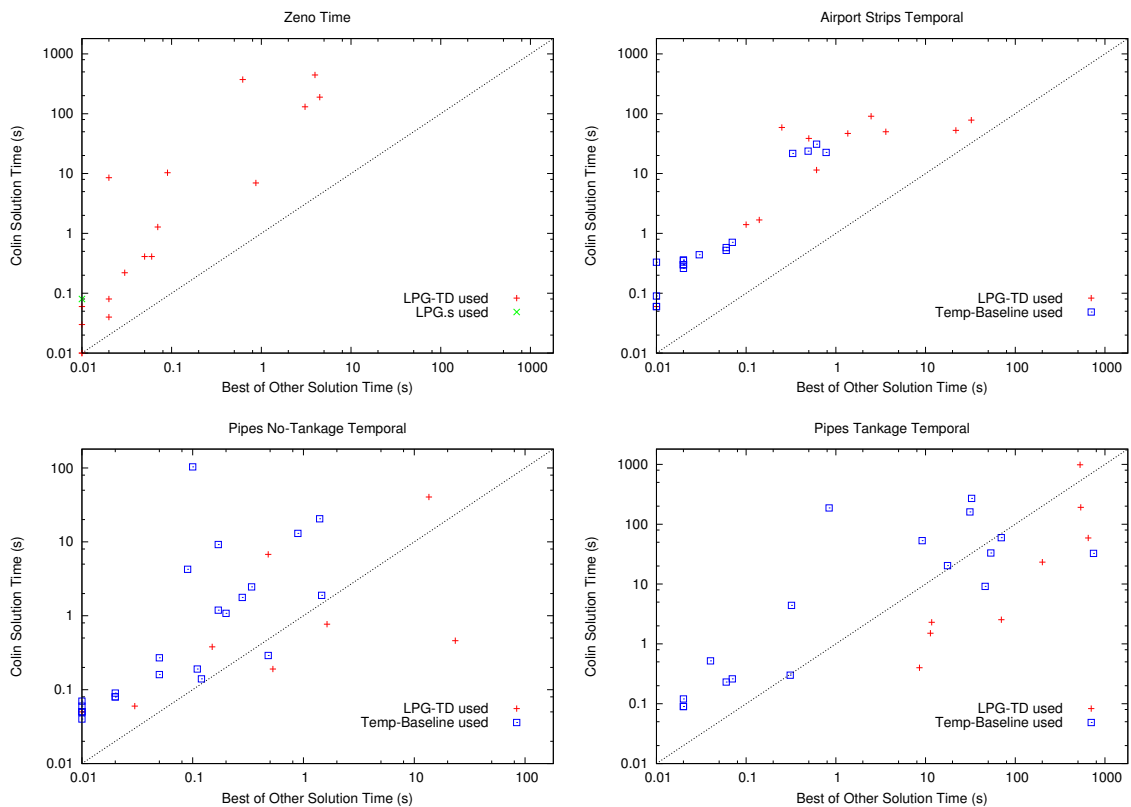

Figure 12: Comparison of time taken to solve problems in more complex temporal planning benchmarks (second set). COLIN is compared to the best of LPG-td, LPG.s, Sapa and a temporal baseline planner, on each problem file — the shape and colour of the points indicate which was the best. Planners not appearing in a particular dataset were not the best on any of the problems in that collection.

Analysis of Figures 10–12 shows that COLIN does indeed pay an overhead in computation time in the solution of temporal problems that do not feature continuous dynamics. The overhead is particularly significant in the simple temporal problems where there is no interesting temporal structure and the temporal baseline planner tends to perform very well. The overhead paid by COLIN is lower in the complex temporal problems, where the temporal reasoning required is sometimes more challenging. The makespan results in Figures 13, 14 and 15 show that COLIN produces good quality plans, especially for the complex temporal problems, although the temporal baseline planner is still competitive in terms of both CPU time and makespan. This suggests that the temporal structure, even in the complex temporal benchmarks, is quite simple and that a planner can do well by ignoring the temporal structure that is present, rather than trying to reason about it in generating plans.

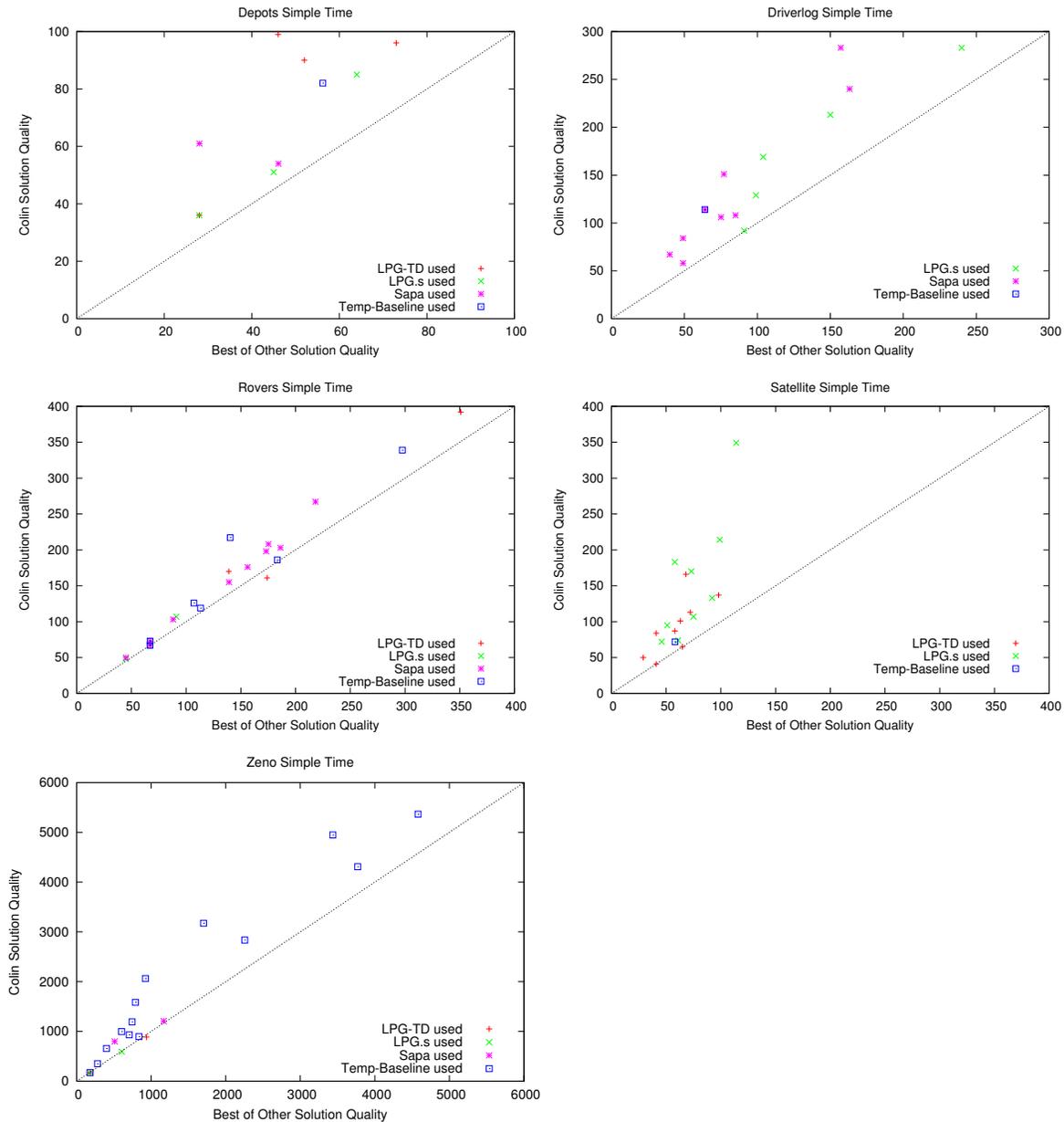

Figure 13: Comparison of plan quality in simple temporal planning benchmarks. COLIN is compared to the best of LPG-td, LPG.s, Sapa and a temporal baseline planner, on each problem file — the shape and colour of the points indicate which was the best. Planners not appearing in a particular dataset were not the best on any of the problems in that collection.

The detailed results of these experiments, showing raw runtime and quality comparisons between the planners used in this experiment, are presented in Appendix E.

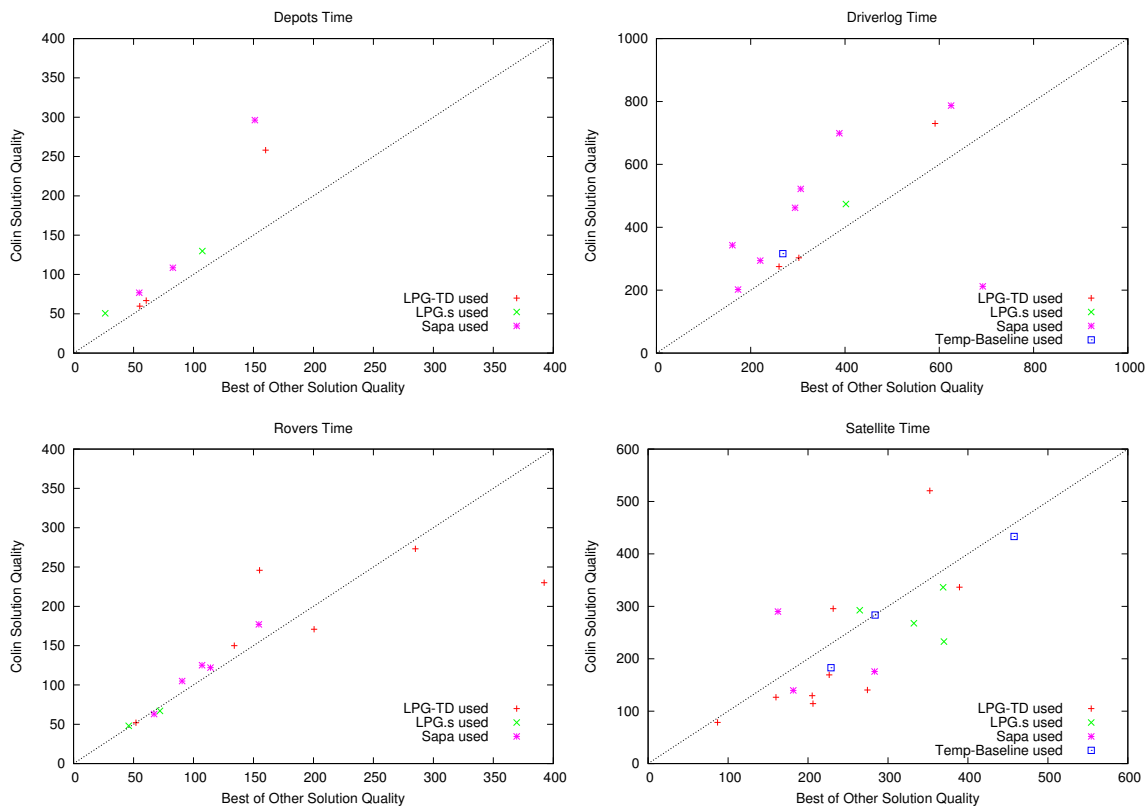

Figure 14: Comparison of plan quality in more complex temporal planning benchmarks (first set). COLIN is compared to the best of LPG-td, LPG.s, Sapa and a temporal baseline planner, on each problem file — the shape and colour of the points indicate which was the best. Planners not appearing in a particular dataset were not the best on any of the problems in that collection.

## 12.2 Solving Problems with Continuous Linear Change and Duration-Dependent Effects

Our focus in this section is on examining the scalability of COLIN on the continuous benchmark domains we have developed and, specifically, on comparing the two variants of the TRPG discussed in Section 9. These are: the basic heuristic, which discretises time, and the refined heuristic, which is capable of handling continuous numeric change directly. The continuous benchmarks, as described in Section 11, are characterised by sophisticated temporal structure (including required concurrency) giving rise to interesting opportunities for concurrent behaviour. Because these problems have time-dependent effects and continuous effects, they are out of the reach of the temporal planners used in the last experiment. The problems used for this experiment are designed to rely on the exploitation of these features, so a baseline planner that ignored these continuous dynamics would be unable to solve the problems.

Results comparing the basic and refined heuristics are shown in Figure 16. Beginning with the Airplane Landing domain and the Rovers domain variant, the performance is the same when either heuristic is used: the relaxed plans found are the same. This is to be expected, because in these two domains the interaction between time and numbers is relatively limited. In the Airplane Landing problem, action durations affect a variable used to measure plan cost but that is not used in any preconditions. Thus, the selection of actions in the TRPG is unaffected. In the Rovers domain, continuous change arises when consuming power during navigate actions, or producing power when recharging. Capturing the time-dependent nature of these more precisely has no effect on the relaxed plans, as the nature of the relaxation leads it to only rarely require recharge actions, and the conditions under which these are needed are not affected by whether the effects are integrated or not. Nevertheless, these two domains illustrate that in guaranteed 'like-for-like' situations, where the heuristic guidance will be the same, the refined heuristic is only negligibly more expensive to compute, despite the additional overheads of tracking gradient effects as the TRPG is expanded. It can also be seen that COLIN scales well across the Airplane Landing instances, although it only manages to solve 9 of the 14 Rovers problems (these well within two minutes).

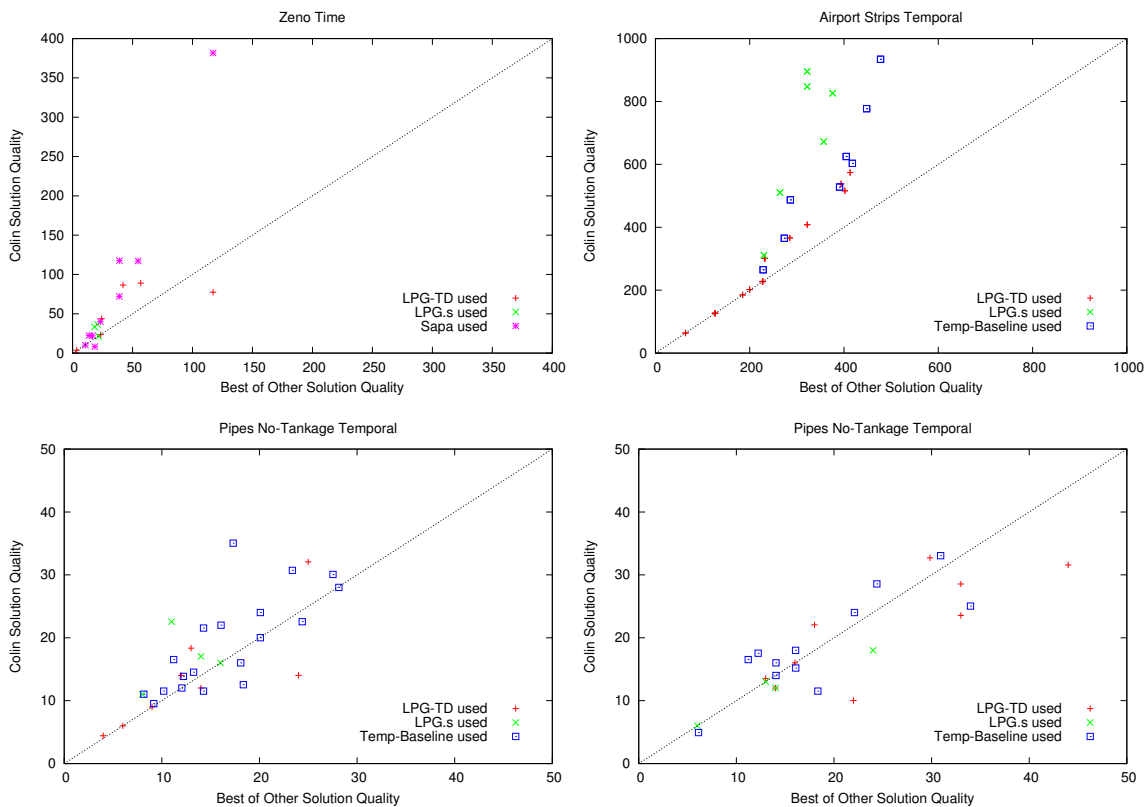

Figure 15: Comparison of plan quality in more complex temporal planning benchmarks (second set). COLIN is compared to the best of LPG-td, LPG.s, Sapa and a temporal baseline planner, on each problem file — the shape and colour of the points indicate which was the best. Planners not appearing in a particular dataset were not the best on any of the problems in that collection.

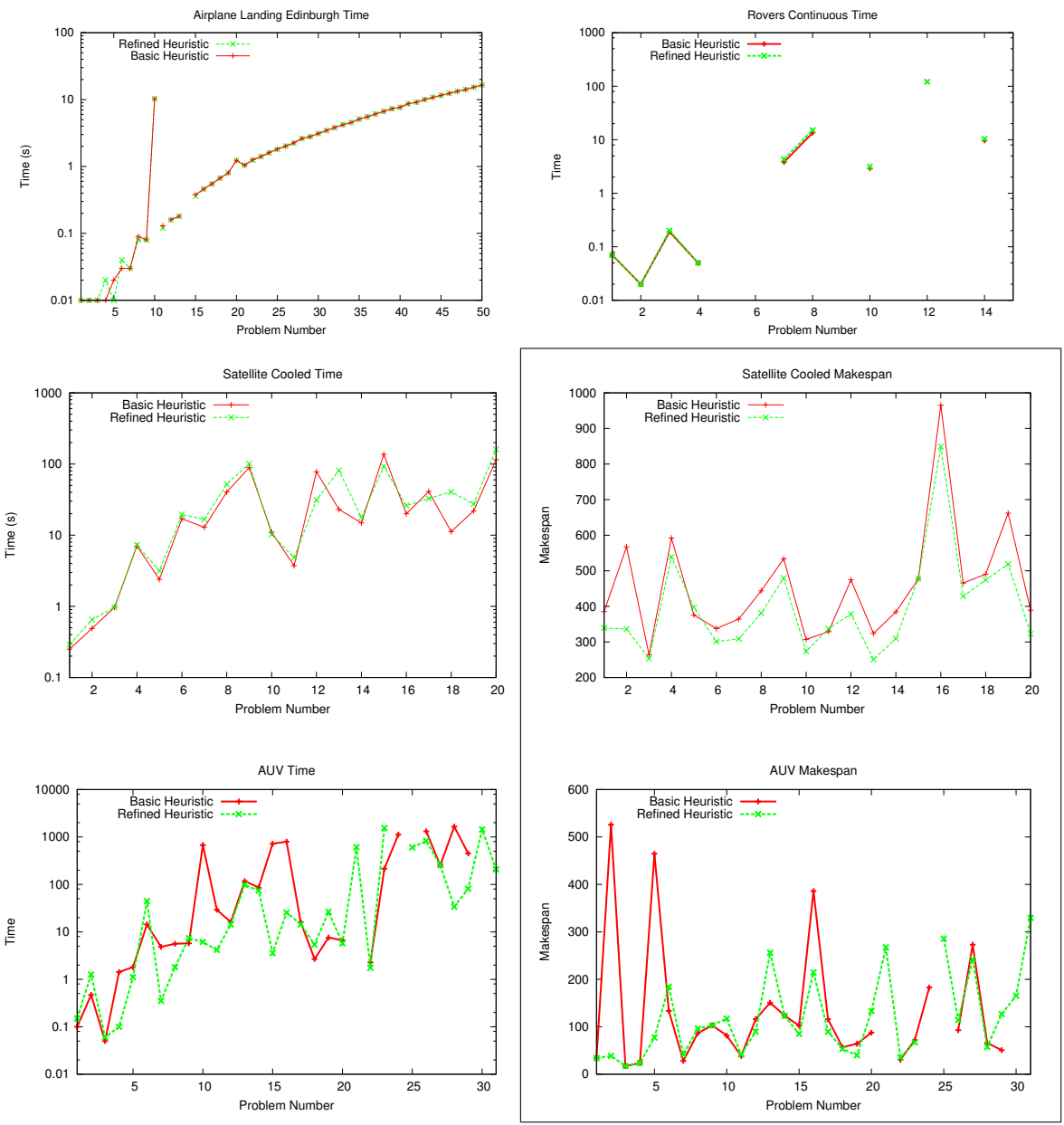

Figure 16: Comparison of the Basic and Refined TRPG variants on continuous domains. The boxed graphs are makespan comparisons for the Satellite Cooled and AUV domains, placed to the right of their corresponding runtime graphs.

In the Satellite 'Cooled' domain, the runtime taken to find the plans when using the refined heuristic is comparable to that when using the basic heuristic: in some problems (e.g. 13, 18) it is slower; but in others (e.g. 12, 15) it is faster. The more interesting comparison to make is in the makespan data (shown to the right). As can be seen, the refined heuristic generally produces better quality plans. The difference in quality is due to the refined heuristic better capturing the relationship between time and numbers, leading to better actions being chosen in the relaxed plan. By way of example, consider the state reached after beginning the sunrise action:

- For the basic heuristic, the LP is used to obtain bounds on the power availability in this state, with free reign over how much time to allow to elapse. The lower-bound found is slightly more than zero (corresponding to allowing $\epsilon$ time to elapse), and the upper-bound found is the peak power availability (corresponding to applying the entirety of the sunrise action). When building a TRPG from these bounds, cooled sensor operation is immediately available, and hence the goals will always be achieved first by actions using sensor cooling: the duration of such actions is lower, making them more attractive. The resulting relaxed plan, and hence helpful actions, will therefore lead search to use sensor cooling.

- With the refined heuristic, the LP is used to obtain bounds on the power availability in this state, but these bounds must be obtained at the soonest possible point. Thus, the lower-bound is still slightly more than zero, but the upper bound is also only slightly more than zero. The positive gradients in effect on the power availability variables are then included in the TRPG, influencing the layers at which different actions become applicable. Specifically, actions without sensor cooling have lower power requirements, and hence appear at earlier layers. Then, for goals first achieved by actions not using sensor cooling (where the increased duration of acquiring the image without cooling is compensated for sufficiently by being able to start taking the image sooner) the relaxed plan, and hence helpful actions, will not use sensor cooling for these goals. It can be seen that this situation is closely analogous to the differences in alternative mortgages in the Borrower domain.

The extent to which this trade-off influences plan quality varies between problems, depending on the initial orientation of the satellites, and the images required. The least benefit arises if a satellite requires substantial reorientation to point it towards its first target — if this is the case, the time taken allows the energy level to rise sufficiently to support sensor cooling. The greatest benefit arises in the opposite situation, where a satellite requires minimal reorientation — then, switching on a sensor in its cooled mode will require a substantial amount of time to elapse to support its energy requirement precondition.

To aid understanding of the scalability implications of these results, the Satellite problems are based on those used in the 2002 IPC, so are of a similar fundamental size. However, the continuous reasoning that has been added to them makes the same underlying problems fundamentally much more difficult to solve.

In the AUV domain, the use of the refined heuristic increases the problem coverage, with 30 problems solved rather than 27. Applying the Wilcoxon Matched-Pairs Signed-Ranks Test to the paired time-taken data for mutually solved problems, we find that we can reject the null hypothesis that the refined heuristic is no better than the basic heuristic, with $p \leq 0.05$. Observing the performance of the planner, this difference in performance arises due to the way in which the drifting process is handled by the two approaches. Specifically, it is accounted for by the difference in how the bounds for fact layer zero of the TRPG are calculated. Consider a state in which an action for an AUV to communicate image data has just been started. The domain encoding ensures that until communication has completed, the AUV cannot perform any other activities. At this point, prior to evaluating the state using a TRPG heuristic, the LP is used to give bounds on the values of each state variable. Considering just the variable recording how far the communicating AUV has drifted — the variable (`distance-from-waypoint auv0`), from now on abbreviated to `dfw0`:

- The basic heuristic employs the approach set out in Section 8.3. A single 'now' timestamp variable is introduced, that must come after the action just started, along with an additional variable and constraint for `dfw0`. Maximising and minimising the value of this additional variable yields bounds on `dfw0`. The lower bound will be infinitesimally larger than it was prior to starting the action, due to $\epsilon$ time having elapsed. The upper bound corresponds to allowing a large amount of time to elapse.

- The refined heuristic employs the approach set out in Section 9.2.1. Here, a 'now' timestamp variable is introduced for each task variable, in this case we are concerned with $tnow(\texttt{dfw0})$. As in the prior case, this is constrained to be after the action just applied. Additionally, however, because the domain model enforces that no other action can refer to the value of the variable until the communicate action has finished, this specific $tnow$ must also come after the (future) end of the action just applied. The bounds on `dfw0` are then found by following the remaining steps of Section 9.2.1: the LP is solved to minimise the value of this $tnow$ variable, the value of the variable is fixed to this minimum and then the LP is solved to maximise and minimise the value of `dfw0`. Critically, because this $tnow$ variable must come after the end of the action just applied, rather than just after its start, the lower bound on `dfw0` is larger.

The increase in the lower bound on `dfw0` then affects whether, in the TRPG, preconditions of the form (`<= (dfw0) c`) are considered satisfied in the initial fact layer. If they are not satisfied, they are delayed until the earliest layer at which a localise action reduces the value of `dfw0`. This difference can then affect the relaxed plan found: during solution extraction, if an action $A$ requiring (`<= (dfw0) c`) is chosen, then if a localise action was necessary to achieve this in the TRPG, the action will be added to the relaxed plan. As $A$ cannot come any earlier than after the end of the communicate action just applied, that is, the point at which the bounds on `dfw0` are calculated, then some sort of localisation is necessary if $A$ is ultimately to be applied. Thus, the bounds for the refined heuristic here lead to better relaxed plans being found, containing localise actions that would otherwise be omitted.

To give an indication of the difficulty of these problems, the AUV problems range from problems with 2 AUVs, 5 waypoints, 2 objectives and 2 goals to those at the harder end with 6 AUVs, 15 waypoints, 6 objectives and 7 goals. The major hurdle preventing COLIN from scaling to even larger problems is the inability to see that an implicit deadline has been created when a shine-torch action is started. The AUV shining the torch has finite energy, so if the planner starts a shine-torch action with one AUV, in preparation for another AUV to take an image, but then adds to the plan some actions involving the second AUV that are unrelated to taking the image, the delay can lead to there being insufficient energy to shine the torch for long enough to gain the required exposure when the photograph taking action is eventually started. This leads the planner to a dead end and it is forced to resort to best-first search, which is much less effective than EHC in this domain. Such implicit deadlines can occur in many planning problems with temporal coordination and the issues COLIN faces could be avoided by using a branch-ordering heuristic that promotes actions whose applicability is time-limited due to the ends of currently-executing actions, or perhaps through relaxing unnecessary ordering constraints imposed by COLIN due to total order search. Both of these are out of the scope of this paper, but are interesting avenues for future work.

```
(:durative-action burning-fuel
:parameters (?a - airplane)
:duration    (>= ?duration (* 60 (engines ?a)))
:condition   (and (at start (not-burning-fuel ?a))
                  (at end (taking-off ?a)))
:effect (and (at start (can-start-engines ?a))
             (at start (not (not-burning-fuel ?a)))
             (increase (wasted-fuel) (* #t (engines ?a)))
         )
)
```

Figure 17: `burning-fuel` action added to the Airport domain

### 12.3 *Post Hoc* Plan Optimisation

In this section we evaluate the effectiveness of our *post hoc* plan optimisation strategy. As described in Section 10, the plan optimisation phase occurs after planning is complete and can never change the actions that are in the plan. By lifting a Partial Order prior to scheduling (Veloso, Pérez, & Carbonell, 1990), we can provide the scheduler with a little more flexibility over the order of actions. As long as the ordering constraints remaining after this (greedy) partial-order lifting are respected, the scheduler can reduce plan cost by altering the time-points at which the actions occur and, where possible, their durations. Minimising an objective other than plan makespan can only have an effect on plan quality in domains where the metric is sensitive to the times at which the actions are applied, since, by default, COLIN minimises makespan in the solution of the final LP for a completed plan. There are few such benchmark domains in the literature, so we make use of the one existing suitable domain and introduce some new variations on existing benchmarks, in order to test this feature.

The first domain, and the only existing domain with this property, is the Airplane Landing domain, used earlier in this section, and described in Section 11. Here, the penalties incurred for each landing depend on whether, and to what extent, it is early or late. Therefore, for a given sequence of landings, the times assigned to them has an impact on the quality of the plan.

The next two of our benchmark problems are variants of problems introduced in the International Planning Competitions of 2002 (Long & Fox, 2003b) and 2004 (Hoffmann & Edelkamp, 2005). First, we consider a modified version of the Satellite domain. We modify the domain by adding time windows (modelled using TILs) during which there is a clear view of a given objective. If the photograph of the objective is taken during such a time window, the quality of the plan improves, as a better quality picture is preferable. In each problem we introduce three such time windows for each objective, of bounded random duration, during which taking a photograph of the objective is preferred. The second adapted benchmark is taken from the IPC2004 Airport domain. Where the Airplane Landing problem described previously is concerned with scheduling landing times for aircraft, the Airport domain is concerned with coordinating the ground traffic: moving planes from gates to runways, and eventually to take-off, whilst respecting the physical separation that must be maintained between aircraft for safety reasons. We add to this domain the metric to minimise the total amount of fuel burnt between an aircraft's engines starting up and when it eventually takes off. To capture this in PDDL2.1, we add the action shown in Figure 17. This action must occur before a plane's engines can be started and cannot then finish until the plane has started to take-off (hence its duration is at least that of the `startup` action). Between these two points it increases the amount of fuel wasted by a rate proportional to the number of engines fitted to the aircraft: larger planes (for which the number of engines is greater) waste more fuel per unit of time. In both the Satellite and the Airport domains we use the standard problem sets from the competitions, adding any minor changes needed to support the modifications made, whilst leaving the underlying problems themselves unaltered.

The final domain we consider is the café domain, first used to evaluate CRIKEY (Coles, Fox, Halsey et al., 2008). In this domain, tea and toast must be made and delivered to each table in a café. The kitchen, however, has only one plug socket, preventing the two items from being made concurrently. This restriction allows the problem to have a number of interesting metric functions: to minimise the total time to serve all customers (the plan makespan), to minimise the time between delivery of tea and toast to a given table, or to minimise the amount by which the items have cooled when each is delivered to the table. We consider the latter two variants here.

The results of our experiments are presented in Figure 18. Starting in the top-left, with the Airplane Landing domain, *post hoc* optimisation gives only a modest improvement in plan quality. This is due to the limited scope for optimisation: even after partial-order lifting, the order in which the planes are going to land is fixed by the plan, so all that can be adjusted is the precise times at which the planes are going to land within that ordering.

Moving to the Airport domain variant with the 'burning-fuel' action — Figure 18 top-right — *post hoc* scheduling is able to give large improvements in plan quality. In the original plans, before optimisation, the burning-fuel action for a given plane can be started at any point prior to when the relevant `can-start-engines` fact is needed and can be ended at any point after the relevant `taking-off` fact is true, so not necessarily in a timely manner. Following *post hoc* optimisation, due to the objective function used, each burning-fuel action starts as late as possible and finishes as early as possible.

In the café domain, the results for the two metrics used are shown in the central graphs in Figure 18. The two diagonal lines correspond to the original plans. On a given problem, the two plans are identical: only the evaluation metric differs. The two lower lines show the quality of the plan after scheduling it with respect to the relevant metric. Observing the post-scheduled plans, the actions are scheduled as one would intuitively expect. When minimising the total delivery window times, the items for a given table are delivered in succession, even if the first item loses heat while waiting for the second item to be prepared. In contrast, when minimising heat loss items are delivered to tables as soon as they have been prepared, even if there is then a delay between the two items being delivered.

Finally, the results for the variant of the Satellite domain with observation windows is shown in the bottom-left of Figure 18. Whilst not as marked as the improvements in the previous two domains, the scheduler is able to make some headway in better scheduling the observations. The original plan for a given problem will, for each satellite, fix the observations it is to make, and the order in which they are to be made. There remains enough flexibility to be able to improve plan quality, reducing plan cost by around a factor of 2.

### 12.4 Comparison with Optimal Solutions

We investigated the difference in quality between optimal solutions and the solutions produced by COLIN in order to form an impression of how close to optimal COLIN can get. To do this, we ran COLIN with an admissible heuristic that uses the makespan estimate produced by the TRPG, using the same value for $\epsilon$ as is used by COLIN in the results presented in Figures 16 and 18. We call this variant optimalCOLIN.

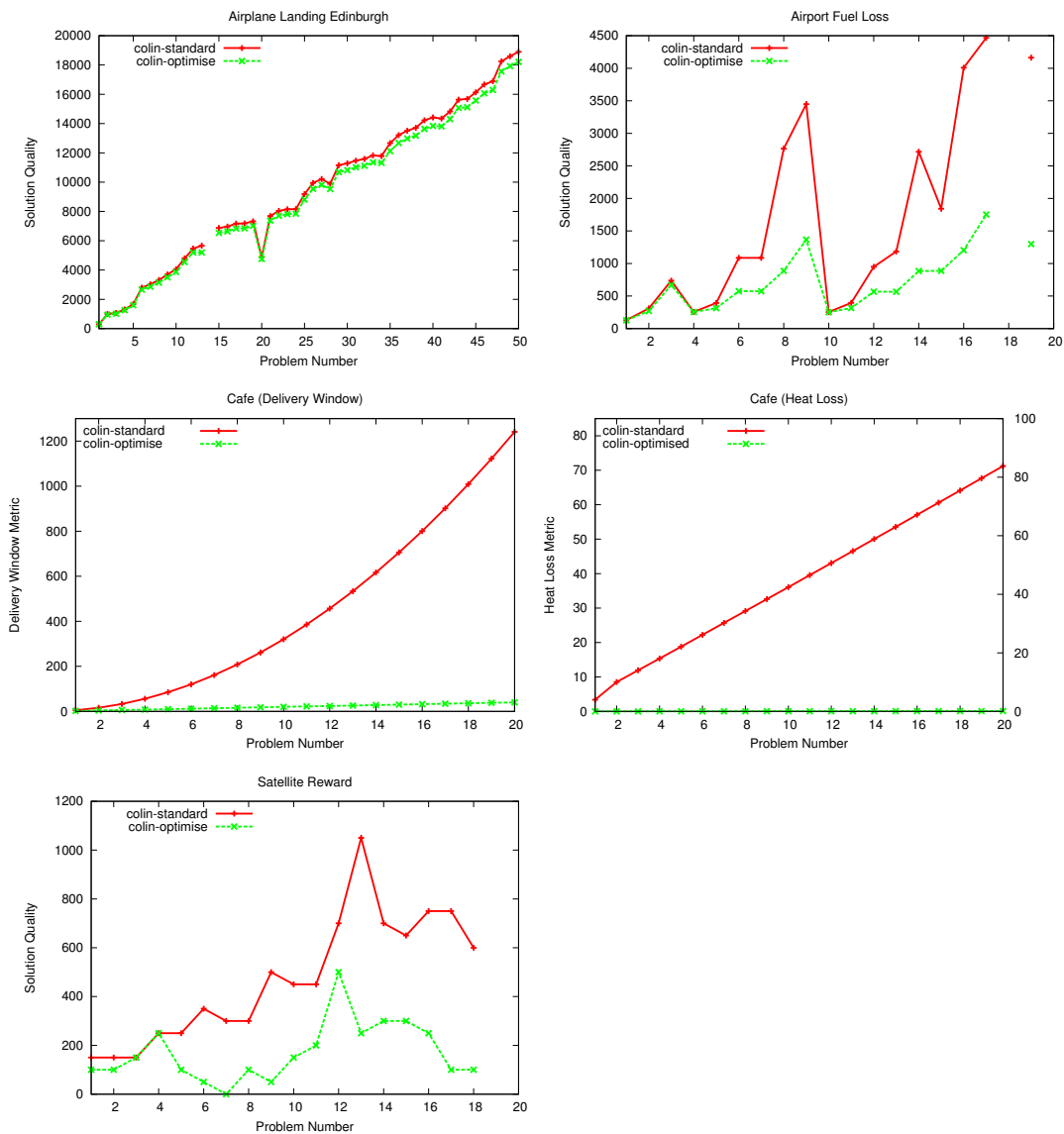

Figure 18: Quality of plans produced by Colin with and without *post hoc* optimisation. On all four graphs, lower is better.

In the AUV and Rover domains, there are variable-duration actions in the domain for which the durations can be chosen to be as small as $\epsilon$ when these actions are used in a plan. $\epsilon$-length actions might be chosen, for example, to relocalise having slightly drifted, or to recharge having used a negligible amount of power. In these domains, optimal search has to consider plans comprising almost entirely actions of $\epsilon$ duration up to the optimal makespan. As an example of the scale of this, on AUV problem 1, solved by COLIN, we find a plan with makespan 34.031. Careful analysis by hand suggests that this plan cannot be improved, so is optimal. OptimalCOLIN must consider plans of up to 34,031 steps in order to prove that this plan is optimal. This means that this problem is completely out of the reach of optimal planning.

| Problem | optimalCOLIN | | COLIN | |
|---------|----------|-------------|----------|-------------|
| [2-5] instance | makespan | time (secs) | makespan | time (secs) |
| 01 | 20.001 | 0.00 | 20.001 | 0.01 |
| 02 | 30.001 | 0.00 | 34.004 | 0.01 |
| 03 | 30.001 | 0.02 | 38.007 | 0.01 |
| 04 | 40.001 | 0.29 | 44.006 | 0.01 |
| 05 | 40.003 | 3.94 | 48.009 | 0.02 |
| 06 | 50.003 | 69.93 | 62.01 | 0.03 |
| 07 | - | - | 66.014 | 0.03 |

Table 4: Comparison of makespans and solution time for airplane landing problems solved by optimalCOLIN and COLIN using the refined heuristic. Problem 7 could not be solved by optimalCOLIN within the 1 hour bound.

A similar problem arises in the Rovers domain, where the recharge action can be as little as $\epsilon$ long, and a series of $\epsilon$-long recharge actions can be applied, reaching ostensibly different states, but without making any progress. Clearly, the potential for $\epsilon$-duration actions can arise in any continuous temporal domain. The same problem of search-space explosion will also arise in any temporal domain where there are orders of magnitude differences between the longest and shortest possible actions.

However, in the Airplane-Landing and Satellite Cooling domains, there are no variable-duration actions in these domains that can be made arbitrarily short during search. Therefore, optimalCOLIN is in principle able to solve problems in these domains. In fact, given 4 Gb of memory and 1 hour of runtime for each instance, it was able to solve 6 airplane landing instances, as shown in Table 4. As the table shows, the time required to solve these problems increases very fast: problem 5 could be solved in 3.94 seconds, problem 6 in 69.93 seconds, and problem 7 could not be solved within the hour available. On this basis we decided it was unnecessary to extend the time available to optimalCOLIN as it would be unlikely to cope with large instances.

Table 4 shows that COLIN sacrifices optimality for speed. This sacrifice is important, but it does pay off in terms of time required to solve problems. COLIN is able to solve 62 of the airplane landing problems, with no instance taking more than 33.02 seconds to solve.

We found that optimalCOLIN could report a candidate solution to the first Satellite domain instance, within 368 seconds. However, it could not prove within the time available that this solution was optimal, so we did not include it.

## 12.5 Costs Associated with LP Scheduling

In the transition from CRIKEY3 to COLIN we switch from solving an STP at each state to solving an LP. An important issue to consider is the impact that this has on the time taken to evaluate the feasibility of the plan constructed to reach every state considered in the search. In its default mode of operation, COLIN uses an STP to evaluate a state unless it has temporal–numeric constructs that necessitate use of an LP. To evaluate whether or not this is appropriate (or whether always using an LP would be faster), and to compare the overheads of STP solving with LP solving on equivalent problems, we created a variant of COLIN that, at every state $S$, schedules the plan to reach $S$ independently using three different schedulers: the original STP solver used in the standard version of COLIN, the equivalent LP solved using CPLEX (IBM ILOG CPLEX Optimization Studio) and the equivalent LP solved using CLP (Lougee-Heimer, 2003). The STP solver used is the incremental STP algorithm due to Cesta and Oddi (1996), as previously used in CRIKEY3. Each of the LP solvers is used with the tighter variable bounds described in Section 8.4. In order to evaluate the cost associated with use of an LP instead of an STP, we modified COLIN to collect data revealing the costs for each technique applied at each node evaluated during the search for a plan. It is not possible to compare the performance straightforwardly, simply by running COLIN using an STP versus COLIN with an LP, because minor variations caused by numerical accuracy can lead to very different trajectories being followed, masking the intended comparison. As an aside, it is interesting to observe that minor (and essentially uncontrollable) differences in computed makespans for relaxed plans can lead to significant variations in performance (relaxed plans with equal $h$-values are sorted by makespan estimates for search).

As we wish to compare the STP and LP approaches, it is necessary to consider domains with which both can reason: that is, those without continuous-numeric or duration-dependent effects. In order to consider some problems for which scheduling is interesting and necessary (in contrast to the temporally simple problems of Section 12.1) we consider domains with required concurrency. Currently very few such benchmarks exist, as few planners attempt to solve such problems. We use representatives of the only competition domains with such features: the compiled 'timed initial literal' domains from IPC2004, from which we use Airport (with Time Windows) and PipesNo-Tankage (with deadlines). We also use the Match-Lift and Driverlog Shift domains (Halsey, 2005). For completeness, we include results for a domain in which the scheduler is not strictly necessary: the PipesNoTankage Temporal domain from IPC2004.

Figure 19 shows the mean time spent scheduling per state, using each approach, on the problems from the above domains. We exclude from the graph data from any problems that were solved by the planner in less than a second, as the accuracy of the profiling data is not sufficiently reliable to measure the time spent in each scheduler when the overall time taken is small. Since there was no interesting variation in results between domains we present all data together across three graphs, sorted by scheduling time per node when using CPLEX. This is intended as a nominal analogue for how hard the scheduling problems in the given planning problem are. An increase in the scheduling time for CPLEX generally corresponds to an increase in the scheduling time for CLP and the STP solver, except on the easier problems where noise can be sufficient to tip the balance as the figures are small. Note the differing y-axis scales on the three graphs, sorting problems according to difficulty allows us to display the data with an appropriate y range to distinguish the results. For the sake of maintaining reasonable y-axis ranges the final problem, problem 60, has been omitted from the graphs; on this problem the figures were CPLEX 239ms, CLP 139ms and STP 38ms.

The results in Figure 19 are, of course, not indicative of the scalability of COLIN, as it is running three schedulers at each state, so is significantly slower than in its usual configuration. In practice, in domains such as these with no continuous or duration dependent effects, COLIN will automatically disable the LP scheduler and use the more efficient STP solver. Further, the planner is here being run with profiling enabled, so is subject to significant overheads.

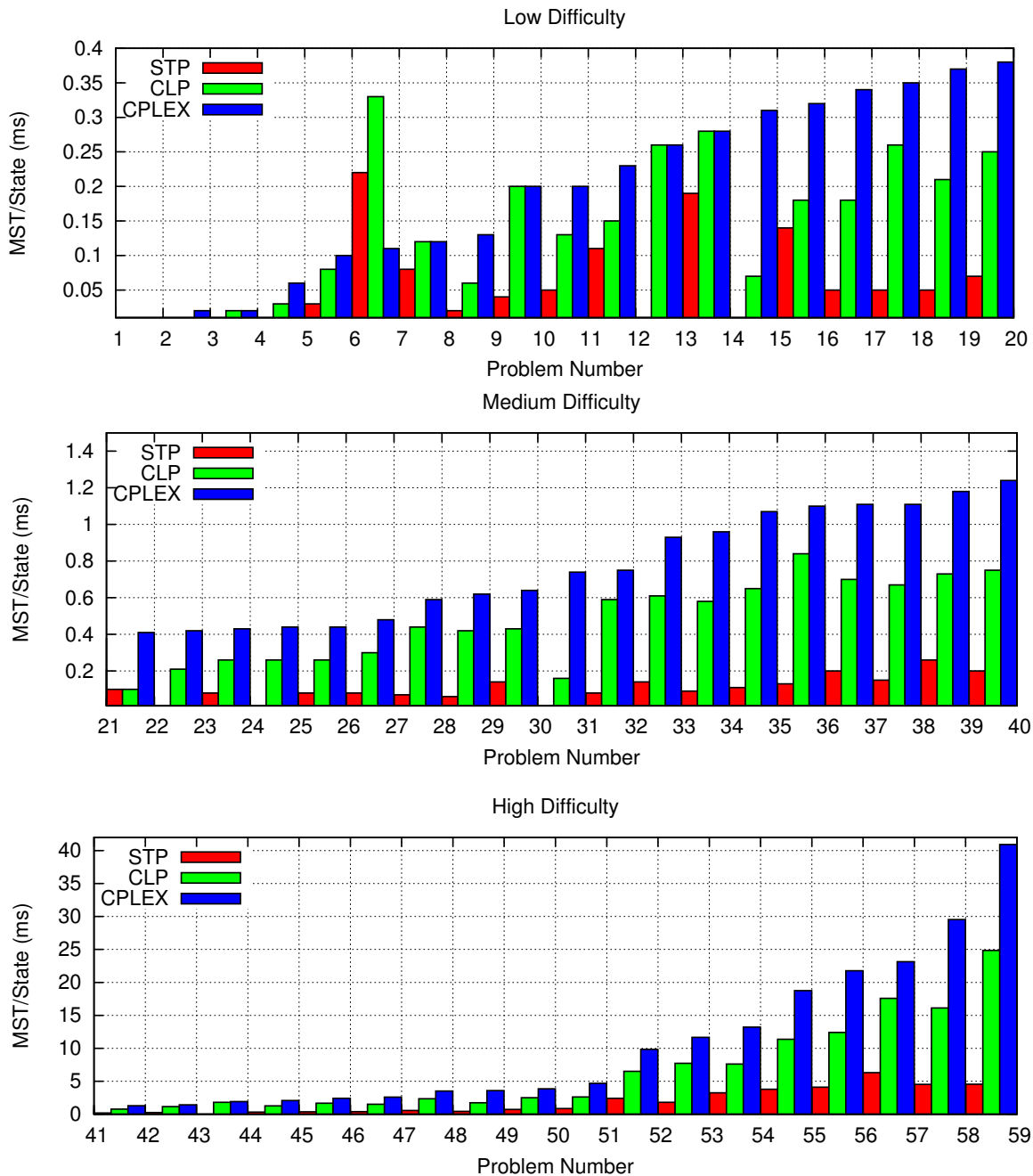

Figure 19: Mean Scheduling Time (MST) per State on Temporal Planning Problems. The problem number appears under the leftmost of the the three corresponding columns in each case.

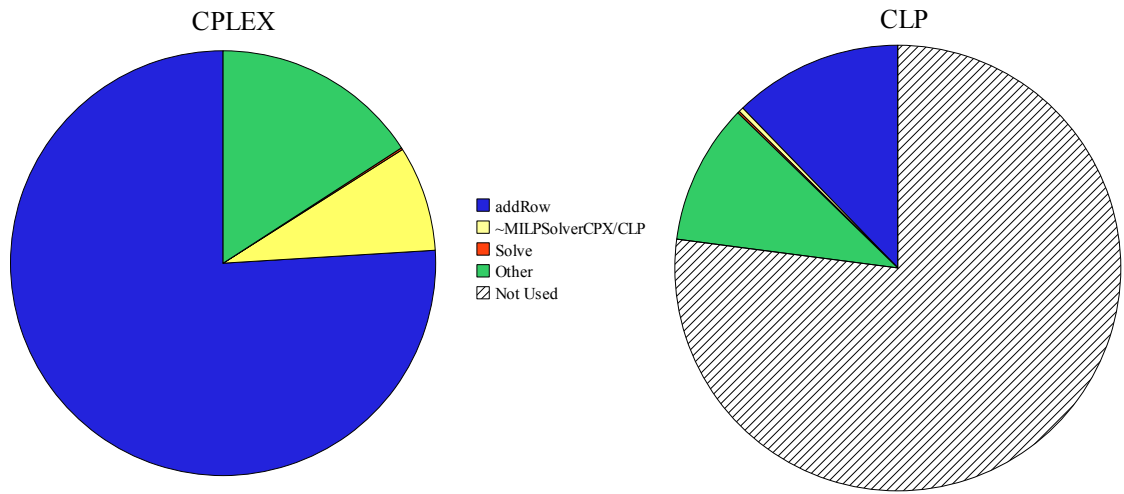

Figure 20: Time spent in various activities by each of the solvers, CPLEX and CLP, viewed as a proportion of the total time spent by CPLEX. The slice labelled '∼MILPSolverCPX/CLP' is time spent in the destructor for the MILP solver in CPLEX or CLP: this is a housekeeping operation in the implementations (which are both written in C++).

Considering the relative performance of the STP and LP solvers, it is clear that there are overheads incurred by the necessary (for domains with continuous effects) move to using an LP rather than an STP. The mean ratio of time spent scheduling on the same problems by CPLEX to that spent using the STP solver is 5.81, the figure for CLP is 3.71. Analysis of the data suggests that these ratios do not change as problem difficulty increases, rather the overhead is a constant factor on harder problems.

Despite increased scheduling overheads it is still worth noting that solving the scheduling problem is a relatively small fraction of the cost of the reasoning done at each state. Once the plan to a given state has been scheduled to check for feasibility the state is evaluated using the temporal RPG heuristic described in Section 9. It is well known, from analysis of the performance of FF and other forward search planners, that the majority of search time is spent in evaluating this heuristic. To give an indication of the relative cost of scheduling versus heuristic computation we give an admissible estimate of the mean fraction of the time spent, per-state, running the scheduler versus computing the heuristic. This estimate is guaranteed to overestimate the true mean because in some states the scheduler will demonstrate that the temporal problem has no solution: in such states the RPG heuristic will never be evaluated, so the heuristic evaluation is actually applied to fewer states than the scheduler. Nonetheless, our data shows that, across these problems, using the STP solver scheduling accounts for on average less than 5% of state evaluation time. For CLP and CPLEX the figures are 13% and 18% respectively. This suggests that, although scheduling does add some overhead to solving problems, these are relatively small compared to the cost of heuristic computation.

A perhaps surprising observation that can be made from Figure 19 is that CLP generally solves the scheduling problems much more efficiently than CPLEX. Given the reputation of CPLEX as a highly efficient commercial LP solver we wanted to investigate why this is the case on our problems.

We performed a further analysis of the profiling data, breaking down the results by function call, to observe the time spent in the various aspects of constructing and solving an LP thorough the CLP and CPLEX library calls. The data, presented in Figure 20 shows the time spent in each function as a fraction of the total time taken by CPLEX to schedule plans (each summed across all problems). The 'not used' section in the CLP data represents the time saved using CLP versus CPLEX. This presentation means that equally sized slices of each of the 'pies' represent the same length of time being taken by either of the solvers in their respective methods.

The important insight that we can gain from this data is that most of the time in both of the LP solvers is not spent in the solve function, indeed it can be observed that the search portion is negligible: it is barely visible. The majority of time is, in fact, spent in adding rows to the LP matrix, i.e. in adding constraints to the LP before it is actually solved. Comparing CPLEX to CLP, it takes over 6 times longer, on average, to add a row to the matrix. The LPs being created for both are identical, and hence involve adding the same number of rows to the matrix. The 'other' portion of the chart corresponds to other methods, many of which also take longer than search, but that are again pre-processing steps such as adding new columns (variables) and setting upper bounds. Since adding rows to the matrix is a significant portion of the time taken in constructing-then-solving the LPs in COLIN, this results in a large overhead. The LPs created by COLIN are very small and simple to solve, compared with the difficult industrial-sized problems for which CPLEX is designed.

Our results suggest that, in fact, the best type of LP solver to use for this task is a relatively light-weight LP solver, with few overheads, that can create models efficiently, even if perhaps it would not scale to other large-scale problems. The other notable, although less marked, difference between the two LP solvers is the time spent in the destructor, called to free up the memory used by the LP solver after each state has been evaluated. Here, it takes 23 times longer, on average, to call the destructor for CPLEX than the destructor for CLP. This has less impact than the row-adding overheads, since the LP is only deleted once per state, rather than once per LP constraint. In general, this would not normally be a noticeable issue when solving a single difficult LP. However, in COLIN, where the number of LPs solved is equal to the number of states evaluated, this overhead does become noticeable.

One interesting outcome of this study is that if, in the future, COLIN were to be extended to non-linear continuous change, requiring the use of a mathematical programming solver at each state (along with other research developments), the overheads may well not be prohibitive. The search within the solver, which is where the greater overhead would occur due to this change, is in fact not the major contributor to the time overheads of using an LP.

## 13. Conclusions

As the range of problems that can be solved effectively by planners grows, so does the range of opportunities for the technology to be applied to real problems. In recent years, planning has extended to solve problems with real temporal structure, requiring temporal coordination, problems that include metric resources and interactions between their use and the causal structure of plans. We have shown how that range can be extended still further, to include linear continuous process effects. Each extension of the power of planners demands several steps. The first is to model the extension in a form that allows the relationship between the constraints imposed on plans by the new expressiveness, and the actions that can be used to solve the problem, to be properly expressed. The second step is to develop a means by which to represent the world state consistently, in order to characterise the space in which the search for a plan is conducted. The third step is to develop a way to compute the progression of states using the action models in this extended representation. Once this step is complete, it is, in principle, possible to plan: a search space can be constructed and searched using classic simple search techniques. In practice, this process is unlikely to lead to solutions of many interesting problems so the fourth step, in order to make the search possible in large spaces, is to construct an informed heuristic to guide the search.

In this paper we have built on earlier work that completed the first steps, adding the third and fourth steps that allow us to solve planning problems with continuous effects. The tools we have used to achieve this are well-established Operations Research tools: LP solvers and their extensions to MILP solvers. The contributions we have made are to show how these tools can be harnessed to check consistency of states, to model state progression and to compute heuristics that can successfully guide search in the large spaces that develop for these planning problems.

An additional contribution is that we have established a collection of benchmark problems for this direction of research in planning. The planning community has witnessed that the creation of benchmarks and their propagation is a powerful aid in the development of the technology, supporting clear empirical evaluation and challenging researchers to improve on the results of others. We have shown that COLIN can solve interesting and complex problems, but there remains much room for improvement. Apart from extending the capability of the planner by improving the informedness of its heuristic and by improving the early pruning of dead end states, there is also the opportunity to extend still further the range of problems that can be expressed and solved. In particular, we are interested in problems with non-linear continuous effects, such as power and thermal curves. It seems possible that such non-linear effects might be approached by a similar approach to that used in COLIN, adapting a NLP solver to the same role as the LP solver in COLIN. Alternatively, it might be possible to approximate non-linear effects with piecewise linear effects, in much the same way that we did for the AUV domain described in this paper, but performing the process automatically.

Planning is becoming an increasingly key technology as robotic systems become more powerful and more complex and we begin to see the limits of low level control strategies in managing the control of these systems. Autonomy demands more powerful predictive control and it is planning that offers possible solutions to this problem. Planning with continuous effects will be an important tool in the collection that we can offer in tackling these new demands.

## Acknowledgments

The authors wish to thank the handling editor, Malte Helmert, and the anonymous reviewers for their considerable contributions to this paper. The authors also wish to thank members of our Planning Group for their helpful discussions during the long gestation of this work.

The authors also wish to acknowledge the EPSRC for their support of this work, specifically through grants EP/G023360/2 and EP/H029001/2.

## Appendix A. Glossary

| Name | Description | First Use |
|------|-------------|-----------|
| $a \downarrow (i, v)$ | The lower bound on assignment effects on variable $v$ due to actions at layer $i$ in a reachability graph. | 62 |
| $a \uparrow (i, v)$ | The upper bound on assignment effects on variable $v$ due to actions at layer $i$ in a reachability graph. | 62 |
| *Action compression* | A technique for simplifying the structure of durative actions by treating them as a simple non-durative action with the union of the effects of both ends of the durative action and the union of their preconditions. | 9 |
| *al* | Action layer in the reachability graph constructed for heuristic purposes. | 26 |
| $ce(i)$ | Function returning the variable corresponding to the end time for a snap-action at position $i$ in the current plan. | 21 |
| $cs(i)$ | Function returning the variable corresponding to the start time for a snap-action at position $i$ in the current plan. | 21 |
| $dec(i, v)$ | Set of (discrete) decreasing effects on variable $v$ at layer $i$ in a reachability graph. | 62 |
| D | The rate of change of variable $v$ (associated with some state achieved during the execution of a plan. | 21 |
| *dmin* (*dmax*) | The minimum (maximum) duration of an action. We use $dmin(a)$ $(dmax(a))$ where the relevant action is required to be explicit and $dmin(a, t)$ $(dmax(a, t))$ where the value is anchored to an action in layer $al(t)$. | 14 |
| E | The event list recording action start times for durative actions whose end points have not yet been included in a plan. | 13 |
| $eff_x^+$ | Propositional add effects of an action, where $x$, when present, indicates whether at the start or end of the action. | 4 |
| $eff_x^-$ | Propositional delete effects of an action, where $x$, when present, indicates whether at the start or end of the action. | 4 |
| $eff_x^n$ | Numeric effects of an action, where $x$, when present, indicates whether at the start or end of the action. | 4 |
| $estep_i$ | The name of the LP variable corresponding to the time at which a durative action will finish, having started as the $i$th step in a plan, but not having finished within the plan constructed so far. | 20 |
| $elapsed(a)$ | The maximum time for which action $a$ could have been executing in a state that is being heuristically evaluated. | 27 |
| $f(i)$ | The variable in the STN in CRIKEY3 that corresponds to the time at which a currently incomplete action will eventually finish. | 15 |

| Name | Description | First Use |
|---|---|---|
| *fl* | Fact layer in the reachability graph constructed for heuristic purposes. | 26 |
| $inc(i, v)$ | Set of (discrete) increasing effects on variable $v$ at layer $i$ in a reachability graph. | 62 |
| $inv(S)$ | The invariants that are active in state $S$. | 14 |
| *Left packing* | A structure of plans with concurrency in which all concurrent actions start simultaneously. | 39 |
| *now* | The name of the variable created to represent the time at the end of the current plan in each STP or LP used to check temporal consistency of a state. | 15 |
| $\langle op, i, dmin, dmax \rangle$ | Event record in a CRIKEY state, containing the durative action, $op$, that started at step $i$, and the minimum and maximum duration of the action. | 14 |
| $pre_\dashv$ | Conditions required to complete an action. | 4 |
| $pre_\leftrightarrow$ | Invariant conditions of a durative action. | 5 |
| $pre_\vdash$ | Conditions required to initiate an action. | 4 |
| $p(a)$ | The bound on the number of instances of durative action $a$ that may execute concurrently. | 28 |
| *remaining(e)* | The maximum amount of remaining time over which an action in event record $e$ could continue to be executing following a state being heuristically evaluated. | 28 |
| *rem(t, a)* | Information associated with durative action $a$ in $al(t)$ in the reachability analysis constructed by COLIN, indicating how much time $a$ could continue to execute from this layer. | 33 |
| $step_i$ | The name of the LP variable corresponding to the time at which action $a_i$ is applied in a plan. | 20 |
| $t(i)$ | The variable in the STP in CRIKEY3 that represents the time at which step $i$ in a plan is to be executed. | 14 |
| *Temporal coordination* | The property of planning problems that require some for of concurrency in order to manage the interactions between the actions or deadlines. | 39 |
| *Time-dependent change* | Action effects that refer to ?duration, causing numeric fluents to change by different amounts according to the length of the action causing the effect. | 2 |

| Name | Description | First Use |
|---|---|---|
| `#t` | Used to describe continuous change: for a complete account of its use and semantics, see original discussion on its use in PDDL2.1 (Fox & Long, 2003). | 3 |
| $ub(\mathbf{w}, \mathbf{x}, \mathbf{y})$ | Function used to calculate bounds on the effects of continuous numeric change. | 29 |
| $\mathbf{v}$ | Used to represent the vector of metric fluents associated with a planning domain, and their values in a state. The vector is treated as indexable: $\mathbf{v}[i]$ is the $i$th entry in $\mathbf{v}$. | 5 |
| $\mathbf{v}'$ | The vector of values of metric fluents at the start of a state, immediately following step effects of application of an action. | 21 |
| $\mathbf{v}_{min}, \mathbf{v}_{max}$ | The vectors of lower and upper bounds on the values of the numeric variables in a state (during plan construction). | 24 |
| $\mathbf{w}$ | Symbol used to represent a vector of constants of equal dimension to the size of the vector of metric fluents in the relevant planning problem. | 5 |

## Appendix B. The Metric Relaxed Planning Graph Heuristic

The Relaxed Planning Graph (RPG) heuristic of Metric-FF (Hoffmann, 2003) has been the most popular numeric planning heuristic over the last decade, being widely used in many planners. The intuition behind the heuristic is to generalise the 'delete-relaxation' to include numeric variables. In the case of propositions, the relaxation is to simply ignore propositional delete effects so, as (relaxed) actions are applied, the set of true propositions is non-decreasing. In the case of numbers, the relaxation replaces exact assignments to numeric variables with bound constraints for their upper and lower bounds. Applying relaxed actions extends the bounds by reducing lower bounds with decrease effects and increasing upper bounds with increase effects. Checking whether a numeric precondition is satisfied is then simply a matter of testing whether the constraint is satisfied by *some* value within the bounds. The delete-relaxed problem can be solved (non-optimally) in polynomial time, and the number of actions in the resulting relaxed plan can then be taken as a heuristic estimate of the distance from the evaluated state to the goal.

The purpose of the RPG is to support this heuristic computation. Relaxed planning is undertaken in two phases: graph expansion, and solution extraction. In the graph expansion phase the purpose is to build an RPG, identifying which facts and actions become reachable. The RPG consists of alternate *fact layers*, consisting of propositions that can hold and optimistic bounds on $\mathbf{v}$, and *action layers*, containing actions whose preconditions are satisfied in the preceding fact layer. In the case of propositional preconditions, a precondition is satisfied if the relevant fact is contained in the previous layer. In the case of numeric preconditions, these are satisfied if some assignment of the variables appearing in the precondition, consistent with the upper and lower bounds, lead to it being satisfied. We define the function $ub(\mathbf{w}, \mathbf{x}, \mathbf{y})$ as:

$$ub(\mathbf{w}, \mathbf{x}, \mathbf{y}) = \sum_{\mathbf{w}[j] \in \mathbf{w}} \left\{ \begin{array}{ll} \mathbf{w}[j] \times \mathbf{y}[j] & \text{if } \mathbf{w}[j] \geq 0 \\ \mathbf{w}[j] \times \mathbf{x}[j] & \text{otherwise} \end{array} \right.$$

(this is the same function as is defined in Section 9.2).

Then, denoting a fact layer $i$ as a set of propositions, $fl(i)$, and upper and lower variable bounds $(\mathbf{v}_{min}(i), \mathbf{v}_{max}(i))$, a precondition $\mathbf{w} \cdot \mathbf{v} \geq c$ of an action in layer $i$ is considered true iff:

$$ub(\mathbf{w}, \mathbf{v}_{min}(i), \mathbf{v}_{max}(i)) \geq c$$

To seed graph construction, fact layer 0 contains all facts that are true in $S$. Thus, action layer 0 consists of all actions whose preconditions are satisfied in fact layer 0. Fact layer 1 is then set to be the optimistic outcome of taking fact layer 0, and applying each of the actions in action layer 0. More formally, considering propositions, applying the actions in action layer $i$, i.e. the actions $al(i)$ leads to a fact layer $i + 1$ where:

$$fl(i + 1) = fl(i) \cup \{eff^+(a) \mid a \in al(i)\}$$

Considering numbers, in action layer $i$ the set of optimistic increase and decrease effects on a variable $v$ across all actions are, respectively:

$$inc(i, v) = \{(ub(\mathbf{w}, \mathbf{v}_{min}(i), \mathbf{v}_{max}(i)) + c) > 0 \mid \exists a \in al(i) \text{ s.t. } \langle v, \texttt{+=}, \mathbf{w} \cdot \mathbf{v} + c \rangle \in eff^n(a)\}$$

$$dec(i, v) = \{(ub(\mathbf{w}, \mathbf{v}_{max}(i), \mathbf{v}_{min}(i)) + c) < 0 \mid \exists a \in al(i) \text{ s.t. } \langle v, \texttt{+=}, \mathbf{w} \cdot \mathbf{v} + c \rangle \in eff^n(a)\}$$

The exchange of the minimum and maximum bounds for $\mathbf{v}$ in these two expressions is important: it causes each expression to be as extreme as possible in the appropriate direction. Similarly, the optimistic upper and lower bounds on $v$, following all available assignment effects, are:

$$a \uparrow (i, v) = \max\{(ub(\mathbf{w}, \mathbf{v}_{min}(i), \mathbf{v}_{max}(i)) + c) \mid \exists a \in al(i) \text{ s.t.} \langle v, =, \mathbf{w} \cdot \mathbf{v} + c \rangle \in \mathit{eff}^n(a)\}$$

$$a \downarrow (i, v) = \min\{(ub(\mathbf{w}, \mathbf{v}_{max}(i), \mathbf{v}_{min}(i)) + c) \mid \exists a \in al(i) \text{ s.t.} \langle v, =, \mathbf{w} \cdot \mathbf{v} + c \rangle \in \mathit{eff}^n(a)\}$$

The new bounds then become:

$$\mathbf{v}_{max}(i+1)[j] = \max\{a \uparrow (i, \mathbf{v}[j]), \mathbf{v}_{max}(i)[j] + \sum inc(i, \mathbf{v}[j])\}$$

$$\mathbf{v}_{min}(i+1)[j] = \min\{a \downarrow (i, \mathbf{v}[j]), \mathbf{v}_{min}(i)[j] + \sum dec(i, \mathbf{v}[j])\}$$

That is, to find the upper (lower) bounds of $\mathbf{v}[j]$ at the next layer, for each we have a choice of applying the largest (smallest) single assignment effect, or the sum of all increase (decrease) effects. Having computed the bounds of all variables in layer $i + 1$, graph expansion then continues iteratively, finding actions applicable in action layer $i + 1$, and hence the facts in layer $i + 2$, and so on. Graph expansion terminates in one of two cases: either a fact layer satisfies all propositional and numeric goals, or the addition of further layers would never lead to more preconditions being satisfied — a condition signalled when no new propositions are appearing and the accumulation of larger or smaller bounds on variables would not lead to any more numeric preconditions becoming satisfied. In this case, the relaxed problem cannot be solved and hence, in the original problem, no plan starting from $S$ can reach $G$. The heuristic value of the state is then set to $\infty$.

Assuming graph expansion terminates with all goals reached, the second phase is to extract a solution from the planning graph. This is a recursive procedure, regressing from the goals back to the initial fact layer. Each fact layer is augmented with a set of goals (facts or numeric preconditions) that are to be achieved at that layer. Beginning by inserting the top-level goals $G$ into the planning graph at the first layers at which they each appeared, solution extraction repeatedly picks the latest outstanding goal in the planning graph and selects a way to achieve it. For propositional goals, a single action (with an effect adding the goal) is chosen, and its preconditions are inserted as goals to be achieved (again, at the earliest possible layers). To satisfy the numeric goal $\mathbf{w} \cdot \mathbf{v} \geq c$ at layer $i$, actions with effects acting upon the variables (with non-zero coefficients) in $\mathbf{v}$ are chosen, until the net increase of $\mathbf{w} \cdot \mathbf{v}$, $k$, is sufficient to allow the residual precondition $\mathbf{w} \cdot \mathbf{v} \geq c - k$ to be satisfied at fact layer $i - 1$. At this point, this residual precondition is added as a goal to be achieved at layer $i - 1$ (or earlier if possible), and the preconditions of all the actions chosen to support this precondition are added as goals to be achieved at previous layers.

Solution extraction terminates when all outstanding goals are to be achieved in fact layer 0, since they are then true in the state being evaluated and need no supporting actions. The actions selected in solution extraction form the *relaxed plan* from $S$ to the goal. The length (number of actions) of this relaxed plan forms the heuristic estimate, $h(S)$. Additionally, the actions in the relaxed plan that were chosen from action layer 0 form the basis of the 'helpful actions' in $S$, used to restrict the states explored by enforced hill-climbing search: any action with an effect in common with the actions chosen from action layer 0 is considered to be helpful.

## Appendix C. Temporal Reasoning in Relaxed Planning Graphs

Several approaches have been proposed for building temporal relaxed planning graphs (TRPGs). There are three additional features that TRPGs can attempt to manage, compared with RPGs:

1. The temporal structure of durative actions: $a_\dashv$ can only be applied if $a_\vdash$ has been applied before it.

2. Action durations: end effects of actions are only available at an appropriate delay after they have started.

3. The PDDL2.1 start–end semantics, allowing effects and preconditions to be attached to both the starts and ends of actions.

The TRPG employed in Sapa (Do & Kambhampati, 2003) satisfies the first two of these, but not the third. In Sapa, each action is compressed into a temporally-extended action obeying the TGP semantics, before discarding delete effects, as a relaxation, and building a TGP-style planning graph (Smith & Weld, 1999). The use of compression and a time-stamped TGP representation captures durations and the start-before-end relationships, but the use of compression causes the heuristic to find dead-ends in cases where there is required concurrency.

The TRGP used in CRIKEY (Coles, Fox, Halsey et al., 2008) avoids action compression, but it ignores the durations of actions. A non-temporal RPG is built in terms of the snap-actions used during search, with an additional precondition on each end snap-action that a particular dummy fact, added by its corresponding start, has appeared in the preceding fact layer. The use of snap-actions means no preconditions or effects are lost (ensuring that the heuristic no longer identifies the false dead-ends created by the approach used in Sapa), but the limitation of the heuristic is that there is no forced separation between the start and and end of an action, but only an ordering constraint.

In CRIKEY3 (Coles, Fox, Long et al., 2008a), the heuristic is constructed to combine the strengths of both of these earlier heuristics, accounting for the durations of actions, whilst also respecting the start–end semantics. We now briefly describe the construction of this TRPG, since it is the basis for the heuristic used in COLIN. The structure of the TRPG is similar to that constructed in Metric-FF, but instead of each fact layer being assigned an index, it is assigned a time-stamp (indicating the minimum amount of time that must pass after the initial layer before the facts in the layer in question can appear). To capture the durations of actions, we record, for each end action $a_\dashv$, the earliest layer $t_{min}(a_\dashv)$ at which it can appear. This value is set to 0 for all actions that are already executing in the state being evaluated (as there is no need to first insert the start of the action into the RPG). For other actions, the value is initialised to $\infty$, before commencing TRPG construction.

To build a TRPG we follow Algorithm 2. First, a number of initialisation steps are performed. The time-zero fact layer $fl(0)$ is initialised (at line 1) to contain all the facts true in $S$[10]. The set $ea$ is initialised to contain all the end snap-actions that must appear in the TRPG — if an action is executing, its end has to be reachable (i.e. appear in the TRPG), or else the state $S$ is a dead end. If $ea$ is empty, and $S$ satisfies the goals $G$ (line 14), then no TRPG need be built, since the plan is complete.

Following initialisation, the TRPG is expanded, beginning with $t = 0$ and using the fact layer $fl(t)$ to determine the action layer $al(t)$. If the preconditions of an action are satisfied in a fact layer

---

10. For simplicity we omit the handling of numeric fluents from this explanation — this is performed exactly as in the earlier description of the RPG heuristic implemented in Metric-FF.

---

**Algorithm 2**: Building a Temporal RPG in CRIKEY3.

---

**Data**: $S = \langle F, E, T \rangle$ - state to be evaluated

**Result**: $R = \langle fls, als \rangle$, a relaxed planning graph

1   $fl(0) \leftarrow F$;

2   $fls \leftarrow \langle fl(0) \rangle$;

3   $als \leftarrow \langle \, \rangle$;

4   $t \leftarrow 0$;

5   $ea \leftarrow \emptyset$;

6   $prev\_al \leftarrow \emptyset$;

7   $prev\_fl \leftarrow fl(0)$;

8   **foreach** $a_{\dashv}$ **do**

9     **if** $\{e \in E \mid e.op = a\} = \emptyset$ **then**

10       $\lfloor \; t_{min}(a_{\dashv}) \leftarrow \infty$;

11     **else**

12       $t_{min}(a_{\dashv}) \leftarrow 0$;

13       $ea \leftarrow ea \cup \{a_{\dashv}\}$;

14   **if** $G \subseteq fl(0) \wedge ea = \emptyset$ **then return** '*S is a goal state*';

15   **while** $t < \infty$ **do**

16     $fl(t + \epsilon) \leftarrow prev\_fl$;

17     $al(t) \leftarrow \{a_{\dashv} \mid pre(a_{\dashv}) \subseteq fl(t) \wedge t_{min}(a_{\dashv}) \leq t\}$;

18     **foreach** $a_{\dashv} \in al(t) - prev\_al$ **do**

19       $\lfloor \; fl(t + \epsilon) \leftarrow fl(t + \epsilon) \cup eff^+(a_{\dashv})$;

20     $al(t) \leftarrow al(t) \cup \{a_{\vdash} \mid pre(a_{\vdash}) \subseteq fl(t)\}$;

21     **foreach** $a_{\vdash} \in al(t) - prev\_al$ **do**

22       $fl(t + \epsilon) \leftarrow fl(t + \epsilon) \cup eff^+(a_{\vdash})$;

23       $t_{min}(a_{\dashv}) = \min[t_{min}(a_{\dashv}), t + dmin(a)]$;

24     $als \leftarrow als + al(t)$;

25     $fls \leftarrow fls + fls(t + \epsilon)$;

26     $prev\_al \leftarrow al(t)$;

27     **if** $G \subseteq fl(t + \epsilon) \wedge ea \subseteq al(t)$ **then**

28       $\lfloor$ **return** $R = \langle fls, als \rangle$;

29     **if** $prev\_fl \neq fl(t + \epsilon)$ **then**

30       $prev\_fl \leftarrow fl(t + \epsilon)$;

31       $t \leftarrow t + \epsilon$;

32     **else**

33       $prev\_fl \leftarrow fl(t + \epsilon)$;

34       $ep = \{t_{min}(a_{\dashv}) \mid pre(a_{\dashv}) \subseteq fl(t) \wedge t_{min}(a_{\dashv}) > t\}$;

35       **if** $ep \neq \emptyset$ **then** $t \leftarrow \min[ep]$;

36       **else** $t \leftarrow \infty$;

37   **return** '*S is a dead end*';

---

$fl(t)$ then whether it can appear in $al(t)$ depends on whether it is a start or end snap-action. The first and simpler case is that, if a start snap-action $a_\vdash$ is applicable, it is added to $al(t)$ and $t_{min}(a_\dashv)$ is set to $t + dmin(a)$, where $dmin(a)$ is an *a priori* lower bound on the duration of $a$. If there is a state-independent measure of the minimum duration of $a$, i.e. a minimum duration constraint referring only to constants, then this is taken as the value of $dmin(a)$. Otherwise, if all the minimum duration constraints depends on the state in which the action is applied, then $dmin(a) = \epsilon$: all that is certain is that *some* time must elapse between the start and the end of the action. The state-dependent terms cannot be evaluated since the TRPG determines a relaxed state, not a real state.

The second case, covering end snap-actions, is that if the preconditions of an end action $a_\dashv$ become satisfied in a fact layer $fl(t)$, then addition of $a_\dashv$ to $al(t)$ depends on whether the start of the action can have occurred sufficiently far in the past (line 17). If $t \geq t_{min}(a_\dashv)$ then $a_\dashv$ is added to $al(t)$; if $t < t_{min}(a_\dashv) < \infty$, then $a_\dashv$ is postponed until $al(t_{min}(a_\dashv))$; otherwise, the start of the action has yet to appear, and $a_\dashv$ is postponed until the relevant start appears.

Having determined which actions newly appear in $al(t)$, the fact layer $fl(t + \epsilon)$ is updated as in the non-temporal RPG case, by taking $fl(t)$ and (optimistically) applying the effects of all the actions in $al(t)$. If $fl(t + \epsilon)$ and $al(t)$ do not contain the necessary goals and end snap-actions (line 27) then it must be decided which fact layer to consider next. Clearly, it is infeasible to create new fact layers at $\epsilon$ spacing between $fl(0)$ and the fact layer at which the goals appear. Fortunately, it is also unnecessary, as many of the fact and action layers in such a graph would be identical. Instead, we determine the next fact layer to consider as follows:

- If there are new facts in $fl(t+\epsilon)$ that were not true in $fl(t)$ (line 29), the next layer to expand is $fl(t+\epsilon)$ — the appearance of new and potentially useful facts makes it necessary to consider whether any actions become applicable in that layer.

- If $fl(t + \epsilon) = fl(t)$, then we know that visiting $fl(t + \epsilon)$ is futile. In this case (line 34), the time-stamp of the next fact layer to visit is the earliest future point at which the postponed end of an action becomes applicable:

$$\min\{t_{min}(a_\dashv) \mid pre(a_\dashv) \subseteq fl(t) \wedge t_{min}(a_\dashv) > t\}$$

If the minimum of these values is $\infty$ (or undefined) then the state can be pruned and the procedure exits early, signalling the result to the search procedure.

When a TRPG is successfully constructed (that is, if the starting state is not a dead end) the graph that is returned contains a finite set of fact and action layers, each associated with a real time value.

Assuming graph expansion terminates with all goals reached, a relaxed solution is extracted. The solution extraction procedure used in Metric-FF needs one minor modification to be suitable for use in a TRPG: if the end of an action $a_\dashv$ is chosen to support a goal at a given fact layer, then if the action $a$ is not already executing in the state being evaluated, the corresponding start $a_\vdash$ must be scheduled for selection (at the layer in which it first appeared). The purpose of this corresponds to that of the dummy facts in CRIKEY: if the end of an action is chosen, its start must also be executed.

As a final remark on this TRPG, timed initial literals (TILs) can be included by employing the machinery introduced to delay the ends of actions until an appropriate layer. If the dummy TIL actions $\{TIL_j ... TIL_m\}$ have yet to be applied then $t_{min}(TIL_j) = 0.0$, since $TIL_j$ could be applied in the first action layer. The intuition here is that the state being evaluated is a snapshot of the world, taken no earlier than after the end of the previous action, but no later than the point at which the next TIL event occurs (due to the constraints discussed in Section 6.1). The minimum timestamps for the later TILs, each $TIL_k \in \{TIL_{j+1} \dots TIL_m\}$, are then set relative to this time point:

$$t_{min}(TIL_k) = ts(TIL_k) - ts(TIL_j).$$

## Appendix D. Post-Hoc Plan Optimisation

This appendix contains details of the MILP construction briefly described in Section 10.2.

### D.1 Optimising for Time Windows

First let us consider the simple case where an action $do$ has a conditional effect on a metric-tracking variable `reward` (where the objective of the problem is to maximise reward), and where the effect occurs depends on the truth value of a single proposition $p$ at some time specifier $ts$ relative to the action $do$ (either `at start`, `over all`, or `at end`):

```
(when (ts (p)) (at end (increase (reward) k))).
```

In the case where $p$ can be manipulated by actions, without allowing the MILP to introduce new actions or completely change the order of the plan steps (with all the complexity these modifications would entail), there is little scope for optimisation. In the case where the truth value of $p$ is dictated by timed initial literals (TILs), we have a more interesting case: changing the time-stamps of the start or the end of $a$ (LP variables $step_i$ and $step_j$), so that the condition is or is not satisfied, has a direct effect on the metric function. This relationship can be encoded within the LP. By way of example, consider the case where $p$ becomes true at time $a$ and false at $b$; then, again, becomes true at $c$ and false at $d$. In this case, we have two time-windows that could potentially satisfy the condition on the effect. Whether the action has to wholly or only partially within one of these windows depends on the time-specifier attached to $p$:

- if $ts =$ `at start`, then $a_\vdash (step_i)$ has to lie within one of the time windows;

- if $ts =$ `at end`, then $a_\dashv (step_j)$ has to lie within one of the time windows;

- otherwise, $ts =$ `over all`, and both $a_\vdash$ and $a_\dashv$ have to lie within one of the time windows.

In all three cases, the question that must be answered is 'does the value of this variable lie within a known range?' — the `over all` case requires a conjunction of two such conditions to hold and, in the other two cases, only one has to hold. For a given step variable $step_i$, and time window $(a, b)$, we can introduce into the (MI)LP a binary variable $switch_{ab}$ corresponding to such an observation, with constraints that take the logical form:

$$switch_{ab} \Leftrightarrow (step_i > a) \land (step_i < b)$$

Thus, if the switch variable takes the value 1, the time-stamp of a point at which $p$ is needed must fall within the time-window $[a, b]$ and vice versa. By introducing two additional binary variables, denoted $ga$ and $lb$, this logical constraint can be represented as a series of inequalities (using $N$ to

denote a large number):

$$
\begin{array}{rcl}
step_i - (a + \epsilon) \times switch_{ab} & \geq & 0 \\
-step_i + (b - \epsilon) \times switch_{ab} & \geq & 0 \\
-step_i + N \times ga & \geq & -a \\
step_i - N \times lb & \geq & b \\
switch_{ab} - ga - lb & \geq & -1
\end{array}
$$

The first two constraints encode the forwards implication: if $switch_{ab}$ is set to 1, then $step_i$ has to lie in the range $[a + \epsilon, b - \epsilon]$ (a non-zero amount of separation, here epsilon, is needed under the PDDL semantics to avoid inspecting the value of $p$ at the same time it is being changed by the TIL). The latter three constraints encode the reverse implication: if $step_i$ is strictly greater than $a$ and strictly less than $b$, then both of $ga$ and $lb$ have to hold the value 1 and thus, so does $switch_{ab}$.

Returning to our example, where the time specifier is `over all`, and the windows are $(a, b)$ and $(c, d)$, the constraints that will be added are:

$$
\begin{array}{rcl}
switch_{ab1} & \Leftrightarrow & (step_i > a) \wedge (step_i < b) \\
switch_{ab2} & \Leftrightarrow & (step_j > a) \wedge (step_j < b) \\
switch_{ab} & \Leftrightarrow & (switch_{ab1} \wedge switch_{ab2}) \\
switch_{cd1} & \Leftrightarrow & (step_i > c) \wedge (step_i < d) \\
switch_{cd2} & \Leftrightarrow & (step_j > c) \wedge (step_j < d) \\
switch_{cd} & \Leftrightarrow & (switch_{cd1} \wedge switch_{cd2}) \\
switch_p & \Leftrightarrow & (switch_{ab} \vee switch_{cd})
\end{array}
$$

That is, $switch_{ab}$ is 1 if the entirety of the action *do* falls within $(a, b)$, $switch_{cd}$ is 1 if it falls within $(c, d)$ and $switch_p$ is 1 if either of these hold. This final switch variable is used to capture the benefit of the effect itself: if it holds the value 1, we increase the value of *reward* at the end of the plan by $k$, that is, we apply the conditional effect. The variable *reward* will already appear in the objective function in the form of the LP variable $reward'_n$, where $n$ is the last step of the plan. Thus, we modify the constraints that define $reward'_n$ so that $k \times switch_p$ is added to this value. This change will then ensure that the variable providing the value of *reward* in the objective function will include the reward of $k$ if the condition on the time at which *do* was executed holds.

Generalising, we can extend this to the case where the conditional effect depends on the truth of a formula $f$ consisting of a conjunction of time-specified propositional facts $[(ts_1\, p_1)...(ts_j\, p_j)]$. For each $(ts_i\, p_i) \in f$, we create constraints, as indicated above, so that the switch variable $switch_{pi}$ can only take the value 1 if $p_i$ holds at the time-specifier $ts_i$. This gives us a list of switch variables $s = [switch_{p1}...switch_{pj}]$. Then, to encode that in fact the conjunction $f$ must hold, we create a variable $switch_f$ and add the constraints:

$$
\begin{array}{c}
-j \times switch_f + 1.switch_{p1} + ... + 1.switch_{pj} \geq 0 \\
switch_f + -1.switch_{p1} + ... + -1.switch_{pj} \geq 1 - j
\end{array}
$$

Defined thus, $switch_f$ takes the value of 1 iff each of the switch variables in $s$ takes the value 1, which is precisely in the case that the conjunct is satisfied. Then, much as before, when updating the constraint dictating the value of the LP variable $reward'_n$, we add $k \times switch_f$ to this value.

### D.2 Optimising Numeric-Dependent Conditions

Perhaps more complex than the case of time windows is where the conditions of a conditional effect depend on the values of the numeric variables in the domain. (The PDDL2.2 definition (Hoffmann

& Edelkamp, 2005) does not include the case where TILs change the values of numeric variables[11], so we do not consider that case here.) In the simple case, the time-specifier of all the numeric conditions is either `at start` or `at end`. More complicated is the case where one or more of the time-specifiers is `over all`. In this case, potentially, all the snap-actions in the plan, between the start and end of the action to which the condition belongs, could affect whether or not the condition associated with the effect is met. We must therefore check the status of the condition at each such point during execution of the action. Suppose we have an action $O$, where $step_k$ and $step_l$ are the variables denoting the start and end time-stamps of the action, and where $O$ has a conditional effect with a numeric precondition in LNF:

```
when (over all (>= (w · v) c)) (at end (increase (reward) k)).
```

To encode this, we need to add constraints to ensure that this conditional outcome occurs iff $\mathbf{w} \cdot \mathbf{v} \geq c$ at all times within $O$. Since all change is linear, then (as with other `over all` conditions on $O$) we only need to check the values of the numeric variables immediately before and immediately after each action time step within $O$, and also immediately following the start and immediately before the end of $O$ itself. Thus, the variables corresponding to values of $\mathbf{v}$ that we must examine are those in the list:

$$e = [\mathbf{v}'_k, \mathbf{v}_{k+1}, \mathbf{v}'_{k+1}, ..., \mathbf{v}_{l-1}, \mathbf{v}'_{l-1}, \mathbf{v}_l].$$

As stated earlier, the case of `at start`/`at end` conditions is somewhat easier: for `at start`, $e = [\mathbf{v}_k]$, and for `at end`, $e = [\mathbf{v}_l]$. Irrespective of the time specifier, on the basis of this list $e$, to capture whether the condition is met, adding a switch variable $switch$ to indicate whether the condition is met for all vectors, and switch variables $[switch_{t1}...switch_{tn}]$ for each element $[1..s]$ of the list $e$, indicating whether it is met for that single vector. The constraints over these (where $\xi$ is a small number) are then:

$$\underset{x \in [1...s]}{\forall} \mathbf{w}.e[x] \geq -N + (N + c) \times switch$$

$$\underset{x \in [1...s]}{\forall} \mathbf{w}.e[x] \leq (c - \xi) + N \times switch_{tx}$$

$$s \times switch + -1.switch_{t0} + ... + -1.switch_{ts} \geq 1 - s.$$

The first quantification ensures that if $switch = 1$, a lower bound of $c$ is imposed on $\mathbf{w} \cdot \mathbf{v}$ for each element of $e$. The second quantification ensures that if the vector $\mathbf{v}$ at index $x$ in $e$ satisfies $\mathbf{w} \cdot \mathbf{v} \geq c$, then the corresponding switch variable $switch_{tx}$ has to take the value 1. The third constraint ensures that if all such switch variables $switch_{tx}$ take the value 1, $switch$ must, too, be set to 1. Having appropriately constrained the $switch$ variable we can update the constraint governing the value of the LP variable $reward'_n$ (the value of $reward$ at the end of the plan) to increase it by the value of $k \times switch$.

### D.3 Optimising Time-Dependent Conditions

The final extension is to allow conditional effects to refer not only to the truth values of timed propositions, or the values of numeric variables, but also to the value of the duration of an action.

---

11. Timed Initial Fluents have been used in some domain models, as an unofficial extension to the language. The semantics of such an extension are as straightforward as Timed Initial Literals.

This situation appears in our example airplane landing problem (Section 10.2) where the value *(total-cost)* is updated by a conditional effect, of which both the condition and the effect depend on `?duration`. We will consider this example in order to show how the MILP can be extended to handle such updates. First, as in the previous cases, we need to add constraints to ensure that if the MILP solver chooses to obtain the conditioned outcome of a conditioned effect, then the condition must be met. So, for the example, we introduce a new variable binary switch variable for each condition, and some new constraints. For the land action for a plane `?p`, starting and finishing at time-stamps action as $step_n$ and $step_m$ respectively, we add a pair of constrained switch variables. For the sake of this example we give these the meaningful names *early* and *late*. The constraints that are added to the LP are then:

$$
\begin{aligned}
\texttt{target p} - step_m + step_n &\leq N \times early \\
step_m - step_n &\leq N - (N + \epsilon - \texttt{target p}) \times early \\
-\texttt{target p} + step_m - step_n &\leq N \times late \\
step_m - step_n &\geq (\texttt{target p} + \epsilon) \times late
\end{aligned}
$$

These new constraints ensure that if the plane lands early, the variable *early* has to take the value 1, and vice versa. Similarly, if it lands late, *late* must take the value 1 and vice versa. In the case of our example, the conditional effects of the action are mutually exclusive, though this is not true in the general case.

Having defined the *early* and *late* switch variables, the objective function for the MILP must be augmented to reflect the conditional outcomes of the action. Two terms must be added — one for each switch variable — for the effect obtained if the switch variable is 1. Abbreviating the terms `earlyPenaltyRate`, `latePenaltyRate` and `latePenalty` to `epr`, `lpr` and `lp`, respectively, the objective terms for plane $p$ are:

$$
early \times (\texttt{epr p}) \times (\texttt{target p} - (step_m - step_n))
$$

$$
late \times (\texttt{lpr p}) \times ((step_m - step_n) - \texttt{target p}) + late \times (\texttt{lp p})
$$

Note that unlike in the previous cases, this objective function is now quadratic: the objective contains terms where a switch variable is multiplied by both a constant and a *step* variable. This arises as, unlike in previous cases, the conditional effect is duration dependent — not a fixed, constant value $k$. Whilst this raises the computational cost of optimising the MILP, the cost is acceptable: it is only incurred once, after a solution plan has been found.

## Appendix E. Details of Empirical Evaluation of Colin

The graphs presented here show the detailed runtime and quality comparisons analysed in Section 12. The comparative data is graphed. Since the graphs sometimes superimpose curves over one another, making it difficult to see how COLIN is performing, Tables 6–13 show the raw time and quality results for COLIN compared with average and best times and qualities for all problems. Best times and qualities are also reported with the corresponding quality or time (respectively) for that solution, and the planner(s) that generated the best result.

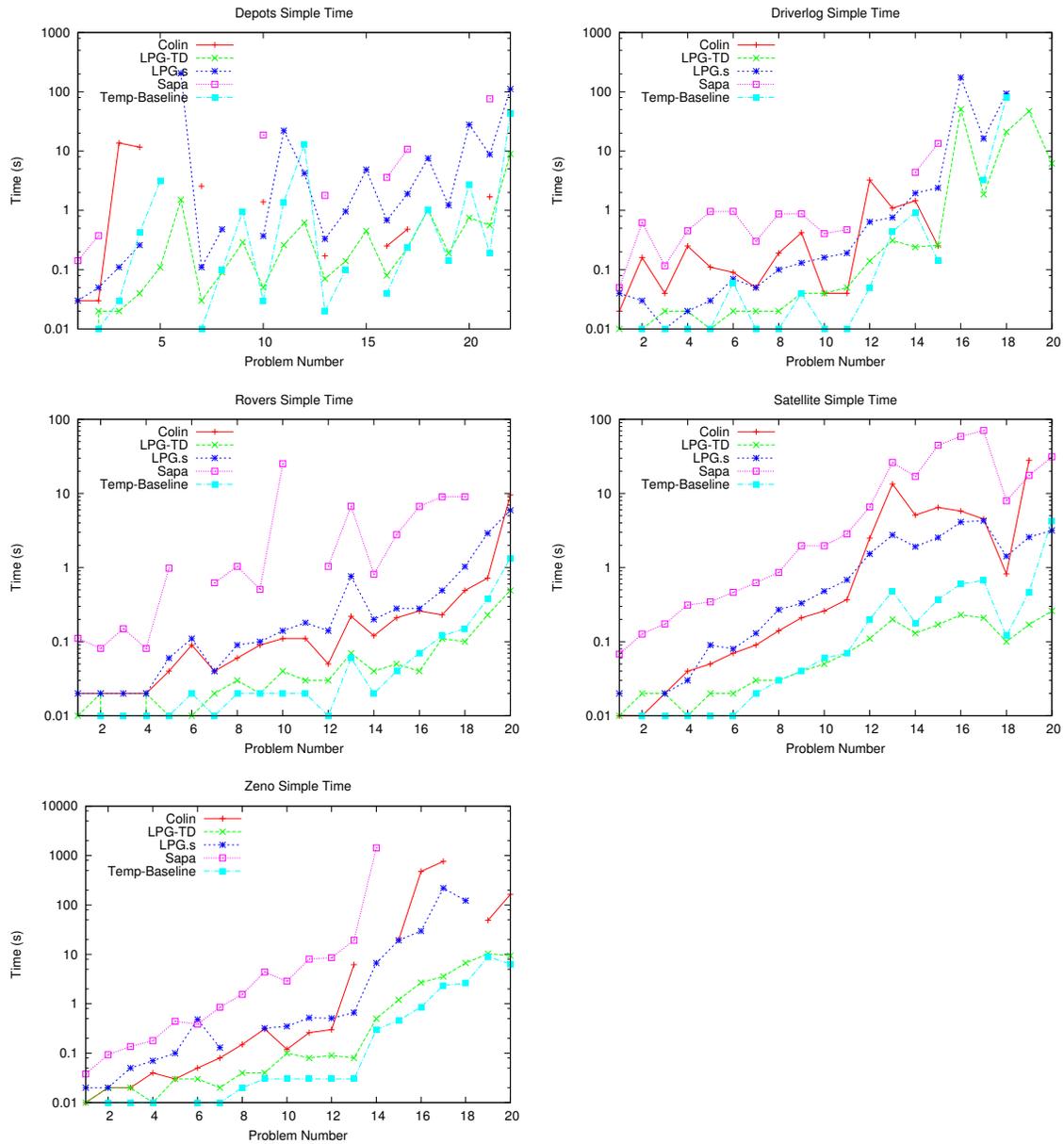

Figure 21: Comparison of time taken to solve problems in various simple temporal planning benchmarks by the planners COLIN, LPG-td, LPG.s, Sapa and a temporal baseline planner. Planners not appearing in a particular dataset did not solve any of the problems in that collection.

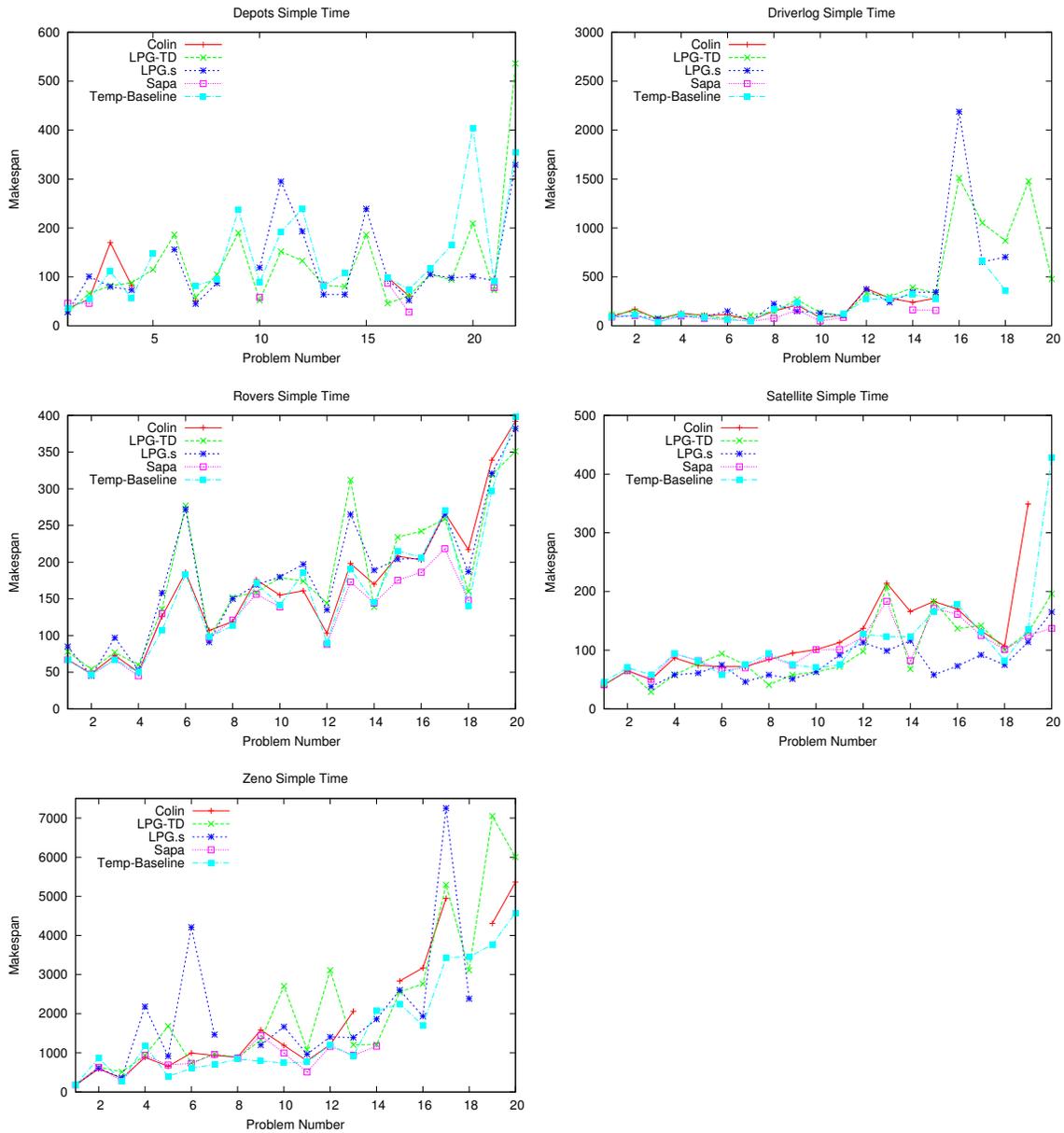

Figure 22: Comparison of plan quality in problems in various simple temporal planning benchmarks by the planners COLIN, LPG-td, LPG.s, Sapa and a temporal baseline planner. Planners not appearing in a particular dataset did not solve any of the problems in that collection.

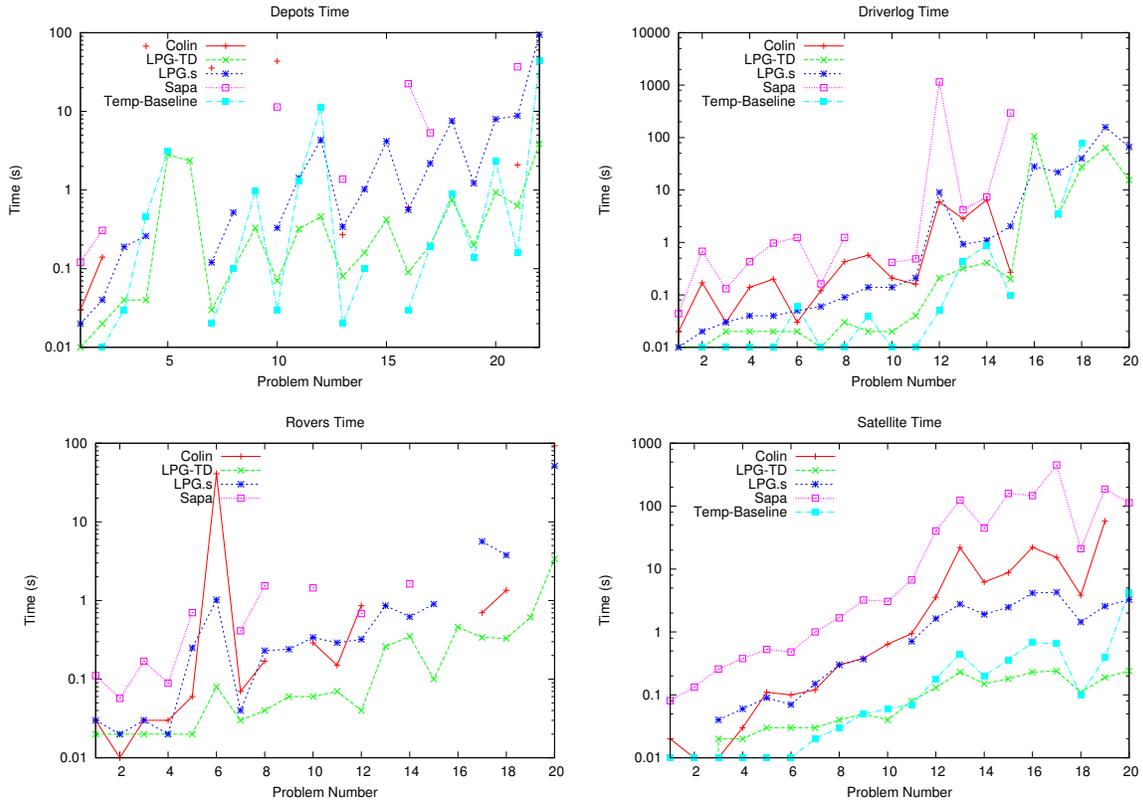

Figure 23: Comparison of time taken to solve problems in more complex temporal planning benchmarks (first set) by the planners COLIN, LPG-td, LPG.s, Sapa and a temporal baseline planner. Planners not appearing in a particular dataset did not solve any of the problems in that collection.

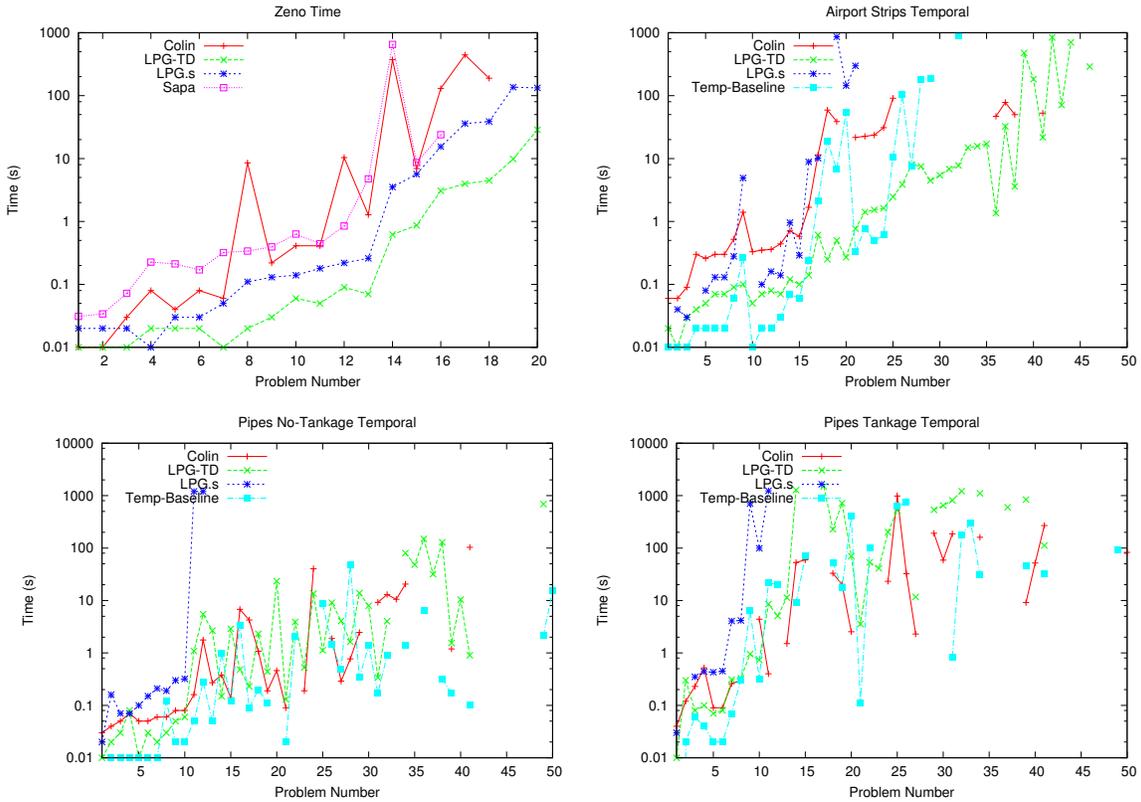

Figure 24: Comparison of time taken to solve problems in more complex temporal planning benchmarks (second set) by the planners COLIN, LPG-td, LPG.s, Sapa and a temporal baseline planner. Planners not appearing in a particular dataset did not solve any of the problems in that collection.

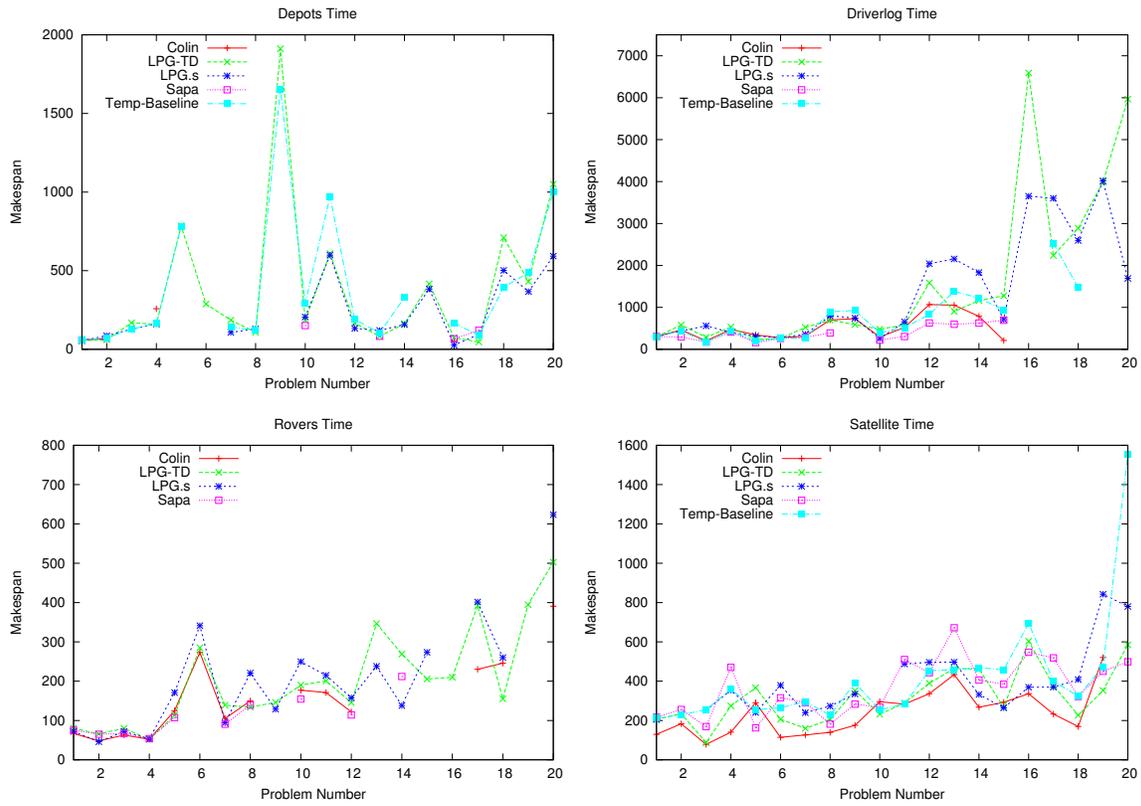

Figure 25: Comparison of plan quality in more complex temporal planning benchmarks (first set) by the planners COLIN, LPG-td, LPG.s, Sapa and a temporal baseline planner. Planners not appearing in a particular dataset did not solve any of the problems in that collection.

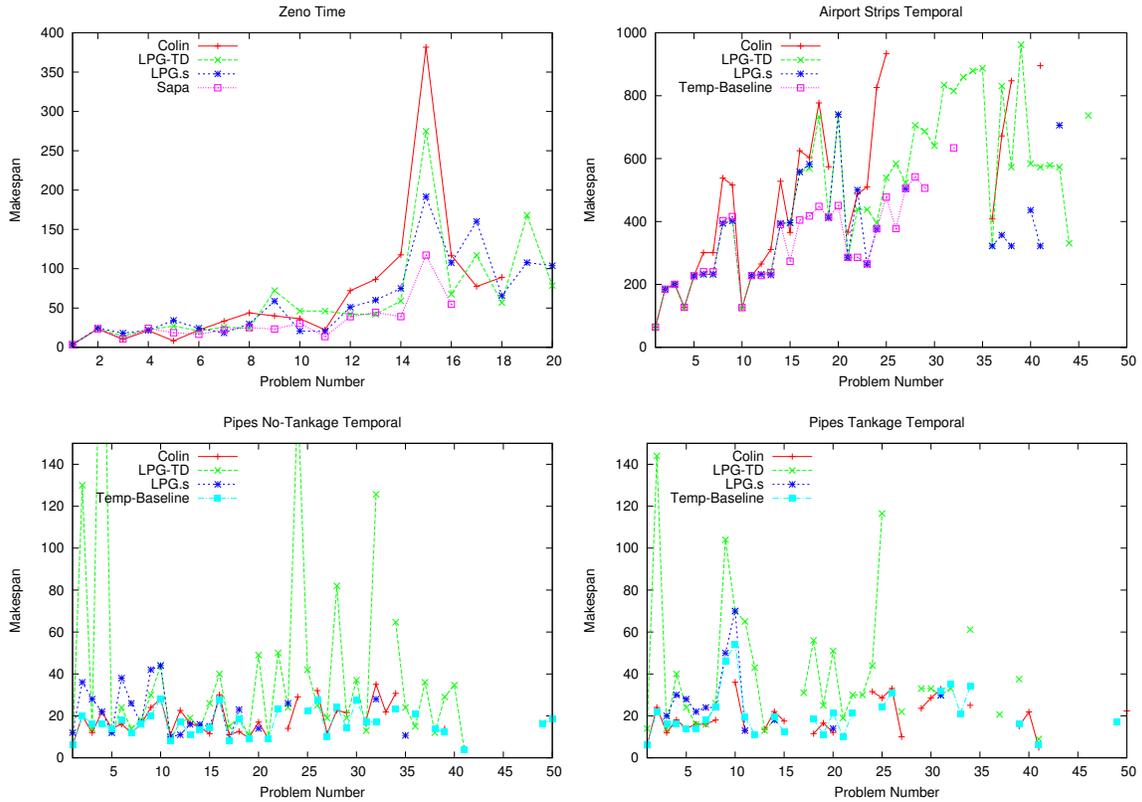

Figure 26: Comparison of plan quality in more complex temporal planning benchmarks (second set) by the planners COLIN, LPG-td, LPG.s, Sapa and a temporal baseline planner. Planners not appearing in a particular dataset did not solve any of the problems in that collection.

| | COLIN | | Average | | Best | | | |
|---|---|---|---|---|---|---|---|---|
| | Time | Quality | Time | Quality | Time | Planner | Quality | Planner |
| depotssimpletime | | | | | | | | |
| 1 | 0.03 | 36.008 | 0.040 | 34.834 | 0.0 (28.000) | LPG-td,TBL | 28.000 (0.0) | LPG-s+td |
| 2 | 0.03 | 54.013 | 0.097 | 64.243 | 0.01 (54.11) | TBL | 46.090 (0.374) | Sapa |
| 3 | 13.69 | 170.037 | 3.462 | 110.817 | 0.02 (82.000) | LPG-td | 80.002 (0.11) | LPG-s |
| 4 | 11.67 | 82.031 | 3.097 | 74.818 | 0.04 (88.000) | LPG-td | 56.24 (0.42) | TBL |
| 5 | | | 1.620 | 131.250 | 0.11 (115.000) | LPG-td | 115.000 (0.11) | LPG-td |
| 6 | | | 103.400 | 171.002 | 1.51 (186.000) | LPG-td | 156.004 (205.29) | LPG-s |
| 7 | 2.56 | 51.024 | 0.677 | 59.049 | 0.01 (82.17) | TBL | 45.001 (0.11) | LPG-s |
| 8 | | | 0.223 | 95.784 | 0.09 (105.000) | LPG-td | 87.003 (0.48) | LPG-s |
| 9 | | | 0.625 | 213.300 | 0.29 (190.000) | LPG-td | 190.000 (0.29) | LPG-td |
| 10 | 1.39 | 90.025 | 4.100 | 81.465 | 0.03 (88.2) | TBL | 52.000 (0.05) | LPG-td |
| 11 | | | 7.933 | 212.845 | 0.26 (152.000) | LPG-td | 152.000 (0.26) | LPG-td |
| 12 | | | 5.880 | 188.251 | 0.62 (133.000) | LPG-td | 133.000 (0.62) | LPG-td |
| 13 | 0.17 | 85.026 | 0.475 | 79.265 | 0.02 (82.17) | TBL | 64.001 (0.33) | LPG-s |
| 14 | | | 0.400 | 83.767 | 0.1 (107.3) | TBL | 64.001 (0.96) | LPG-s |
| 15 | | | 2.655 | 212.502 | 0.45 (186.000) | LPG-td | 186.000 (0.45) | LPG-td |
| 16 | 0.25 | 99.025 | 0.928 | 85.273 | 0.04 (98.18) | TBL | 46.000 (0.08) | LPG-td |
| 17 | 0.48 | 61.016 | 2.716 | 55.261 | 0.23 (61.000) | LPG-td | 28.050 (10.728) | Sapa |
| 18 | | | 3.217 | 109.144 | 1.03 (117.43) | TBL | 105.000 (1.04) | LPG-td |
| 19 | | | 0.517 | 118.787 | 0.14 (164.36) | TBL | 94.000 (0.19) | LPG-td |
| 20 | | | 10.443 | 238.291 | 0.75 (209.000) | LPG-td | 101.002 (27.84) | LPG-s |
| 21 | 1.69 | 96.029 | 17.448 | 85.872 | 0.19 (90.18) | TBL | 73.000 (0.56) | LPG-td |
| 22 | | | 54.667 | 406.609 | 8.95 (536.000) | LPG-td | 329.007 (111.46) | LPG-s |
| driverlogssimpletime | | | | | | | | |
| 1 | 0.02 | 92.006 | 0.024 | 96.025 | 0.0 (91.05) | TBL | 91.000 (0.04) | LPG-s |
| 2 | 0.16 | 169.021 | 0.166 | 127.460 | 0.01 (110.19) | LPG-td,TBL | 104.001 (0.03) | LPG-s |
| 3 | 0.04 | 67.012 | 0.039 | 59.215 | 0.01 (40.04) | LPG-s,TBL | 40.02 (0.116) | Sapa |
| 4 | 0.25 | 129.018 | 0.150 | 112.460 | 0.01 (110.15) | TBL | 99.001 (0.02) | LPG-s |
| 5 | 0.11 | 106.017 | 0.223 | 92.054 | 0.01 (83.17) | LPG-td,TBL | 75.08 (0.957) | Sapa |
| 6 | 0.09 | 114.011 | 0.241 | 93.031 | 0.02 (74.000) | LPG-td | 64.070 (0.963) | Sapa |
| 7 | 0.05 | 58.011 | 0.086 | 63.838 | 0.01 (51.09) | TBL | 49.090 (0.3) | Sapa |
| 8 | 0.19 | 151.023 | 0.238 | 153.071 | 0.01 (167.24) | TBL | 77.090 (0.869) | Sapa |
| 9 | 0.42 | 213.026 | 0.301 | 204.692 | 0.04 (232.26) | LPG-td,TBL | 150.002 (0.13) | LPG-s |
| 10 | 0.04 | 84.021 | 0.131 | 93.645 | 0.01 (71.11) | TBL | 49.090 (0.404) | Sapa |
| 11 | 0.04 | 108.019 | 0.153 | 104.058 | 0.01 (119.18) | TBL | 85.090 (0.474) | Sapa |
| 12 | 3.22 | 380.041 | 1.012 | 339.586 | 0.05 (274.3) | TBL | 274.3 (0.05) | TBL |
| 13 | 1.09 | 283.039 | 0.650 | 274.338 | 0.31 (299.000) | LPG-td | 240.003 (0.76) | LPG-s |
| 14 | 1.45 | 240.036 | 1.790 | 291.536 | 0.24 (391.000) | LPG-td | 163.22 (4.388) | Sapa |
| 15 | 0.25 | 283.043 | 3.299 | 278.920 | 0.14 (278.34) | TBL | 157.210 (13.457) | Sapa |
| 16 | | | 112.980 | 1849.014 | 50.78 (1510.000) | LPG-td | 1510.000 (50.78) | LPG-td |
| 17 | | | 7.133 | 790.049 | 1.86 (1052.000) | LPG-td | 653.008 (16.28) | LPG-s |
| 18 | | | 64.420 | 644.560 | 21.03 (869.000) | LPG-td | 361.67 (79.06) | TBL |
| 19 | | | 47.080 | 1478.000 | 47.08 (1478.000) | LPG-td | 1478.000 (47.08) | LPG-td |
| 20 | | | 6.190 | 478.000 | 6.19 (478.000) | LPG-td | 478.000 (6.19) | LPG-td |

Table 6: Results for Simple Domains: Best results show best time (corresponding quality) and which planner(s) achieved this time and best quality (corresponding time) and planner(s) achieving this quality. TBL is the Temporal Baseline planner in this and following tables.

| | COLIN | | Average | | Best | | | |
|---|---|---|---|---|---|---|---|---|
| | Time | Quality | Time | Quality | Time | Planner | Quality | Planner |
| roverssimpletime | | | | | | | | |
| 1 | 0.02 | 67.007 | 0.032 | 72.834 | 0.0 (67.08) | TBL | 67.007 (0.02) | COLIN |
| 2 | 0.02 | 48.006 | 0.030 | 48.223 | 0.01 (47.06) | TBL | 45.000 (0.02) | LPG-s |
| 3 | 0.02 | 73.01 | 0.040 | 76.238 | 0.0 (77.000) | LPG-td | 67.089993 (0.15) | Sapa |
| 4 | 0.02 | 50.005 | 0.030 | 51.621 | 0.01 (50.06) | TBL | 45.039997 (0.081) | Sapa |
| 5 | 0.04 | 126.014 | 0.220 | 131.461 | 0.01 (107.15) | LPG-td,TBL | 107.15 (0.01) | TBL |
| 6 | 0.09 | 186.028 | 0.058 | 229.580 | 0.01 (277.000) | LPG-td | 183.29 (0.02) | TBL |
| 7 | 0.04 | 107.013 | 0.146 | 97.851 | 0.01 (98.13) | TBL | 91.001 (0.04) | LPG-s |
| 8 | 0.06 | 119.018 | 0.247 | 131.066 | 0.02 (113.19) | TBL | 113.19 (0.02) | TBL |
| 9 | 0.09 | 176.028 | 0.148 | 166.296 | 0.02 (159.000) | LPG-td,TBL | 156.18999 (0.508) | Sapa |
| 10 | 0.11 | 155.019 | 5.110 | 158.878 | 0.02 (141.23) | TBL | 139.14 (25.24) | Sapa |
| 11 | 0.11 | 161.025 | 0.085 | 179.322 | 0.02 (185.26) | TBL | 161.025 (0.11) | COLIN |
| 12 | 0.05 | 103.014 | 0.253 | 112.041 | 0.01 (90.11) | TBL | 88.08 (1.035) | Sapa |
| 13 | 0.22 | 198.029 | 1.565 | 227.708 | 0.06 (190.33) | TBL | 173.17998 (6.716) | Sapa |
| 14 | 0.12 | 170.021 | 0.239 | 157.483 | 0.02 (145.22) | TBL | 139.000 (0.04) | LPG-td |
| 15 | 0.21 | 208.038 | 0.671 | 207.310 | 0.04 (215.29) | TBL | 175.21999 (2.776) | Sapa |
| 16 | 0.26 | 203.032 | 1.464 | 208.513 | 0.04 (242.000) | LPG-td | 186.20998 (6.67) | Sapa |
| 17 | 0.23 | 267.036 | 1.992 | 255.936 | 0.11 (259.000) | LPG-td | 218.22993 (9.01) | Sapa |
| 18 | 0.49 | 217.039 | 2.158 | 170.494 | 0.1 (160.000) | LPG-td | 140.28 (0.15) | TBL |
| 19 | 0.72 | 339.047 | 1.058 | 319.138 | 0.23 (319.000) | LPG-td | 297.5 (0.38) | TBL |
| 20 | 9.51 | 392.063 | 4.317 | 380.932 | 0.49 (351.000) | LPG-td | 351.000 (0.49) | LPG-td |
| satellitesimpletime | | | | | | | | |
| 1 | 0.01 | 41.008 | 0.022 | 43.032 | 0.0 (46.07) | TBL | 41.000 (0.01) | LPG-td |
| 2 | 0.01 | 65.012 | 0.042 | 66.310 | 0.01 (65.012) | COLIN,TBL | 65.000 (0.02) | LPG-td |
| 3 | 0.02 | 50.01 | 0.049 | 45.040 | 0.01 (58.09) | TBL | 29.000 (0.02) | LPG-td |
| 4 | 0.04 | 87.019 | 0.080 | 78.268 | 0.01 (58.000) | LPG-td,TBL | 58.000 (0.01) | LPG-td |
| 5 | 0.05 | 74.016 | 0.103 | 75.259 | 0.01 (82.13) | TBL | 61.001 (0.09) | LPG-s |
| 6 | 0.07 | 72.019 | 0.128 | 72.846 | 0.01 (58.09) | TBL | 58.09 (0.01) | TBL |
| 7 | 0.09 | 72.022 | 0.179 | 67.655 | 0.02 (75.12) | TBL | 46.001 (0.13) | LPG-s |
| 8 | 0.14 | 84.024 | 0.266 | 73.267 | 0.03 (41.000) | LPG-td,TBL | 41.000 (0.03) | LPG-td |
| 9 | 0.21 | 95.028 | 0.517 | 70.866 | 0.04 (58.000) | LPG-td,TBL | 51.001 (0.33) | LPG-s |
| 10 | 0.26 | 101.029 | 0.563 | 79.678 | 0.05 (63.000) | LPG-td | 63.000 (0.05) | LPG-td |
| 11 | 0.37 | 113.031 | 0.806 | 90.667 | 0.07 (72.000) | LPG-td,TBL | 72.000 (0.07) | LPG-td |
| 12 | 2.49 | 137.041 | 2.183 | 119.533 | 0.11 (98.000) | LPG-td | 98.000 (0.11) | LPG-td |
| 13 | 13.38 | 214.061 | 8.604 | 165.517 | 0.2 (208.000) | LPG-td | 99.002 (2.75) | LPG-s |
| 14 | 5.11 | 166.039 | 4.859 | 111.078 | 0.13 (68.000) | LPG-td | 68.000 (0.13) | LPG-td |
| 15 | 6.47 | 183.056 | 10.845 | 152.335 | 0.17 (183.000) | LPG-td | 58.001 (2.54) | LPG-s |
| 16 | 5.78 | 170.045 | 13.916 | 143.923 | 0.23 (137.000) | LPG-td | 73.002 (4.12) | LPG-s |
| 17 | 4.52 | 133.041 | 16.070 | 124.495 | 0.21 (142.000) | LPG-td | 92.002 (4.29) | LPG-s |
| 18 | 0.82 | 107.031 | 2.084 | 93.269 | 0.1 (101.000) | LPG-td | 75.002 (1.42) | LPG-s |
| 19 | 27.98 | 349.075 | 9.740 | 170.718 | 0.17 (130.000) | LPG-td | 114.003 (2.57) | LPG-s |
| 20 | | | 9.739 | 231.793 | 0.26 (196.000) | LPG-td | 137.24 (31.225) | Sapa |

Table 7: Results for Simple Domains: Best results show best time (corresponding quality) and which planner(s) achieved this time and best quality (corresponding time) and planner(s) achieving this quality.

| | COLIN | | Average | | Best | | | |
|---|---|---|---|---|---|---|---|---|
| | Time | Quality | Time | Quality | Time | Planner | Quality | Planner |
| zenosimpletime | | | | | | | | |
| 1 | 0.01 | 173.001 | 0.016 | 178.602 | 0.0 (180) | TBL | 173.001 (0.01) | COLIN |
| 2 | 0.02 | 592.008 | 0.033 | 666.022 | 0.01 (866.05) | TBL | 592.008 (0.02) | COLIN |
| 3 | 0.02 | 350.007 | 0.047 | 356.619 | 0.01 (280.04) | TBL | 280.04 (0.01) | TBL |
| 4 | 0.04 | 885.013 | 0.062 | 1226.041 | 0.01 (936.000) | LPG-td,TBL | 885.013 (0.04) | COLIN |
| 5 | 0.03 | 656.011 | 0.121 | 868.832 | 0.0 (400.06) | TBL | 400.06 (0.0) | TBL |
| 6 | 0.05 | 995.016 | 0.191 | 1456.434 | 0.01 (603.06) | TBL | 603.06 (0.01) | TBL |
| 7 | 0.08 | 931.014 | 0.218 | 1006.035 | 0.01 (706.08) | TBL | 706.08 (0.01) | TBL |
| 8 | 0.15 | 895.013 | 0.440 | 869.291 | 0.02 (836.07) | TBL | 836.07 (0.02) | TBL |
| 9 | 0.31 | 1583.027 | 1.021 | 1268.456 | 0.03 (789.12) | TBL | 789.12 (0.03) | TBL |
| 10 | 0.12 | 1191.022 | 0.696 | 1461.053 | 0.03 (743.13) | TBL | 743.13 (0.03) | TBL |
| 11 | 0.26 | 796.015 | 1.785 | 823.433 | 0.03 (763.1) | TBL | 510.050 (8.037) | Sapa |
| 12 | 0.3 | 1208.027 | 1.900 | 1617.856 | 0.03 (1199.13) | TBL | 1166.120 (8.568) | Sapa |
| 13 | 6.19 | 2062.04 | 5.261 | 1303.254 | 0.03 (923.14) | TBL | 923.14 (0.03) | TBL |
| 14 | | | 360.254 | 1577.821 | 0.3 (2068.18) | TBL | 1169.100 (1433.534) | Sapa |
| 15 | 19.48 | 2836.051 | 10.105 | 2561.808 | 0.46 (2254.18) | TBL | 2254.18 (0.46) | TBL |
| 16 | 477.47 | 3173.045 | 127.658 | 2394.072 | 0.86 (1702.24) | TBL | 1702.24 (0.86) | TBL |
| 17 | 759.19 | 4947.071 | 246.098 | 5232.603 | 2.36 (3436.34) | TBL | 3436.34 (2.36) | TBL |
| 18 | | | 43.737 | 2983.101 | 2.61 (3453.3) | TBL | 2383.003 (121.90) | LPG-s |
| 19 | 48.95 | 4309.079 | 22.740 | 5043.146 | 8.91 (3769.36) | TBL | 3769.36 (8.91) | TBL |
| 20 | 163.2 | 5364.096 | 59.687 | 5315.165 | 6.43 (4578.4) | TBL | 4578.4 (6.43) | TBL |

Table 8: Results for Simple Domains: Best results show best time (corresponding quality) and which planner(s) achieved this time and best quality (corresponding time) and planner(s) achieving this quality.

| | COLIN | | Average | | Best | | | |
|---|---|---|---|---|---|---|---|---|
| | Time | Quality | Time | Quality | Time | Planner | Quality | Planner |
| airportstripstemporal | | | | | | | | |
| 1 | 0.06 | 64.007 | 0.030 | 64.026 | 0.01 (64.07) | TBL | 64.000 (0.02) | LGP-td |
| 2 | 0.06 | 185.008 | 0.030 | 185.022 | 0.01 (185.000) | LGP-td,TBL | 185.000 (0.01) | LGP-td |
| 3 | 0.09 | 202.011 | 0.040 | 200.533 | 0.01 (200.12) | TBL | 200.000 (0.03) | LGP-td |
| 4 | 0.30 | 127.019 | 0.120 | 127.070 | 0.02 (127.19) | TBL | 127.000 (0.04) | LGP-td |
| 5 | 0.26 | 227.02 | 0.103 | 227.055 | 0.02 (227.2) | TBL | 227.000 (0.05) | LGP-td |
| 6 | 0.30 | 301.04 | 0.130 | 251.363 | 0.02 (240.41) | TBL | 232.000 (0.07) | LGP-td |
| 7 | 0.30 | 301.04 | 0.130 | 251.363 | 0.02 (240.41) | TBL | 232.000 (0.07) | LGP-td |
| 8 | 0.52 | 538.059 | 0.237 | 432.165 | 0.06 (402.6) | TBL | 394.000 (0.09) | LGP-td |
| 9 | 1.40 | 516.058 | 1.672 | 433.918 | 0.10 (402.000) | LGP-td | 402.000 (0.10) | LGP-td |
| 10 | 0.33 | 126.017 | 0.130 | 126.062 | 0.01 (126.17) | TBL | 126.000 (0.05) | LGP-td |
| 11 | 0.35 | 228.02 | 0.135 | 228.055 | 0.02 (228.2) | TBL | 228.000 (0.07) | LGP-td |
| 12 | 0.36 | 265.035 | 0.155 | 239.352 | 0.02 (228.37) | TBL | 228.37 (0.02) | TBL |
| 13 | 0.44 | 311.034 | 0.170 | 254.852 | 0.03 (237.37) | TBL | 230.002 (0.14) | LGP-s |
| 14 | 0.71 | 528.057 | 0.465 | 426.657 | 0.07 (390.57) | TBL | 390.57 (0.07) | TBL |
| 15 | 0.58 | 365.049 | 0.257 | 357.633 | 0.06 (273.48) | TBL | 273.48 (0.06) | TBL |
| 16 | 1.68 | 625.076 | 2.738 | 536.440 | 0.14 (558.000) | LGP-td | 404.68 (0.24) | TBL |
| 17 | 11.47 | 603.084 | 6.093 | 542.947 | 0.61 (569.000) | LGP-td | 417.7 (2.16) | TBL |
| 18 | 58.89 | 777.095 | 25.983 | 652.658 | 0.25 (733.000) | LGP-td | 447.88 (18.81) | TBL |
| 19 | 38.65 | 574.076 | 226.607 | 453.467 | 0.50 (413.000) | LGP-td | 413.000 (0.50) | LGP-td |
| 20 | | | 66.563 | 643.625 | 0.27 (740.000) | LGP-td | 450.87 (55.03) | TBL |
| 21 | 21.65 | 366.1 | 80.240 | 305.519 | 0.33 (285.97) | TBL | 285.000 (0.77) | LGP-td |
| 22 | 22.52 | 487.147 | 6.178 | 428.104 | 0.78 (286.26) | TBL | 286.26 (0.78) | TBL |
| 23 | 23.68 | 510.166 | 6.425 | 369.363 | 0.49 (265.28) | TBL | 264.004 (510.166) | LGP-s |
| 24 | 31.00 | 826.156 | 8.307 | 493.835 | 0.61 (377.18) | TBL | 376.003 (826.156) | LGP-s |
| 25 | 90.81 | 934.204 | 34.577 | 650.231 | 2.47 (539.000) | LGP-td | 477.49 (10.45) | TBL |
| 26 | | | 53.480 | 480.785 | 3.88 (584.000) | LGP-td | 377.57 (103.08) | TBL |
| 27 | | | 5.190 | 511.111 | 7.50 (505.33) | TBL | 504.004 () | LGP-s |
| 28 | | | 94.760 | 623.910 | 7.53 (706.000) | LGP-td | 541.82 (181.99) | TBL |
| 29 | | | 94.955 | 596.410 | 4.47 (687.000) | LGP-td | 505.82 (185.44) | TBL |
| 30 | | | 5.420 | 641.000 | 5.42 (641.000) | LGP-td | 641.000 (5.42) | LGP-td |
| 31 | | | 6.820 | 834.000 | 6.82 (834.000) | LGP-td | 834.000 (6.82) | LGP-td |
| 32 | | | 443.075 | 724.570 | 7.78 (815.000) | LGP-td | 634.14 (878.37) | TBL |
| 33 | | | 14.950 | 859.000 | 14.95 (859.000) | LGP-td | 859.000 (14.95) | LGP-td |
| 34 | | | 15.710 | 879.000 | 15.71 (879.000) | LGP-td | 879.000 (15.71) | LGP-td |
| 35 | | | 17.110 | 887.000 | 17.11 (887.000) | LGP-td | 887.000 (17.11) | LGP-td |
| 36 | 46.83 | 408.108 | 16.063 | 350.705 | 1.36 (322.000) | LGP-td | 322.000 (1.36) | LGP-td |
| 37 | 78.05 | 672.136 | 36.820 | 620.047 | 32.41 (831.000) | LGP-td | 357.006 (672.136) | LGP-s |
| 38 | 49.89 | 847.217 | 17.833 | 580.741 | 3.61 (573.000) | LGP-td | 322.006 (847.217) | LGP-s |
| 39 | | | 476.650 | 962.000 | 476.65 (962.000) | LGP-td | 962.000 (476.65) | LGP-td |
| 40 | | | 91.395 | 510.004 | 182.79 (584.000) | LGP-td | 436.007 () | LGP-s |
| 41 | 52.64 | 895.241 | 24.803 | 596.749 | 21.77 (573.000) | LGP-td | 322.006 (895.241) | LGP-s |
| 42 | | | 841.960 | 579.000 | 841.96 (579.000) | LGP-td | 579.000 (841.96) | LGP-td |
| 43 | | | 35.450 | 639.506 | 70.90 (573.000) | LGP-td | 573.000 (70.90) | LGP-td |
| 44 | | | 701.750 | 331.000 | 701.75 (331.000) | LGP-td | 331.000 (701.75) | LGP-td |
| 46 | | | 289.270 | 737.000 | 289.27 (737.000) | LGP-td | 737.000 (289.27) | LGP-td |

Table 9: Results for More Complex Domains: Best results show best time (corresponding quality) and which planner(s) achieved this time and best quality (corresponding time) and planner(s) achieving this quality.

| | COLIN | | Average | | Best | | | |
|---|---|---|---|---|---|---|---|---|
| | Time | Quality | Time | Quality | Time | Planner | Quality | Planner |
| depotstime | | | | | | | | |
| 1 | 0.03 | 59.746 | 0.036 | 57.698 | 0.00 (58.9311) | TBL | 55.181 (0.01) | LGP-td |
| 2 | 0.14 | 66.79 | 0.103 | 71.719 | 0.01 (73.2211) | TBL | 60.556 (0.02) | LGP-td |
| 3 | | | 0.087 | 141.362 | 0.03 (127.592) | TBL | 127.592 (0.03) | TBL |
| 4 | 67.41 | 258.125 | 17.043 | 186.587 | 0.04 (160.139) | LGP-td | 160.139 (0.04) | LGP-td |
| 5 | | | 2.970 | 781.855 | 2.82 (786.1666) | LGP-td | 777.544 (3.12) | TBL |
| 6 | | | 2.350 | 287.543 | 2.35 (287.543) | LGP-td | 287.543 (2.35) | LGP-td |
| 7 | 35.56 | 129.741 | 8.933 | 141.337 | 0.02 (142.158) | TBL | 107.340 (0.12) | LGP-s |
| 8 | | | 0.240 | 119.161 | 0.10 (108.3988) | LGP-td,TBL | 108.399 (0.10) | LGP-td |
| 9 | | | 0.650 | 1783.468 | 0.33 (1911.6665) | LGP-td | 1655.27 (0.97) | TBL |
| 10 | 43.34 | 296.354 | 11.027 | 226.698 | 0.03 (294.293) | TBL | 151.173 (11.364) | Sapa |
| 11 | | | 1.020 | 726.121 | 0.32 (608.9167) | LGP-td | 599.323 (1.42) | LGP-s |
| 12 | | | 5.387 | 163.032 | 0.46 (165.5889) | LGP-td | 132.893 (4.33) | LGP-s |
| 13 | 0.27 | 108.493 | 0.417 | 99.751 | 0.02 (104.436) | TBL | 82.834 (1.373) | Sapa |
| 14 | | | 0.427 | 216.623 | 0.10 (329.133) | TBL | 157.701 (1.02) | LGP-s |
| 15 | | | 2.290 | 399.697 | 0.42 (416.4197) | LGP-td | 382.973 (4.16) | LGP-s |
| 16 | 0.60 | 50.588 | 4.754 | 76.900 | 0.03 (166.236) | TBL | 26.390 (0.56) | LGP-s |
| 17 | | | 1.977 | 87.495 | 0.19 (88.26) | TBL | 46.2 (0.20) | LGP-s |
| 18 | | | 3.077 | 535.942 | 0.75 (709.8509) | LGP-td | 395.768 (0.89) | TBL |
| 19 | | | 0.520 | 428.210 | 0.14 (486.722) | TBL | 366.474 (1.22) | LGP-s |
| 20 | | | 3.743 | 880.639 | 0.93 (1050.2797) | LGP-td | 592.561 (7.98) | LGP-s |
| 21 | 2.08 | 76.753 | 9.693 | 71.388 | 0.16 (101.373) | TBL | 54.731 (36.794) | Sapa |
| 22 | | | 47.470 | 445.232 | 3.86 (343.1095) | LGP-td | 343.110 (3.86) | LGP-td |
| driverlogtime | | | | | | | | |
| 1 | 0.02 | 303.006 | 0.017 | 302.625 | 0.00 (302.05) | TBL | 302 (0.01) | LGP-td |
| 2 | 0.17 | 462.019 | 0.178 | 442.260 | 0.01 (438.19) | LGP-td,TBL | 294.090 (0.678) | Sapa |
| 3 | 0.03 | 202.009 | 0.044 | 279.018 | 0.01 (173.06) | TBL | 173.020 (0.131) | Sapa |
| 4 | 0.14 | 474.019 | 0.128 | 453.060 | 0.01 (441.15) | TBL | 402.001 (0.04) | LGP-s |
| 5 | 0.20 | 343.019 | 0.249 | 250.058 | 0.01 (195.18) | TBL | 161.090 (0.973) | Sapa |
| 6 | 0.03 | 275.007 | 0.280 | 266.432 | 0.02 (260) | LGP-td | 260 (0.02) | LGP-td |
| 7 | 0.12 | 316.012 | 0.072 | 350.041 | 0.01 (268.1) | LGP-td,TBL | 268.1 (0.01) | TBL |
| 8 | 0.43 | 699.024 | 0.361 | 695.275 | 0.01 (894.24) | TBL | 388.110 (1.244) | Sapa |
| 9 | 0.57 | 730.028 | 0.193 | 748.078 | 0.02 (591) | LGP-td | 591 (0.02) | LGP-td |
| 10 | 0.21 | 294.031 | 0.159 | 328.450 | 0.01 (394.11) | TBL | 220.110 (0.415) | Sapa |
| 11 | 0.16 | 522.026 | 0.181 | 515.068 | 0.01 (512.18) | TBL | 306.13 (0.486) | Sapa |
| 12 | 5.90 | 1066.038 | 233.369 | 1230.325 | 0.05 (829.32) | TBL | 627.260 (1151.634) | Sapa |
| 13 | 2.84 | 1052.031 | 1.743 | 1218.305 | 0.32 (903) | LGP-td | 597.180 (4.193) | Sapa |
| 14 | 6.51 | 787.034 | 3.248 | 1125.322 | 0.41 (1161) | LGP-td | 625.150 (7.34) | Sapa |
| 15 | 0.27 | 212.042 | 59.469 | 766.155 | 0.10 (939.43) | TBL | 212.042 (0.27) | COLIN |
| 16 | | | 66.385 | 5120.504 | 28.03 (3652.0081) | LGP-s | 3652.008 (28.03) | LGP-s |
| 17 | | | 9.580 | 2787.062 | 3.31 (2238) | LGP-td | 2238 (3.31) | LGP-td |
| 18 | | | 48.780 | 2319.229 | 27.48 (2885) | LGP-td | 1476.68 (78.94) | TBL |
| 19 | | | 110.925 | 4005.005 | 63.87 (3993) | LGP-td | 3993 (63.87) | LGP-td |
| 20 | | | 41.060 | 3828.003 | 15.56 (5965) | LGP-td | 1691.007 (66.56) | LGP-s |

Table 10:  Results for More Complex Domains: Best results show best time (corresponding quality) and which planner(s) achieved this time and best quality (corresponding time) and planner(s) achieving this quality.

| | COLIN | | Average | | Best | | | |
|---|---|---|---|---|---|---|---|---|
| | Time | Quality | Time | Quality | Time | Planner | Quality | Planner |
| pipesnotankagetemporal | | | | | | | | |
| 1 | 0.03 | 6.003 | 0.015 | 7.506 | 0.00 (6.02) | TBL | 6.000 (0.01) | LGP-td |
| 2 | 0.04 | 20.011 | 0.058 | 51.526 | 0.01 (20.09) | TBL | 20.011 (0.04) | COLIN |
| 3 | 0.05 | 12.008 | 0.040 | 17.520 | 0.01 (16.07) | TBL | 12.008 (0.05) | COLIN |
| 4 | 0.07 | 22.012 | 0.058 | 83.021 | 0.01 (16.07) | TBL | 16.07 (0.01) | TBL |
| 5 | 0.05 | 14.006 | 0.043 | 13.017 | 0.01 (12.000) | LGP-td,TBL | 12.000 (0.01) | LGP-td |
| 6 | 0.05 | 16.009 | 0.060 | 24.023 | 0.01 (18.08) | TBL | 16.009 (0.05) | COLIN |
| 7 | 0.06 | 12.007 | 0.075 | 16.015 | 0.01 (12.05) | TBL | 12.007 (0.06) | COLIN |
| 8 | 0.06 | 16.009 | 0.100 | 16.520 | 0.03 (18.000) | LGP-td | 16.001 (0.19) | LGP-s |
| 9 | 0.08 | 24.013 | 0.113 | 29.026 | 0.02 (20.09) | TBL | 20.09 (0.02) | TBL |
| 10 | 0.08 | 28.015 | 0.120 | 36.037 | 0.02 (28.13) | TBL | 28.015 (0.08) | COLIN |
| 11 | 0.16 | 11.026 | 300.430 | 9.544 | 0.05 (8.15) | TBL | 8.15 (0.05) | TBL |
| 12 | 1.77 | 22.554 | 302.110 | 16.971 | 0.28 (17.33) | TBL | 11.002 (1200.92) | LGP-s |
| 13 | 0.27 | 16.535 | 0.745 | 15.687 | 0.05 (11.21) | TBL | 11.21 (0.05) | TBL |
| 14 | 0.38 | 14.532 | 0.380 | 14.446 | 0.15 (14.000) | LGP-td | 13.25 (0.99) | TBL |
| 15 | 0.14 | 11.526 | 0.785 | 16.700 | 0.12 (14.27) | TBL | 11.526 (0.14) | COLIN |
| 16 | 6.78 | 30.07 | 2.667 | 31.401 | 0.48 (40.000) | LGP-td | 27.53 (3.41) | TBL |
| 17 | 4.25 | 11.023 | 1.143 | 10.544 | 0.09 (8.15) | TBL | 8.002 (11.023) | LGP-s |
| 18 | 1.08 | 12.529 | 0.900 | 18.221 | 0.20 (18.35) | TBL | 12.529 (1.08) | COLIN |
| 19 | 0.19 | 9.525 | 0.247 | 9.898 | 0.11 (9.17) | TBL | 9.17 (0.11) | TBL |
| 20 | 0.46 | 17.039 | 7.977 | 26.681 | 0.46 (0.46) | COLIN | 14.003 (17.039) | LGP-s |
| 21 | 0.09 | 9.017 | 0.080 | 9.062 | 0.02 (9.17) | TBL | 9.000 (0.13) | LGP-td |
| 22 | | | 2.975 | 36.700 | 2.05 (23.4) | TBL | 23.4 (2.05) | TBL |
| 23 | 0.19 | 14.018 | 0.240 | 21.340 | 0.19 (14.018) | COLIN | 14.018 (0.19) | COLIN |
| 24 | 40.46 | 29.032 | 27.015 | 101.016 | 13.57 (173.000) | LGP-td | 29.032 (40.46) | COLIN |
| 25 | | | 4.850 | 32.165 | 1.12 (42.000) | LGP-td | 22.33 (8.58) | TBL |
| 26 | 1.89 | 32.053 | 4.140 | 28.148 | 1.46 (27.39) | TBL | 25.000 (9.07) | LGP-td |
| 27 | 0.29 | 11.527 | 1.613 | 13.572 | 0.29 (11.527) | COLIN | 10.19 (0.48) | TBL |
| 28 | 0.77 | 22.551 | 16.697 | 42.977 | 0.77 (22.551) | COLIN | 22.551 (0.77) | COLIN |
| 29 | 2.46 | 21.546 | 5.560 | 18.272 | 0.34 (14.27) | TBL | 14.27 (0.34) | TBL |
| 30 | | | 4.665 | 32.205 | 1.39 (27.41) | TBL | 27.41 (1.39) | TBL |
| 31 | 9.21 | 18.357 | 3.240 | 16.098 | 0.17 (16.9367) | TBL | 13.000 (0.34) | LGP-td |
| 32 | 13.01 | 35.038 | 4.485 | 51.504 | 0.89 (17.31) | TBL | 17.31 (0.89) | TBL |
| 33 | 10.53 | 21.874 | 10.530 | 21.874 | 10.53 (21.874) | COLIN | 21.874 (10.53) | COLIN |
| 34 | 20.52 | 30.709 | 34.070 | 39.582 | 1.40 (23.370033) | TBL | 23.370033 (1.40) | TBL |
| 35 | | | 24.200 | 17.334 | 48.40 (24.000) | LGP-td | 10.669 () | LGP-s |
| 36 | | | 78.245 | 18.078 | 6.49 (21.156633) | TBL | 15.000 (150.00) | LGP-td |
| 37 | | | 31.760 | 36.000 | 31.76 (36.000) | LGP-td | 36.000 (31.76) | LGP-td |
| 38 | | | 64.855 | 12.822 | 0.32 (13.6433) | TBL | 12.000 (129.39) | LGP-td |
| 39 | 1.19 | 13.857 | 0.967 | 18.356 | 0.17 (12.21) | TBL | 12.21 (0.17) | TBL |
| 40 | | | 10.410 | 34.665 | 10.41 (34.665) | LGP-td | 34.665 (10.41) | LGP-td |
| 41 | 103.46 | 4.43 | 34.820 | 4.173 | 0.10 (4.09) | TBL | 4.000 (0.90) | LGP-td |
| 49 | | | 346.235 | 16.493 | 2.15 (16.32) | TBL | 16.32 (2.15) | TBL |
| 50 | | | 15.400 | 18.380 | 15.40 (18.38) | TBL | 18.38 (15.40) | TBL |

Table 11: Results for More Complex Domains: Best results show best time (corresponding quality) and which planner(s) achieved this time and best quality (corresponding time) and planner(s) achieving this quality.

| | COLIN | | Average | | Best | | | |
|---|---|---|---|---|---|---|---|---|
| | Time | Quality | Time | Quality | Time | Planner | Quality | Planner |
| pipestankagetemporal | | | | | | | | |
| 1 | 0.04 | 6.003 | 0.020 | 8.006 | 0.00 (6.02) | TBL | 6.000 (0.03) | LGP-s |
| 2 | 0.12 | 24.013 | 0.147 | 63.371 | 0.02 (22.1) | TBL | 22.1 (0.02) | TBL |
| 3 | 0.23 | 12.008 | 0.180 | 15.520 | 0.06 (16.07) | TBL | 12.008 (0.23) | COLIN |
| 4 | 0.52 | 18.009 | 0.275 | 26.020 | 0.04 (16.07) | TBL | 16.07 (0.04) | TBL |
| 5 | 0.09 | 14.007 | 0.153 | 20.017 | 0.02 (14.06) | TBL | 14.007 (0.09) | COLIN |
| 6 | 0.09 | 16.01 | 0.160 | 17.018 | 0.02 (14.06) | TBL | 14.06 (0.02) | TBL |
| 7 | 0.26 | 16.008 | 1.175 | 18.522 | 0.07 (18.08) | TBL | 16.000 (0.31) | LGP-td |
| 8 | 0.30 | 18.012 | 1.275 | 23.031 | 0.30 (18.012) | COLIN | 18.012 (0.30) | COLIN |
| 9 | | | 238.057 | 66.741 | 0.95 (104.000) | LGP-td | 46.22 (6.34) | TBL |
| 10 | 4.40 | 36.02 | 26.135 | 57.571 | 0.32 (54.26) | TBL | 36.02 (4.40) | COLIN |
| 11 | 0.40 | 13.031 | 316.585 | 27.601 | 0.40 (13.031) | COLIN | 13.003 (1235.31) | LGP-s |
| 12 | | | 12.480 | 27.105 | 5.08 (43.000) | LGP-td | 11.21 (19.88) | TBL |
| 13 | 1.51 | 13.53 | 6.420 | 13.265 | 1.51 (13.53) | COLIN | 13.000 (11.33) | LGP-td |
| 14 | 53.09 | 22.052 | 334.625 | 19.357 | 9.17 (19.37) | TBL | 18.000 (1276.24) | LGP-td |
| 15 | 59.61 | 17.541 | 64.690 | 14.886 | 59.61 (17.541) | COLIN | 12.23 (69.77) | TBL |
| 17 | | | 1500.780 | 31.000 | 1500.78 (31.000) | LGP-td | 31.000 (1500.78) | LGP-td |
| 18 | 33.06 | 11.527 | 105.157 | 28.626 | 33.06 (11.527) | COLIN | 11.527 (33.06) | COLIN |
| 19 | 20.29 | 16.537 | 252.310 | 17.582 | 17.59 (11.21) | TBL | 11.21 (17.59) | TBL |
| 20 | 2.54 | 12.029 | 118.323 | 24.610 | 2.54 (12.029) | COLIN | 12.029 (2.54) | colin |
| 21 | | | 1.830 | 14.595 | 0.11 (10.19) | TBL | 10.19 (0.11) | TBL |
| 22 | | | 77.570 | 25.690 | 53.50 (30.000) | LGP-td | 21.38 (101.64) | TBL |
| 23 | | | 41.080 | 30.000 | 41.08 (30.000) | LGP-td | 30.000 (41.08) | LGP-td |
| 24 | 23.25 | 31.564 | 112.000 | 37.782 | 23.25 (31.564) | COLIN | 31.564 (23.25) | COLIN |
| 25 | 983.78 | 28.558 | 710.420 | 56.483 | 527.52 (116.500) | LGP-td | 24.39 (619.96) | TBL |
| 26 | 32.55 | 33.053 | 389.255 | 31.991 | 32.55 (33.053) | COLIN | 30.93 (745.96) | TBL |
| 27 | 2.30 | 10.024 | 7.000 | 16.012 | 2.30 (10.024) | COLIN | 10.024 (2.30) | COLIN |
| 29 | 191.49 | 23.548 | 364.450 | 28.274 | 191.49 (23.548) | COLIN | 23.548 (191.49) | COLIN |
| 30 | 58.98 | 28.555 | 354.695 | 30.777 | 58.98 (28.555) | COLIN | 28.555 (58.98) | COLIN |
| 31 | 187.22 | 32.713 | 249.670 | 31.041 | 0.84 (31.7833) | TBL | 29.833 (810.62) | LGP-td |
| 32 | | | 696.495 | 34.158 | 179.24 (35.3167) | TBL | 33.000 (1213.75) | LGP-td |
| 33 | | | 304.470 | 21.133 | 304.47 (21.133333) | TBL | 21.133333 (304.47) | TBL |
| 34 | 160.42 | 25.038 | 432.870 | 40.065 | 31.22 (33.99) | TBL | 25.038 (160.42) | COLIN |
| 37 | | | 598.180 | 20.665 | 598.18 (20.665) | LGP-td | 20.665 (598.18) | LGP-td |
| 39 | 9.14 | 15.191 | 298.620 | 22.925 | 9.14 (15.191) | COLIN | 15.191 (9.14) | COLIN |
| 40 | 51.82 | 21.877 | 51.820 | 21.877 | 51.82 (21.877) | COLIN | 21.877 (51.82) | COLIN |
| 41 | 269.57 | 4.93 | 138.057 | 6.687 | 32.63 (6.13) | TBL | 4.93 (269.57) | COLIN |
| 49 | | | 91.770 | 17.340 | 91.77 (17.34) | TBL | 17.34 (91.77) | TBL |
| 50 | 82.82 | 22.378 | 82.820 | 22.378 | 82.82 (22.378) | COLIN | 22.378 (82.82) | COLIN |

Table 12: Results for More Complex Domains: Best results show best time (corresponding quality) and which planner(s) achieved this time and best quality (corresponding time) and planner(s) achieving this quality.

| | COLIN | | Average | | Best | | | |
|---|---|---|---|---|---|---|---|---|
| | Time | Quality | Time | Quality | Time | Planner | Quality | Planner |
| roverstime | | | | | | | | |
| 1 | 0.03 | 67.007 | 0.048 | 73.524 | 0.02 (80) | LGP-td | 67.007 (67.007) | COLIN |
| 2 | 0.01 | 48.006 | 0.027 | 56.269 | 0.01 (48.006) | COLIN | 46.0007 (0.02) | LGP-s |
| 3 | 0.03 | 63.01 | 0.062 | 70.525 | 0.02 (80) | LGP-td | 63.01 (0.03) | COLIN |
| 4 | 0.03 | 52.006 | 0.040 | 52.733 | 0.02 (52) | LGP-s,LGP-td | 52 (0.02) | LGP-td |
| 5 | 0.06 | 125.014 | 0.258 | 129.069 | 0.02 (113.2) | LGP-td | 107.12 (0.702) | Sapa |
| 6 | 40.93 | 273.118 | 14.010 | 299.794 | 0.08 (284.965) | LGP-td | 273.118 (40.93) | COLIN |
| 7 | 0.07 | 105.017 | 0.138 | 107.121 | 0.03 (138.9286) | LGP-td | 90.538574 (0.411) | Sapa |
| 8 | 0.17 | 149.944 | 0.494 | 160.906 | 0.04 (134) | LGP-td | 134 (0.04) | LGP-td |
| 9 | | | 0.150 | 137.345 | 0.06 (146) | LGP-td | 128.6895 (0.24) | LGP-s |
| 10 | 0.29 | 177.022 | 0.534 | 192.855 | 0.06 (190.3077) | LGP-td | 154.48767 (1.448) | Sapa |
| 11 | 0.15 | 170.881 | 0.170 | 195.395 | 0.07 (200.5) | LGP-td | 170.881 (0.15) | COLIN |
| 12 | 0.86 | 122.022 | 0.476 | 134.818 | 0.04 (145.5294) | LGP-td | 114.130005 (0.682) | Sapa |
| 13 | | | 0.560 | 292.200 | 0.26 (346.5833) | LGP-td | 237.8161 (0.86) | LGP-s |
| 14 | | | 0.865 | 206.147 | 0.35 (268.7368) | LGP-td | 137.7917 (0.62) | LGP-s |
| 15 | | | 0.500 | 239.333 | 0.10 (205.3755) | LGP-td | 205.3755 (0.10) | LGP-td |
| 16 | | | 0.460 | 210.000 | 0.46 (210) | LGP-td | 210 (0.46) | LGP-td |
| 17 | 0.70 | 230.036 | 2.223 | 341.230 | 0.34 (392.353) | LGP-td | 230.036 (0.70) | COLIN |
| 18 | 1.35 | 245.864 | 1.817 | 220.290 | 0.33 (155.0909) | LGP-td | 155.0909 (0.33) | LGP-td |
| 19 | | | 0.610 | 394.591 | 0.61 (394.5915) | LGP-td | 394.5915 (0.61) | LGP-td |
| 20 | 92.59 | 390.804 | 49.107 | 505.667 | 3.38 (502.7423) | LGP-td | 390.804 (92.59) | COLIN |
| satellitetime | | | | | | | | |
| 1 | 0.02 | 129.596 | 0.026 | 193.553 | 0.01 (205.28) | LGP-s,LGP-td,TBL | 129.596 (0.02) | COLIN |
| 2 | 0.01 | 182.916 | 0.038 | 225.716 | 0.00 (235.12) | LGP-td | 182.916 (0.01) | COLIN |
| 3 | 0.01 | 78.616 | 0.068 | 168.343 | 0.01 (78.616) | COLIN,TBL | 78.616 (0.01) | COLIN |
| 4 | 0.03 | 140.42 | 0.099 | 319.688 | 0.01 (359.28) | TBL | 140.42 (0.03) | COLIN |
| 5 | 0.11 | 290.12 | 0.154 | 262.900 | 0.01 (254.563) | TBL | 162.39699 (0.528) | Sapa |
| 6 | 0.10 | 114.38 | 0.138 | 255.894 | 0.01 (264.51) | TBL | 114.38 (0.10) | COLIN |
| 7 | 0.12 | 126.544 | 0.265 | 222.818 | 0.02 (296.16) | TBL | 126.544 (0.12) | COLIN |
| 8 | 0.30 | 139.608 | 0.469 | 205.241 | 0.03 (226.503) | TBL | 139.608 (0.30) | COLIN |
| 9 | 0.38 | 175.74 | 0.813 | 307.457 | 0.05 (351.8529) | LGP-td,TBL | 175.74 (0.38) | COLIN |
| 10 | 0.64 | 295.549 | 0.949 | 262.206 | 0.04 (231.485) | LGP-td | 231.485 (0.04) | LGP-td |
| 11 | 0.94 | 283.351 | 1.703 | 370.878 | 0.07 (283.916) | TBL | 283.351 (0.94) | COLIN |
| 12 | 3.57 | 336.599 | 9.147 | 423.836 | 0.13 (389.4588) | LGP-td | 336.599 (3.57) | COLIN |
| 13 | 21.89 | 433.308 | 29.900 | 504.815 | 0.23 (464.408) | LGP-td | 433.308 (21.89) | COLIN |
| 14 | 6.19 | 267.73 | 10.618 | 387.271 | 0.15 (461.855) | LGP-td | 267.73 (6.19) | COLIN |
| 15 | 8.76 | 292.436 | 34.225 | 333.289 | 0.18 (267.4431) | LGP-td | 264.6641 (2.48) | LGP-s |
| 16 | 22.18 | 336.356 | 34.882 | 510.100 | 0.23 (602.7849) | LGP-td | 336.356 (22.18) | COLIN |
| 17 | 15.29 | 232.66 | 94.390 | 380.053 | 0.24 (378.459) | LGP-td | 232.66 (15.29) | COLIN |
| 18 | 3.85 | 169.256 | 5.306 | 290.099 | 0.10 (324.406) | TBL | 169.256 (3.85) | COLIN |
| 19 | 57.66 | 520.602 | 49.424 | 527.707 | 0.19 (352.355) | LGP-td | 352.355 (0.19) | LGP-td |
| 20 | | | 30.264 | 854.958 | 0.24 (584.663) | LGP-td | 498.90106 (113.277) | Sapa |

Table 13: Results for More Complex Domains: Best results show best time (corresponding quality) and which planner(s) achieved this time and best quality (corresponding time) and planner(s) achieving this quality.

| | COLIN | | Average | | Best | | | |
|---|---|---|---|---|---|---|---|---|
| | Time | Quality | Time | Quality | Time | Planner | Quality | Planner |
| zenotime | | | | | | | | |
| 1 | 0.01 | 3.672 | 0.018 | 3.489 | 0.01 (0.01) | COLIN,LPG-td | 3.424 (0.01) | LPG-td |
| 2 | 0.01 | 23.435 | 0.019 | 23.672 | 0.01 (0.01) | COLIN,LPG-td | 23.431 (0.01) | LPG-td |
| 3 | 0.03 | 10.089 | 0.033 | 13.336 | 0.01 (14.4211) | LPG-td | 10.089 (0.03) | COLIN |
| 4 | 0.08 | 21.287 | 0.084 | 22.340 | 0.01 (21.708) | LPG-s | 21.287 (0.08) | COLIN |
| 5 | 0.04 | 8.196 | 0.075 | 22.016 | 0.02 (27.0419) | LPG-td | 8.196 (0.04) | COLIN |
| 6 | 0.08 | 21.966 | 0.075 | 20.921 | 0.02 (20.8864) | LPG-td | 16.578 (0.17) | Sapa |
| 7 | 0.06 | 33.31 | 0.110 | 24.863 | 0.01 (25.6744) | LPG-td | 18.283 (0.05) | LPG-s |
| 8 | 8.53 | 43.722 | 2.250 | 30.689 | 0.02 (24.2375) | LPG-td | 24.238 (0.02) | LPG-td |
| 9 | 0.22 | 39.949 | 0.194 | 48.528 | 0.03 (72.1579) | LPG-td | 23.238 (0.395) | Sapa |
| 10 | 0.41 | 36.458 | 0.310 | 33.487 | 0.06 (46.1848) | LPG-td | 20.864 (0.14) | LPG-s |
| 11 | 0.41 | 22.264 | 0.271 | 25.576 | 0.05 (46.1576) | LPG-td | 13.666 (0.443) | Sapa |
| 12 | 10.34 | 72.139 | 2.874 | 51.105 | 0.09 (42.1671) | LPG-td | 38.992 (0.846) | Sapa |
| 13 | 1.28 | 86.473 | 1.592 | 58.235 | 0.07 (42.0593) | LPG-td | 42.059 (0.07) | LPG-td |
| 14 | 371.61 | 117.625 | 256.813 | 72.619 | 0.62 (58.8983) | LPG-td | 39.064 (651.493) | Sapa |
| 15 | 6.95 | 381.626 | 5.524 | 241.318 | 0.87 (274.8496) | LPG-td | 117.171 (8.644) | Sapa |
| 16 | 130.33 | 117.052 | 43.183 | 86.693 | 3.07 (67.554) | LPG-td | 54.717 (23.883) | Sapa |
| 17 | 443.79 | 77.332 | 161.243 | 118.179 | 3.97 (117.1512) | LPG-td | 77.332 (443.79) | COLIN |
| 18 | 188.85 | 89.082 | 77.357 | 70.542 | 4.48 (56.8345) | LPG-td | 56.835 (4.48) | LPG-td |
| 19 | | | 73.040 | 137.909 | 9.83 (168.0886) | LPG-td | 107.729 (136.25) | LPG-s |
| 20 | | | 80.660 | 91.146 | 28.69 (78.4703) | LPG-td | 78.470 (28.69) | LPG-td |

Table 14: Results for More Complex Domains: Best results show best time (corresponding quality) and which planner(s) achieved this time and best quality (corresponding time) and planner(s) achieving this quality.